\documentclass[11pt]{article}

\usepackage[final]{acl}

\usepackage{times}
\usepackage{latexsym}

\usepackage[T1]{fontenc}

\usepackage[utf8]{inputenc}

\usepackage{microtype}

\usepackage{inconsolata}

\usepackage{graphicx}

\usepackage{amsmath}
\usepackage{amssymb}
\usepackage{mathtools}
\usepackage{amsthm}
\usepackage{booktabs}
\usepackage{subcaption}

\usepackage{adjustbox}
\usepackage{bbm}
\usepackage{enumitem}
\def\ours{CoDyRA}

\definecolor{lightgrey}{rgb}{0.95, 0.95, 0.95}
\definecolor{softblue}{rgb}{0.22, 0.59, 1.00}
\definecolor{softorange}{rgb}{1.00, 0.52, 0.00}
\definecolor{softgreen}{rgb}{0.45, 0.62, 0.5}
\definecolor{lightorange}{RGB}{255, 223, 186}

\usepackage{multirow}
\usepackage{colortbl}
\usepackage[dvipsnames]{xcolor}
\usepackage{makecell}
\usepackage{rotating}

\usepackage{amsmath,amsfonts,bm}

\def\eqref#1{equation~\ref{#1}}

\def\1{\bm{1}}

\DeclareMathAlphabet{\mathsfit}{\encodingdefault}{\sfdefault}{m}{sl}
\SetMathAlphabet{\mathsfit}{bold}{\encodingdefault}{\sfdefault}{bx}{n}

\usepackage[dvipsnames]{xcolor}
\definecolor{FigComp}{HTML}{C8E4F8}      
\definecolor{FigCompAux}{HTML}{EAF4FF}   
\definecolor{FigCompNone}{HTML}{F2F2F2}  
\definecolor{FigCont}{HTML}{F5E4F0}      
\definecolor{FigRef}{HTML}{FFF3E8}       
\definecolor{FigNaive}{HTML}{F0EEDC}     
\definecolor{FigCap}{HTML}{F3ECFF}       
\usepackage[most]{tcolorbox}
\newtcolorbox{takeawaybox}{
  enhanced, breakable,
  colback=Purple!6,
  colframe=Purple!35,
  boxrule=0.6pt,
  arc=2.5pt, outer arc=2.5pt,
  left=8pt, right=8pt, top=5pt, bottom=5pt,
  before skip=6pt, after skip=6pt,
}
\newcommand{\takeaway}[2]{\begin{takeawaybox}\textcolor{Purple}{\textbf{Takeaway~#1.}}~#2\end{takeawaybox}}
\theoremstyle{plain}
\newtheorem{proposition}{Proposition}
\newcommand{\highblue}[1]{{\textbf{\color[RGB]{30, 85, 170}#1}}}
\newcommand{\highred}[1]{{\textbf{\color[RGB]{220, 20, 60}#1}}}
\newcommand{\colora}[1]{{\textcolor{Green}{#1}}}
\newcommand{\colorb}[1]{{\textcolor{Purple}{#1}}}

\newcommand{\reva}[1]{#1}

\def\ie{\textit{i.e.}}

\title{Take Only What You Need: Rank Minimization \\as an Implicit Forgetting Regularizer in Continual Learning}

\author{
Haodong Lu$^{1,2}$\ \ \ \ Chongyang Zhao$^{1}$\ \ \ \  Minhui Xue$^2$\ \ \ \ \\ \textbf{Lina Yao}$^{1}$\ \ \ \  \textbf{Kristen Moore}$^2$\ \ \ \  \textbf{Dong Gong}$^1$\thanks{D. Gong is the corresponding author. H. Lu was supported by a CSIRO's scholarship.}\\
$^1$University of New South Wales,~~$^2$CSIRO\\
\texttt{ \small \{haodong.lu, chongyang.zhao, lina.yao, dong.gong\}@unsw.edu.au}, \\ \texttt{ \small \{jason.xue, kristen.moore\}@csiro.au}
}

\begin{document}
\maketitle
\begin{abstract}
The central tension in continual learning (CL) is the trade-off between \emph{plasticity} (acquiring new knowledge) and \emph{stability} (retaining prior knowledge). We study how a pre-trained backbone can be continually updated to absorb new knowledge while preserving existing capabilities, via \emph{capacity control}: regulating the \emph{effective rank} of each parameter update, a per-step quantity directly controllable inside a LoRA update. A controlled probe of LoRA rank and placement across modules and tasks reveals a consistent trade-off, with a moderate-rank sweet spot that varies by placement and task, leaving no universally optimal fixed rank; a formal bound links forgetting to update rank. Building on these findings, we propose \textbf{Co}ntinual \textbf{Dy}namic \textbf{R}ank-Selective LoR\textbf{A} (\textbf{CoDyRA}), which jointly trains each LoRA update with rank minimization via sparsity-promoting regularization on per-component importance weights. The supervised objective drives plasticity; rank minimization regularizes forgetting. We show that rank minimization serves as an implicit forgetting regularizer in the CL regime, protecting general capability and prior-task knowledge simultaneously by controlling forgetting against the current model state. Across MTIL, X-TAIL, and TRACE (CLIP, LLaMA, Gemma), CoDyRA outperforms prior CL methods on new knowledge learning and forgetting, achieving a strong plasticity--stability balance. 
Project page: \href{https://artificer-ai-lab.github.io/CoDyRA/}{\texttt{artificer-ai-lab.github.io/CoDyRA}}.

\end{abstract}

\section{Introduction}

\begin{figure}[!t]
    \centering
    \includegraphics[width=\linewidth]{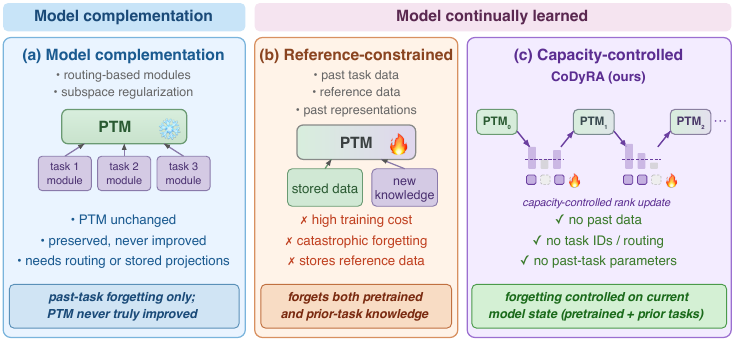}
    \vspace{-0.6cm}
    \caption{CL taxonomy. \textbf{Model complementation}: freeze the PTM and add external modules. \textbf{Model continually learned}: update the PTM directly. \textbf{(a) Model complementation}: per-task structure retained across the stream as task-specific auxiliary modules \citep{dualprompt,l2p,boosting,rail} or as maintained prior-task projections that keep new updates orthogonal \citep{wang2023orthogonal,inflora}. While yielding strong stability and performance, both rely on external modules alongside the PTM. \textbf{(b) Reference-constrained}: PTM updates constrained by stored data or memory \citep{li2017learning,wortsman2022robust,zscl}, at high training cost with visible forgetting. \textbf{(c) Capacity-controlled (ours)}: PTM updates bounded by effective rank, each task merged into the PTM under a single forgetting criterion (Sec.~\ref{sec:theory}) covering pretrained capability and prior tasks, with no past data, no task prediction, and no extra inference overhead.}
    \label{fig:concept}
    \vspace{-0.6cm}
\end{figure}

Continual learning (CL) \citep{hadsell2020embracing, de2021continual, wang2024comprehensive} hinges on balancing \colorb{plasticity} (acquiring new knowledge) and \colora{stability} (retaining prior knowledge) \citep{mccloskey1989catastrophic,nguyen2019toward}. With pre-trained models (PTMs), this trade-off sharpens into two existing regimes (Fig.~\ref{fig:concept}). \emph{Model complementation} methods (Fig.~\ref{fig:concept}(a)) retain per-task structure across the stream, either by routing each task through task-specific auxiliary modules \citep{dualprompt,l2p,boosting,rail,wang2024self} or by keeping new updates orthogonal to the maintained prior-task projections \citep{wang2023orthogonal,inflora}. While yielding strong stability and performance, both rely on external modules alongside the PTM. \emph{Model continually learned} methods update the PTM weights, with existing approaches constraining each step by stored data or memory (\emph{reference-constrained}, Fig.~\ref{fig:concept}(b)) \citep{rebuffi2017icarl,li2017learning,wortsman2022robust,zscl}, at high training cost, and visible forgetting persists. This raises a central CL question: \emph{how can we continually update the backbone to absorb new knowledge while preserving existing capabilities?}

To answer this, we study a general CL scenario (Fig.~\ref{fig:concept}(c)): unlike (a) and (b), at each step only the current task data and model state are available, with no external dependencies. The only design knob is \emph{how parameters are continually updated}. LoRA \citep{lora} is a natural fit: each low-rank update merges into the backbone, leaving the architecture and parameter count unchanged.

To better understand parameter-update dynamics, in a controlled study, we probe LoRA rank and placement across modules and tasks (Sec.~\ref{sec:motivation}). Our analyses reveal that \emph{not all} LoRA ranks and placements contribute \emph{equally} to downstream learning and forgetting: for arbitrary weights, \colorb{relatively higher-rank} LoRA facilitates learning new tasks (\ie, \colorb{\emph{plasticity}}) but tends to increase forgetting, while \colora{lower-rank} LoRA mitigates forgetting (\ie, \colora{\emph{stability}}) but limits adaptation. 
Importantly, we identify that there exists a \textbf{\emph{balance} between \colorb{\emph{plasticity}} (learning new knowledge) and \colora{\emph{stability}} (regularizing forgetting)} at a moderately small (but not too small) rank, potentially maximizing CL benefits. 
Moreover, we observe that this balance \emph{varies} significantly across different parameter placements. 

These findings identify a unified design knob, the \emph{effective rank} of each parameter update $\Delta\theta_t$, which we term \emph{capacity control}\label{sec:intro_capacity}: a per-step quantity distinct from model-level capacity (parameter count, VC dimension \citep{vapnik2013nature}). Rank is directly controllable in a LoRA update and bounds forgetting (Sec.~\ref{sec:theory}). Existing CL families adjust this lever only indirectly through replay, per-coordinate penalties, modular isolation, or fixed-rank direction constraints; we make update rank itself the design knob.

This calls for a mechanism that adaptively finds the per-module sweet-spot rank end-to-end. We propose \textbf{Co}ntinual \textbf{Dy}namic \textbf{R}ank-Selective LoR\textbf{A} (\textbf{CoDyRA}), which jointly trains each LoRA update alongside \colora{rank minimization}, with \colora{sparsity-promoting regularization} on per-component importance weights \citep{sora,adalora}: the differentiable convex relaxation of discrete rank reduction. The \colorb{supervised objective} drives \colorb{plasticity}; \colora{rank minimization} regularizes \colora{forgetting}, keeping each update minimal and the model close to its previous state without task-specific assumptions or stored statistics. Our key reinterpretation: \colora{rank minimization} serves as a \colora{forgetting regularizer} under the CL regime, simultaneously protecting pretrained capability and prior-task knowledge under a single criterion. Across MTIL, X-TAIL, and TRACE (CLIP, LLaMA, Gemma), CoDyRA is competitive or better with prior CL methods on accuracy and forgetting, achieving a strong plasticity--stability balance.

\section{Related Work}

We organize prior work by how each family controls the update applied to the model.

\noindent\textbf{Isolating the model via learning external modules.}
Modular methods \citep{wang2022beef,wang2022foster,l2p,dualprompt,sprompts,zhou2022model,codaprompt,roy2024convolutional,boosting,rail,wang2024self,mcdonnell2024ranpac,zhou2025same,tang2025mind,huang2025mind,liu2025c,zhao2026token} freeze the PTM and route each task through dedicated prompts, adapters, or experts with an inference-time selector, setting PTM update capacity to zero.
\reva{Orthogonal-subspace LoRA variants \citep{wang2023orthogonal,inflora,qiao2024learn,wu2025sdlora,zhu2025bilora,qian2025treelora} constrain per-task updates to fixed-rank subspaces orthogonal to prior tasks and leave the backbone frozen, while CoDyRA regularizes the update's rank via sparsity and stores no per-task subspaces, complementary to orthogonal methods (App.~\ref{supp:ortho}).}

\noindent\textbf{Continually updating the model with past-task information constraints.}
Experience replay and reference-data approaches \citep{li2017learning,rebuffi2017icarl,chaudhry2018riemannian,chaudhry2018efficient,aljundi2019gradient,liu2020mnemonics,der,yan2022learning,luo2023class,zhao2024learning,lu2026little}, including CLIP-specific variants \citep{zscl,garg2023tic,jha2024clapclip,zhang2024overcoming}, constrain each update via a data-grounded loss; parameter-regularization methods (EWC \citep{kirkpatrick2017overcoming}, SI \citep{zenke2017continual}, MAS \citep{aljundi2018memory}, and variants \citep{aljundi2019task,liang2022mind,npcl}) instead soft-bound capacity coordinate-wise via per-task importance. 

\noindent\textbf{Adaptive-rank LoRA.}
Adaptive-rank LoRA mechanisms \citep{adalora,sora,dylora,alora,milora,pissa,moslora,wu2024reft,biderman2024lora,schulman2025lora} focus on parameter-efficient fine-tuning for single downstream tasks, allocating sufficient rank budget for each task.
\reva{These methods neither address continual learning nor connect rank to forgetting; under matched placement, budget, and rank, CoDyRA outperforms them on all CL metrics since our rank selection acts as an implicit forgetting regularizer, whereas AdaLoRA and DyLoRA drive rank selection from a task-learning criterion alone (App.~\ref{supp:adalora}).}

\section{Methodology}
\subsection{Preliminaries}

\noindent\textbf{Transformer backbones.}
We consider PTMs built from stacked transformer blocks, each combining a multi-head attention (MHA) module and a multilayer perceptron (MLP). The attention block at layer $l$ uses Query/Key/Value/Output projections $\mathbf{W}_l^{\mathrm{Q}}, \mathbf{W}_l^{\mathrm{K}}, \mathbf{W}_l^{\mathrm{V}}, \mathbf{W}_l^{\mathrm{O}}\in\mathbb{R}^{d\times d}$; the MLP block uses up- and down-projections $\mathbf{W}_l^{\mathrm{FC}}\in\mathbb{R}^{d\times d_m}, \mathbf{W}_l^{\mathrm{Proj}}\in\mathbb{R}^{d_m\times d}$. Collectively, we denote the trainable backbone weights as $\theta=\{\mathbf{W}_l^{(\cdot)}\}$, and an additive update to $\theta$ as $\Delta\theta$.

\noindent\textbf{LoRA updates.}
Low-Rank Adaptation \citep{lora} writes the update to a weight matrix $\mathbf{W}\in\mathbb{R}^{d\times k}$ as $\Delta\mathbf{W}=\mathbf{B}\mathbf{A}$ with $\mathbf{B}\in\mathbb{R}^{d\times r}, \mathbf{A}\in\mathbb{R}^{r\times k}$, $r\ll\min(d,k)$. After training, $\Delta\mathbf{W}$ merges back into $\mathbf{W}$, so the inference architecture is unchanged. The \emph{effective rank} of $\Delta\mathbf{W}$ is $\mathrm{rank}(\mathbf{B}\mathbf{A})\le r$. This is the quantity we treat as ``capacity'' (Sec.~\ref{sec:intro_capacity}).

\noindent\textbf{Continual learning.}
CoDyRA operates under the general regime of Sec.~\ref{sec:intro_capacity}: at step $t\in\{1,\dots,T\}$, only the current task's dataset $\mathcal{D}^t$ and the current model state $\theta_{t-1}$ are available. After training task $t$, the LoRA update merges into $\theta_{t-1}$ to give $\theta_t$; nothing about previous tasks is retained. Since $\theta_{t-1}$ already encodes \emph{both} the general capability and prior-task knowledge, targeting forgetting with respect to $\theta_{t-1}$ unifies both under a single criterion, unlike replay-based methods (which target only prior-task data) and modular methods (which isolate the pretrained side). Baselines retain their original setups; comparisons are performed at the benchmark level.

\subsection{Update Rank as the Capacity Lever for Plasticity and Stability}

\subsubsection{Empirical Analysis}
\label{sec:motivation}

\begin{figure}[!t]
\centering
\includegraphics[width=0.95\linewidth]{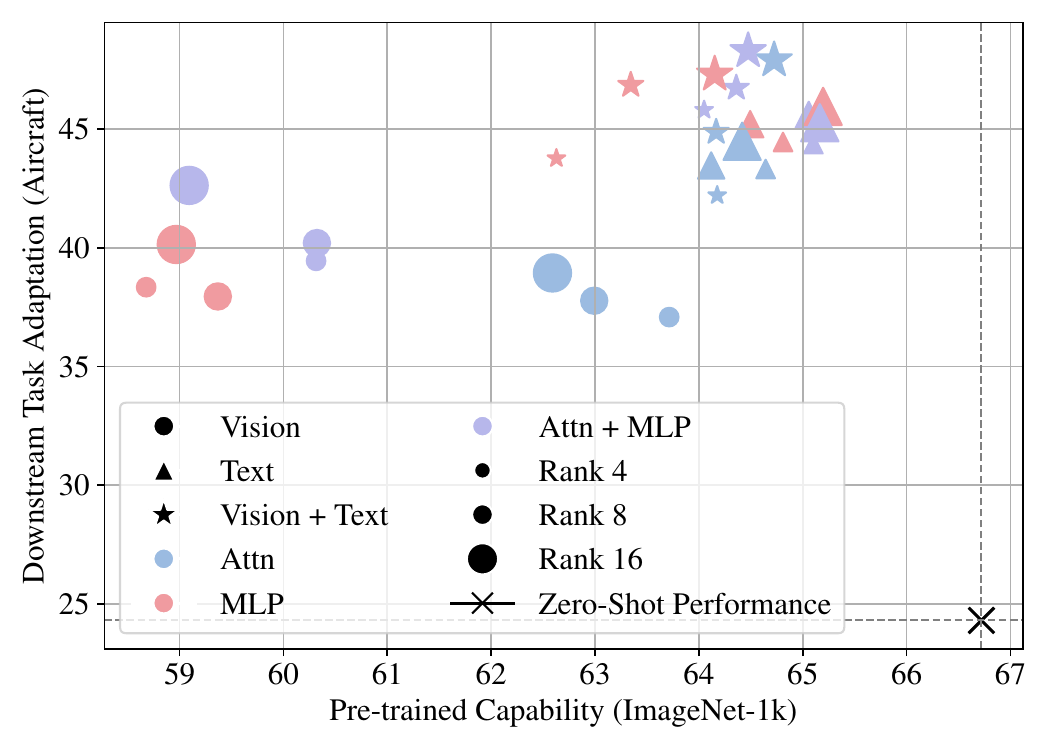}\
\caption{Task adaptation and zero-shot capability retention after training CLIP with different LoRA insertion points and ranks. Shapes indicate encoders, colors denote transformer modules, and sizes reflect rank values.
}
\label{fig:lora_forgetting}
\vspace{0.2cm}
\includegraphics[width=\linewidth]{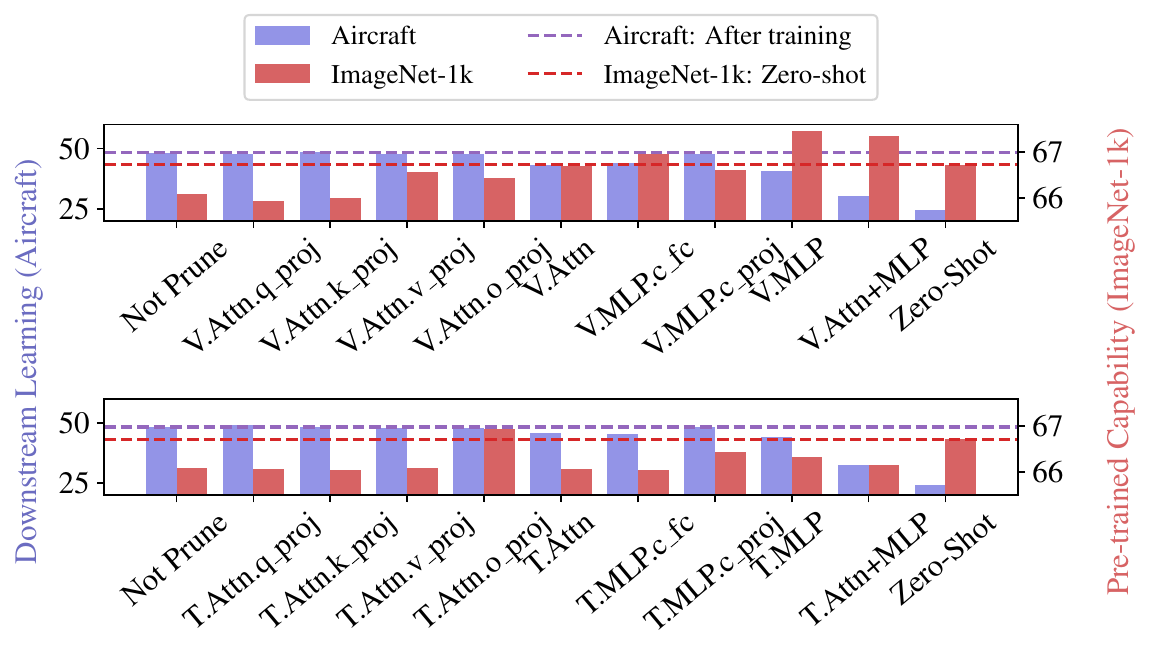}\\
\caption{Impact of removing trained LoRA modules on downstream task adaptation (Aircraft, left y-axis) and retention of existing capabilities (ImageNet-1k, right y-axis). ``T'' and ``V'' indicate text-encoder and vision-encoder modules, respectively.}
\label{fig:prune_motivation}
\end{figure}

We probe how LoRA's rank and placement affect plasticity and stability on a representative CLIP task (Aircraft, with ImageNet-1k as reference) through two complementary analyses \citep{pearl2022direct,meng2022locating}: \textbf{(1)} insert LoRA of varying rank at different locations in pretrained CLIP to measure each component's contribution (Fig.~\ref{fig:lora_forgetting}); \textbf{(2)} apply LoRA to all weights and prune learned components to assess their standalone effect (Fig.~\ref{fig:prune_motivation}). Each setting is evaluated on the new task to measure \colorb{learning of new knowledge}, and on a reference dataset to measure \colora{retention of existing capabilities}.

\noindent\textbf{Applying LoRA on all weights.} 
Unlike prior work \citep{lora,milora,pissa} that only applies LoRA to limited components by default, we first examine the impact of LoRA placements. 
As shown in Fig. \ref{fig:lora_forgetting} and Fig. \ref{fig:prune_motivation}, 
applying LoRA to only specific modules alone may constrain the update capacity (the rank of $\Delta\theta$) in different ways:
\textbf{(1)} LoRA placement strongly impacts downstream performance and retention of pre-trained capabilities; 
\textbf{(2)} Updating only the vision encoder causes more forgetting; 
\textbf{(3)} Updating only attention modules mitigates forgetting but limits learning on new tasks.
These observations reflect the distinct roles of PTM components.
\reva{\takeaway{1}{LoRA placement is a plasticity--stability lever: no single placement or rank simultaneously optimizes both across tasks.}}

\noindent\textbf{Effect of LoRA rank on the learning--forgetting trade-off.}
Rank governs the trade-off directly. As shown in Fig.~\ref{fig:lora_forgetting}, higher-rank LoRA \colorb{facilitates learning} but \colora{increases forgetting}, while lower-rank LoRA \colora{reduces forgetting} but \colorb{limits adaptation}. The extreme is informative: a rank-zero LoRA leaves $\theta$ unchanged and induces no forgetting. A \emph{balance} emerges (top-right of Fig.~\ref{fig:lora_forgetting}) where a moderately small rank reduces forgetting while leaving enough capacity for adaptation. The supervised objective drives \colorb{new-knowledge acquisition} while rank minimization regularizes \colora{forgetting}; jointly optimizing both navigates to this balance. 
\reva{\takeaway{2}{Rank governs the \colorb{plasticity}--\colora{stability} balance: \colorb{higher rank} favors plasticity, \colora{lower rank} favors stability. For a given level of task adaptation, forgetting is minimized at a moderate-rank sweet spot.}}

\noindent\textbf{The balance varies by location and task.} 
Fig. \ref{fig:lora_forgetting} shows that the optimal rank for balance depends on both parameter placement and the learning task, consistent with the pruning-based sufficiency analysis in Fig. \ref{fig:prune_motivation}. For example, removing updates to the Value projection in the Attention module of the vision encoder has little impact on \colorb{adaptation} but substantially restores \colora{pre-trained performance}. Interestingly, pruning the MLP layer slightly reduces \colorb{task performance} while enhancing \colora{pre-trained capabilities}.
\reva{Extended analyses on more data are broadly consistent with these findings across placements and ranks (App.~\ref{supp:multi_dataset}).}
\takeaway{3}{The sweet-spot rank is not universal: its location varies systematically by module and by downstream task.}

\subsubsection{Theoretical Bounds}
\label{sec:theory}

To motivate why rank is the right control knob, we give a forgetting bound that ties Sec.~\ref{sec:motivation}'s empirical sweet spot to a single, controllable quantity. Let $\Delta\mathbf{W} = \mathbf{B}\mathbf{A}$ be a LoRA update with $\mathbf{B}\in\mathbb{R}^{d\times r}$, $\mathbf{A}\in\mathbb{R}^{r\times k}$, and $\rho := \mathrm{rank}(\Delta\mathbf{W}) \le r$ its \emph{effective rank}. Let $\ell(\cdot;\mathcal{D}_{\mathrm{ref}})$ be the loss on a reference distribution and $F(\theta_t) := \ell(\theta_t;\mathcal{D}_{\mathrm{ref}})-\ell(\theta_{t-1};\mathcal{D}_{\mathrm{ref}})$ the resulting forgetting. Proof in Appendix~\ref{supp:theory}.

\begin{proposition}[Rank-dependent forgetting bound]
\label{prop:forget}
If $\ell(\cdot;\mathcal{D}_{\mathrm{ref}})$ is $L$-smooth on a convex neighborhood of $\theta_{t-1}$ containing $\theta_t$, then
\begin{equation}
\label{eq:forget_bound}
F(\theta_t) \;\le\; \|\nabla\ell(\theta_{t-1})\|_F\,\|\Delta\mathbf{W}\|_F \,+\, \tfrac{L}{2}\,\|\Delta\mathbf{W}\|_F^2,
\end{equation}
with $\|\Delta\mathbf{W}\|_F \le \sqrt{\rho}\,\|\mathbf{B}\|_2\|\mathbf{A}\|_2$. Moreover, $\Delta\mathbf{W}$ has row and column space of dimension at most $\rho$.
\end{proposition}

Both terms in the upper bound shrink as $\rho$ decreases\reva{; the operator norms $\|\mathbf{B}\|_2, \|\mathbf{A}\|_2$ remain near-constant across ranks under weight decay on $\mathbf{A}, \mathbf{B}$ (App.~\ref{supp:norm_stability})}. While the subspace property follows from the rank constraint by construction, its practical implication is that interference is confined to at most $\rho$ input and output directions. Reducing $\rho$ thus tightens the bound along two axes, \emph{magnitude} and \emph{subspace dimension}, under a criterion that is agnostic to $\mathcal{D}_{\mathrm{ref}}$ and hence covers general capability and prior-task knowledge alike. Sec.~\ref{sec:our_method} controls $\rho$ through the $\ell_1$ relaxation of $\ell_0$ \citep{tibshirani1996regression}.
\reva{A finer decomposition (App.~\ref{supp:two_mechanisms}) attributes forgetting to an interference (first-order) mechanism and a magnitude (second-order) mechanism, with CoDyRA operating on the latter.}

\begin{figure*}[!t]
\centering
\includegraphics[width=1\linewidth]{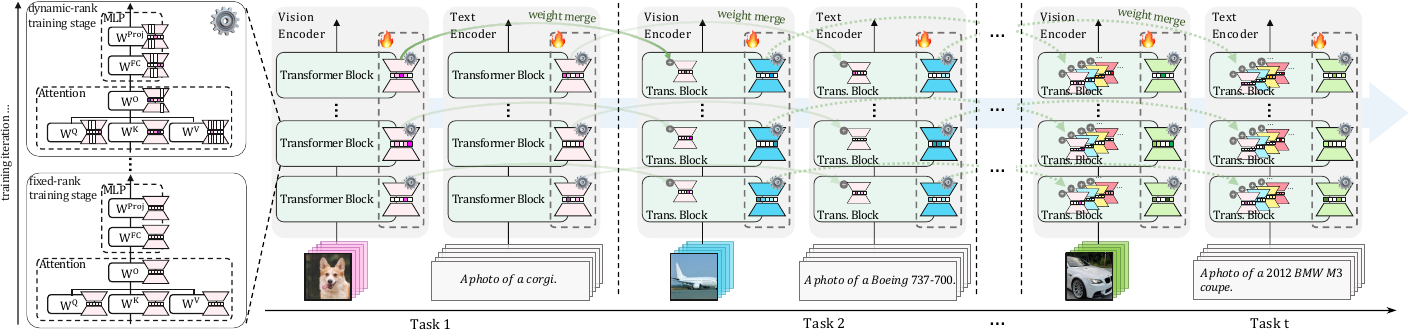}\
\caption{Overview of \ours: dynamic rank-selective LoRA enables each pre-trained weight matrix to adaptively retain only the necessary ranks for downstream adaptation while preserving pre-trained capabilities. After each task, updates merge into the backbone (as with standard LoRA), incurring no inference overhead.
}
\label{fig:main}
\end{figure*}

\subsection{CoDyRA: Implicit Forgetting Regularization with Rank Minimization}
\label{sec:our_method}
Takeaways~1--3 together specify the design: placement matters (Takeaway~1); rank governs the plasticity--stability balance with no universally optimal fixed choice (Takeaway~2); and the sweet-spot rank varies by location and task (Takeaway~3). We therefore apply LoRA across \emph{all} weight matrices and let the per-location rank be selected adaptively. Adaptive-rank LoRA mechanisms were developed for parameter efficiency in single-task fine-tuning \citep{adalora,dylora,sora,alora}; in the CL context, the same $\ell_1$ penalty on per-component importance weights serves a fundamentally different purpose. Because rank directly governs forgetting (Prop.~\ref{prop:forget}), \colora{rank minimization} acts as a \colora{forgetting regularizer} rather than a compression tool, while the \colorb{supervised objective} drives \colorb{plasticity}. Concretely, CoDyRA inserts the rank-selective updates in the Attention and MLP modules of both encoders (Fig.~\ref{fig:main}) and merges them back into the backbone after each task.

\noindent\textbf{Regularized low-rank updates with rank importance.}
Let $\{\mathbf{W}^{t,m}_0\}_{m=1}^{M}$ denote the $M$ weight matrices updated by LoRA at task $t$ (each carrying any prior-task updates). For each matrix, LoRA \citep{lora} introduces $\mathbf{B}^{t,m}\in\mathbb{R}^{d\times r}$ and $\mathbf{A}^{t,m}\in\mathbb{R}^{r\times k}$ with $r\ll\min(d,k)$:
\begin{equation}
\label{eq:lora}
\mathbf{W}^{t,m} = \mathbf{W}_0^{t,m} + \mathbf{B}^{t,m}\mathbf{A}^{t,m}.
\end{equation}

The forgetting upper bound (Sec.~\ref{sec:theory}) tightens as the effective rank of $\Delta\mathbf{W}^{t,m}=\mathbf{B}^{t,m}\mathbf{A}^{t,m}$ decreases, so reducing the active rank is a principled lever for reducing forgetting. We capture this with a regularizer $R(\Delta\mathbf{W}^{t,m})$ that vanishes when the update has rank zero (no change to $\mathbf{W}^{t,m}$, no forgetting) and grows with the active rank. Jointly optimizing $R$ alongside the \colorb{supervised objective} balances \colorb{plasticity} and \colora{stability} (Takeaway~2).

To automatically and adaptively minimize the active ranks in LoRA, we formulate $R(\Delta \mathbf{W}^{t,m})$ through a rank-selective LoRA. 
We introduce a learnable importance weight vector $\mathbf{w}^{t,m} \in \mathbb{R}^r$ to indicate the active ranks for a LoRA adapter:
\begin{equation}
\label{eq:dyra}
\Delta \mathbf{W}^{t,m} = \sum\nolimits_{i=1}^{r} \mathbf{w}_i^{t,m} \mathbf{B}_{:,i}^{t,m} \mathbf{A}_{i,:}^{t,m}
\end{equation}
where $\mathbf{w}_i^{t,m}$ is the learnable importance weight associated with rank $i$, $\mathbf{B}_{:,i}^{t,m}$ denotes the $i$-th column of $\mathbf{B}^{t,m}$, and $\mathbf{A}_{i,:}^{t,m}$ denotes the $i$-th row of $\mathbf{A}^{t,m}$. By construction, the effective rank satisfies $\rho \le \|\mathbf{w}^{t,m}\|_0$, so $\|\mathbf{w}^{t,m}\|_1$ (the convex envelope of $\|\cdot\|_0$ on $[-1,1]^r$ \citep{tibshirani1996regression}) relaxes the rank-dependent forgetting bound, turning $R(\Delta\mathbf{W}^{t,m})$ into a tractable forgetting surrogate.

\begin{table*}[!t]
\centering
    \resizebox{0.95\linewidth}{!}{
	\renewcommand{\arraystretch}{1.05}
	\begin{tabular}{l>{\centering\arraybackslash}p{1cm} >{\centering\arraybackslash}p{1cm}>{\centering\arraybackslash}p{1cm} >{\centering\arraybackslash}p{1cm} >{\centering\arraybackslash}p{1cm} >{\centering\arraybackslash}p{1cm} >{\centering\arraybackslash}p{1cm} >{\centering\arraybackslash}p{1cm} >{\centering\arraybackslash}p{1cm} >{\centering\arraybackslash}p{1cm} >{\centering\arraybackslash}p{1cm} >{\centering\arraybackslash}p{1.5cm}}
		\toprule
		{\quad}\textbf{Method} & \makecell[c]{\rotatebox{90}{Aircraft}} & \makecell[c]{\rotatebox{90}{Caltech101}} & \makecell[c]{\rotatebox{90}{CIFAR100}} & \makecell[c]{\rotatebox{90}{DTD}} & \makecell[c]{\rotatebox{90}{EuroSAT}} & \makecell[c]{\rotatebox{90}{Flowers}} & \makecell[c]{\rotatebox{90}{Food}} & \makecell[c]{\rotatebox{90}{MNIST}} & \makecell[c]{\rotatebox{90}{OxfordPet}} & \makecell[c]{\rotatebox{90}{Cars}} & \makecell[c]{\rotatebox{90}{SUN397}} & \makecell[c]{\textbf{\textit{Average}}} \\
		\midrule
		\rowcolor{Green!10}\multicolumn{13}{l}{\emph{CLIP zero-shot reference}}\\
		{\quad}Zero-shot \citep{clip} & 24.3 & 88.4 & 68.2 & 44.6 & 54.9 & 71.0 & 88.5 & 59.4 & 89.0 & 64.7 & 65.2 & 65.3  \\
		\midrule
		\rowcolor{Green!10}\multicolumn{13}{l}{\emph{Transfer}}\\
		{\quad}Zero-shot \citep{clip} & -- & \highblue{88.4}&\highblue{68.2}& 44.6 &\highred{54.9}&\highred{71.0}& \highblue{88.5} & 59.4 & \highblue{89.0} &\highred{64.7}&\highred{65.2}& \highblue{69.4}\\
		\multicolumn{13}{l}{\colorbox{FigComp}{\textbf{\textit{\small Model complementation}}}} \\
		{\quad}MoE-Adapter$\dagger$ \citep{boosting} & -- & 87.9 & \highblue{68.2} & 44.1 & 48.1 & 64.7 & \highred{88.8} & \highred{69.0} & \highred{89.1} & \highblue{64.5} & \highblue{65.1} & 68.9\\
		{\quad}RAIL-Primal$\dagger$ \citep{rail} & -- & \highblue{88.4}&\highblue{68.2}& 44.6 &\highred{54.9}&\highred{71.0}& \highblue{88.5} & 59.4 & \highblue{89.0} &\highred{64.7}&\highred{65.2}& \highblue{69.4}\\
		\multicolumn{13}{l}{\colorbox{FigCont}{\textbf{\textit{\small Model continually learned}}}} \\
		\multicolumn{13}{l}{\hspace{0.5em}\colorbox{FigRef}{\textit{\small reference-constrained}}} \\
		{\quad}LwF \citep{li2017learning} & -- & 72.1 & 49.2 & 35.9 & 44.5 & 41.1 & 66.6 & 50.5 & 69.0 & 19.0 & 51.7 & 50.0   \\
		{\quad}LwF-VR \citep{ding2022don} & -- & 82.2 & 62.5 & 40.1 & 40.1 & 56.3 & 80.0 & 60.9 & 77.6 & 40.5 & 60.8 & 60.1   \\
		{\quad}WiSE-FT \citep{wortsman2022robust} &--  & 77.6 & 60.0 & 41.3 & 39.4 & 53.0 & 76.6 & 58.1 & 75.5 & 37.3 & 58.2 & 57.7   \\
		{\quad}ZSCL \citep{zscl} & -- & 84.0 & 68.1 & \highblue{44.8} & 46.8 & 63.6 & 84.9 & 61.4 & 81.4 & 55.5 & 62.2 & 65.3\\
		\multicolumn{13}{l}{\hspace{0.5em}\colorbox{FigCap}{\textit{\small capacity-controlled (ours)}}} \\
		{\quad}\textbf{\ours} & -- & \highred{92.4} & \highred{68.4} & \highred{45.8} & \highblue{54.5} & \highblue{69.6} & 87.4 & \highblue{65.2} & 88.5 & 64.2 & 64.5 & \highred{70.1}\\
		\midrule
		\rowcolor{Green!10}\multicolumn{13}{l}{\emph{Average}}\\
		\multicolumn{13}{l}{\colorbox{FigComp}{\textbf{\textit{\small Model complementation}}}} \\
		{\quad}MoE-Adapter$\dagger$ \citep{boosting}& 30.0 & 89.6 & \highblue{73.9}& 58.7& 69.3 &79.3 & \highblue{88.1}& \highred{76.5} & 89.1 & 65.3 &\highred{65.8} & 71.4 \\
		{\quad}RAIL-Primal$\dagger$ \citep{rail}& 32.9 &94.5 &69.9 &58.1& 71.8 &\highred{84.4} &\highred{88.5} &70.4 &89.0 &\highblue{66.1} & \highblue{65.7} & 71.9\\
		\multicolumn{13}{l}{\colorbox{FigCont}{\textbf{\textit{\small Model continually learned}}}} \\
		\multicolumn{13}{l}{\hspace{0.5em}\colorbox{FigRef}{\textit{\small reference-constrained}}} \\
		{\quad}LwF \citep{li2017learning} &23.5 &77.4 &43.5 &41.7 &43.5 &52.2 &54.6 & 63.4 & 68.0 & 21.3 & 52.6 & 49.2\\
		{\quad}LwF-VR \citep{ding2022don} &24.9 & 89.1 &64.2 &53.4 &54.3 &70.8 &79.2 &66.5 &79.2 & 44.1 & 61.6 & 62.5\\
		{\quad}WiSE-FT \citep{wortsman2022robust} & 32.0 & 87.7 & 61.0 & 55.8 & 68.1 & 69.3 & 76.8 & 71.5 & 77.6 & 42.0 & 59.3 & 63.7  \\
		{\quad}ZSCL \citep{zscl} & 28.2 & 88.6 & 66.5 & 53.5 & 56.3 & 73.4 & 83.1 & 56.4 & 82.4 & 57.5 & 62.9 & 64.4 \\
		\multicolumn{13}{l}{\hspace{0.5em}\colorbox{FigCap}{\textit{\small capacity-controlled (ours)}}} \\
		{\quad}\textbf{\ours} & \highred{34.6} & \highred{95.8} & \highred{73.9} & \highred{60.0} & \highred{77.1} & \highblue{81.3} & 86.6 & \highblue{75.9} & \highred{89.9} & \highred{66.1} & 65.3 & \highred{73.3}\\
		\midrule
		\rowcolor{Green!10}\multicolumn{13}{l}{\emph{Last}}\\
		\multicolumn{13}{l}{\colorbox{FigComp}{\textbf{\textit{\small Model complementation}}}} \\
		{\quad}MoE-Adapter$\dagger$ \citep{boosting} & 30.1 & 89.3 & \highred{74.9} & \highred{64.0} & \highblue{82.3} & 89.4 & \highblue{87.1} & 89.0 & \highblue{89.1} & 69.5 & \highblue{72.5} & 76.1 \\
		{\quad}RAIL-Primal$\dagger$ \citep{rail} & \highred{32.9} & \highblue{95.1} & 70.3 & 63.2 & 81.5 & \highred{95.6} & \highred{88.5} & 89.7 & 89.0 & \highblue{72.5} & 71.0 & \highblue{77.2}\\
		\multicolumn{13}{l}{\colorbox{FigCont}{\textbf{\textit{\small Model continually learned}}}} \\
		\multicolumn{13}{l}{\hspace{0.5em}\colorbox{FigRef}{\textit{\small reference-constrained}}} \\
		{\quad}LwF \citep{li2017learning} & 22.1 & 58.2 & 17.9 & 32.1 & 28.1 & 66.7 & 46.0 & 84.3 & 64.1 & 31.5 & 60.1 & 46.5 \\
		{\quad}LwF-VR \citep{ding2022don} & 22.9 & 89.8 & 59.3 & 57.1 & 57.6 & 79.2 & 78.3 & 77.7 & 83.6 & 60.1 & 69.8 & 66.9  \\
		{\quad}WiSE-FT \citep{wortsman2022robust} & 30.8 & 88.9 & 59.6 & 60.3 & 80.9 & 81.7 & 77.1 & \highred{94.9} & 83.2 & 62.8 & 70.0 & 71.9 \\
		{\quad}ZSCL \citep{zscl} & 26.8 & 88.5 & 63.7 & 55.7 & 60.2 & 82.1 & 82.6 & 58.6 & 85.9 & 66.7 & 70.4 & 67.4 \\
		\multicolumn{13}{l}{\hspace{0.5em}\colorbox{FigCap}{\textit{\small capacity-controlled (ours)}}} \\
		{\quad}\textbf{\ours} & \highblue{31.6} & \highred{95.5} & \highblue{72.8} & \highblue{63.5} & \highred{85.0} & \highblue{89.7} & 85.0 & \highblue{94.7} & \highred{93.2} & \highred{73.6} & \highred{73.0} & \highred{78.0}\\
		\bottomrule
	\end{tabular}}
    \vspace{-0.1cm}
    \caption{5-shot MTIL. CoDyRA trains a capacity-bounded LoRA on the current task and merges it into the backbone, storing no past-task data or parameters. \highred{Best}/\highblue{second best} in \highred{red}/\highblue{blue}. $\dagger$ indicates domain or distribution detection \citep{boosting, rail}. More detailed results in Appendix Table~\ref{tab:few_xtail} (X-TAIL) and~\ref{supp:full_mtil} (full-shot MTIL).}\label{tab:few_mtil}
\end{table*}

\noindent\textbf{Capacity-controlled updates with sparsity-promoting regularization.}
The formulation in Eq.~\ref{eq:dyra} enables the model to dynamically optimize the importance of each rank and minimize active ranks through gradient descent.
To only {retain the essential ranks} for {learning new tasks} (thus reducing forgetting), we can achieve optimizing \colora{$R(\Delta \mathbf{W}^{t,m})$} through an $\ell_1$ norm-based sparsity-promoting regularization \citep{tibshirani1996regression,sora} on $\mathbf{w}^{t,m}$, \ie, \colora{$\|\mathbf{w}^{t,m}\|_1$}. 
The optimization objective for parameters $\{\mathbf{w}^{t,m},\mathbf{B}^{t,m},\mathbf{A}^{t,m}\}_{m=1}^M$ learned on task $t$ is:
\begin{equation}
\mathcal{L}_{\text{train}}^{t} := \colorb{\mathcal{L}_{\text{sup}}^t} + \colora{\lambda \sum\nolimits_{m=1}^{M} \|\mathbf{w}^{t,m}\|_1},
\label{eq:loss}
\end{equation}
where $\mathcal{L}_{\text{sup}}^t$ is the supervised training loss for the current task, $M$ is the number of weight matrices with an inserted $r$-rank update, and $\mathbf{w}^{t,m}$ represents the importance weights of each rank added to the weight matrix $\mathbf{W}^{t,m}_0$. The $\ell_1$ regularization strength is controlled by $\lambda$.
By Eq.~\ref{eq:dyra}, driving $\|\mathbf{w}^{t,m}\|_1$ toward zero shrinks both the active rank and $\|\Delta\mathbf{W}^{t,m}\|_F$, tightening the forgetting bound. Further discussion of how rank minimization connects to forgetting reduction is in Appendix~\ref{supp:ortho}. \reva{The $\ell_1$ penalty thus serves as an implicit forgetting regularizer: minimizing $\|\mathbf{w}^{t,m}\|_1$ tightens the bound of Prop.~\ref{prop:forget} and controls forgetting against the current model state without any past-task information.}

To handle the non-differentiable $\ell_1$ regularization applied to the importance weights, we adopt the proximal gradient method~\citep{beck2009fast}.
The $i$-th element of $\hat{\mathbf{w}}^{t,m}$ for rank $i$ is updated via soft-thresholding:
\begin{equation}
    \mathbf{w}_i^{t,m} := \mathbbm{1}(\lvert \hat{\mathbf{w}}_i^{t,m} \rvert > \kappa) \cdot (\hat{\mathbf{w}}_i^{t,m} {-} \mathrm{sign}(\hat{\mathbf{w}}_i^{t,m}) \cdot \kappa),
    \label{eq:sparse_update}
\end{equation}
where $\hat{\mathbf{w}}_i^{t,m}$ denotes the value of $\mathbf{w}_i^{t,m}$ after applying the gradient update from the supervised loss $\mathcal{L}_\text{sup}^t$. 
The threshold $\kappa$ increases from zero to $\kappa_{\text{max}}$, analogous to linearly scheduling $\lambda$, and adaptively induces sparsity in $\mathbf{w}$ based on the supervised objective. 
The indicator $\mathbbm{1}(\cdot)$ returns 1 if the condition holds and 0 otherwise, while $\mathrm{sign}(\cdot)$ gives the input's sign ($\pm$). The full derivation is in Appendix~\ref{supp:derivation}.

This approach preserves only the ranks important for the current task and prunes those with low relevance, adapting across weights and tasks. The importance weights on preserved ranks softly highlight their relative significance. 
Early in training, dense updates are applied without Eq.~\ref{eq:sparse_update}, allowing all ranks to capture task-relevant information before sparsity is enforced. Following standard conventions, $\mathbf{A}^{t,m}$ is initialized randomly and $\mathbf{B}^{t,m}$ to zero; $\mathbf{w}^{t,m}$ is initialized randomly.

\begin{table*}[!t]
    \centering
    \resizebox{\linewidth}{!}{
    \small
\begin{tabular}{@{}l c ccccc ccc c c@{}}
\toprule
 & \multicolumn{1}{c}{} & \multicolumn{5}{c}{\colorbox{FigComp}{\textbf{\textit{Model complementation}}}} & \multicolumn{5}{c}{\colorbox{FigCont}{\textbf{\textit{Model continually learned}}}} \\
\cmidrule(lr){3-7} \cmidrule(lr){8-12}
 & \multicolumn{1}{c}{\colorbox{FigCompNone}{\textit{no training}}} & \multicolumn{5}{c}{\colorbox{FigCompAux}{\textit{auxiliary modules}}} & \multicolumn{3}{c}{\colorbox{FigRef}{\textit{reference-constrained}}} & \colorbox{FigNaive}{\textit{naive LoRA}} & \colorbox{FigCap}{\textit{capacity-controlled}} \\
\cmidrule(lr){2-2} \cmidrule(lr){3-7} \cmidrule(lr){8-10} \cmidrule(lr){11-11} \cmidrule(lr){12-12}
 & FIX(ICL) & L2P & DualPrompt & HiDeLoRA & O-LoRA & TreeLoRA & OGD & EWC & GEM & SeqLoRA & \textbf{CoDyRA} \\ \midrule
\rowcolor{Green!10}\multicolumn{12}{l}{meta-llama / LLaMA-2-7B-Chat} \\ \midrule
OP & 38.94 $\pm$ 0.3 & 36.23 $\pm$ 0.8 & 37.69 $\pm$ 1.2 & 41.60 $\pm$ 0.8 & 42.78 $\pm$ 0.8 & 43.52 $\pm$ 1.0 & 42.09 $\pm$ 1.6 & 42.36 $\pm$ 1.2 & 40.08 $\pm$ 1.6 & 34.30 $\pm$ 1.2 & \textbf{43.82 $\pm$ 0.9} \\
\reva{Forgetting} & -- & 8.25 $\pm$ 0.8 & 8.03 $\pm$ 0.8 & 7.12 $\pm$ 0.4 & 7.16 $\pm$ 0.4 & 3.46 $\pm$ 0.4 & 8.06 $\pm$ 1.2 & 5.97 $\pm$ 0.8 & 6.77 $\pm$ 1.2 & 18.50 $\pm$ 0.8 & \textbf{3.25 $\pm$ 0.5} \\ \midrule
\rowcolor{Green!10}\multicolumn{12}{l}{google / Gemma-2B-it} \\ \midrule
OP & 32.3 $\pm$ 0.2 & 31.14 $\pm$ 1.2 & 32.42 $\pm$ 1.0 & 33.25 $\pm$ 0.9 & 33.73 $\pm$ 0.8 & 33.41 $\pm$ 0.9 & 32.85 $\pm$ 1.4 & 28.35 $\pm$ 1.6 & 26.48 $\pm$ 1.5 & 31.89 $\pm$ 0.8 & \textbf{33.96 $\pm$ 1.0} \\
\reva{Forgetting} & -- & 15.77 $\pm$ 0.7 & 14.25 $\pm$ 0.5 & 13.66 $\pm$ 0.5 & 12.36 $\pm$ 0.4 & 8.50 $\pm$ 0.5 & 12.27 $\pm$ 0.9 & 16.96 $\pm$ 1.2 & 18.25 $\pm$ 0.9 & 15.28 $\pm$ 0.4 & \textbf{7.94 $\pm$ 0.8} \\ \midrule
\rowcolor{Green!10}\multicolumn{12}{l}{meta-llama / LLaMA-3-1B-Instruct} \\ \midrule
OP & 31.16 $\pm$ 0.4 & 29.38 $\pm$ 1.2 & 30.76 $\pm$ 1.2 & 33.73 $\pm$ 1.2 & 32.94 $\pm$ 0.8 & 36.14 $\pm$ 0.7 & 30.12 $\pm$ 2.0 & 31.96 $\pm$ 1.6 & 32.19 $\pm$ 2.0 & 29.73 $\pm$ 1.6 & \textbf{37.46 $\pm$ 0.8} \\
\reva{Forgetting} & -- & 13.57 $\pm$ 0.8 & 11.34 $\pm$ 0.8 & 12.36 $\pm$ 0.8 & 12.89 $\pm$ 1.2 & 7.36 $\pm$ 0.8 & 15.2 $\pm$ 1.6 & 11.62 $\pm$ 1.2 & 10.74 $\pm$ 1.6 & 17.03 $\pm$ 1.2 & \textbf{5.11 $\pm$ 0.8} \\ \bottomrule
\end{tabular}
    }
    \vspace{-0.2cm}
    \caption{TRACE LLM-CL results. We report Overall Performance (OP\,\%; higher is better) and \reva{Forgetting (\%; lower is better)}; averaged over 3 runs $\pm$ std; best in bold.}
    \label{tab:llm_cl}
\end{table*}

\section{Experiments}
\reva{We evaluate on CLIP \citep{clip,openclip} with multi-domain task-incremental learning (MTIL \citep{zscl,boosting}) and cross-domain task-agnostic incremental learning (X-TAIL \citep{rail}), and on LLaMA \citep{touvron2023llama,dubey2024llama} and Gemma \citep{team2024gemma} with TRACE \citep{wang2023trace}. On CLIP CL (MTIL, X-TAIL), we report the Transfer, Average, and Last metrics of \citet{zscl}; on TRACE (LLM CL), Overall Performance (OP) of \citet{wang2023trace}. Both settings report a Forgetting metric (lower is better); formal definitions of all metrics are in Appendix~\ref{supp:metrics}. Full settings and extended results are in Appendix~\ref{supp:experiment}.}

\subsection{Experimental Results}
\label{sec:main_exp}

\noindent\textbf{CLIP continual learning.}
On MTIL (Table~\ref{tab:few_mtil}), CoDyRA is competitive or better across all metrics without replay, inference-time task IDs, or architectural change. \textit{Transfer} measures retention of pre-trained (zero-shot) capability on yet-to-be-learned tasks; \textit{Last} measures end-of-stream task accuracy, combining new-task learning with retention of prior tasks. On Transfer, reference-constrained baselines (LwF, ZSCL, WiSE-FT) drift below the zero-shot reference (forgetting), model complementation methods (MoE-Adapter, RAIL) are upper-bounded by it (RAIL-Primal ties the zero-shot ceiling at 69.4 by construction), and CoDyRA is the only method that strictly surpasses it (Transfer 70.1); CoDyRA amortizes new knowledge into the backbone rather than erasing or merely preserving general capability. CoDyRA also leads on Last (78.0), indicating strong learning with low forgetting on prior tasks. All updates merge into the backbone, and CoDyRA uses the fewest trainable parameters among baselines with directly comparable training procedures (Table~\ref{tab:comp_cost}). 
We also include comparisons on related adaptive rank \citep{adalora,dylora} methods (Appendix~\ref{supp:adalora}).

\begin{table}[!tb]
\centering\small
\resizebox{0.95\linewidth}{!}{\begin{tabular}{@{}lcc@{}}
\toprule
Method & Train.\ Params. & Add.\ Inf.\ Params.\,/\,Mem. \\ \midrule
LwF \citep{li2017learning}        & 129.6M & None \\
ZSCL \citep{zscl}                 & 129.6M & None \\
MoE-Adapters \citep{boosting}     & 59.8M  & 13.35M \\
RAIL \citep{rail}                 & N/A    & 24.18M\,/\,9.01M \\
\midrule
\rowcolor{Purple!8}
\textbf{\ours}                    & \textbf{4.4M} & \textbf{None} \\ \bottomrule
\end{tabular}}
\vspace{-0.2cm}
\caption{Computation cost on CLIP CL. CoDyRA's updates merge into the model, adding no inference overhead.}
\label{tab:comp_cost}
\end{table}
\begin{table}[!tb]
\centering\small
\resizebox{\linewidth}{!}{\begin{tabular}{@{}lcccc>{\columncolor{Purple!8}}c@{}}
\toprule
Method & ZSCL & MoE-Ad. & InfLoRA &  LoRA ($r{=}16$) & \textbf{CoDyRA} \\ \midrule
\reva{Forgetting} & 8.78 & 5.43 & 3.12  & 14.86 & \textbf{1.87} \\ \bottomrule
\end{tabular}}
\vspace{-0.2cm}
\caption{\reva{Forgetting} (\%; lower is better) on CLIP CL.}
\label{tab:bwt_main}
\vspace{-0.3cm}
\end{table}

\noindent\textbf{LLM continual learning.}
On TRACE (Table~\ref{tab:llm_cl}), regularization (EWC \citep{kirkpatrick2017overcoming}, OGD \citep{farajtabar2020orthogonal}), rehearsal (GEM \citep{lopez2017gradient}), and prompt methods (L2P \citep{l2p}, DualPrompt \citep{dualprompt}) often underperform In-Context Learning (FIX(ICL); \citealp[ICL,][]{brown2020language}), and naive sequential LoRA fine-tuning (SeqLoRA) forgets severely. O-LoRA \citep{wang2023orthogonal}, HiDeLoRA \citep{wang2025hide}, and TreeLoRA \citep{qian2025treelora} improve stability via orthogonality, hierarchical decomposition, or tree expansion. CoDyRA performs comparably or better with capacity-controlled updates alone.

\subsection{Discussions}
\label{sec:discussion}

\reva{\noindent\textbf{Capacity control reduces forgetting.}
CoDyRA reduces Forgetting on the CLIP CL stream to $1.87\%$, the lowest among replay-based (ZSCL), modular (MoE-Adapter), orthogonal-subspace (InfLoRA), and fixed-rank LoRA baselines (Table~\ref{tab:bwt_main}). This aligns with Prop.~\ref{prop:forget}: rank minimization tightens the bound; the gap to fixed-rank LoRA isolates \colora{rank minimization}'s contribution. The same reduction holds on TRACE (Table~\ref{tab:llm_cl}).}

\begin{figure*}[!t]
    \begin{minipage}[t]{0.48\textwidth}
        \centering
        \includegraphics[width=0.95\textwidth]{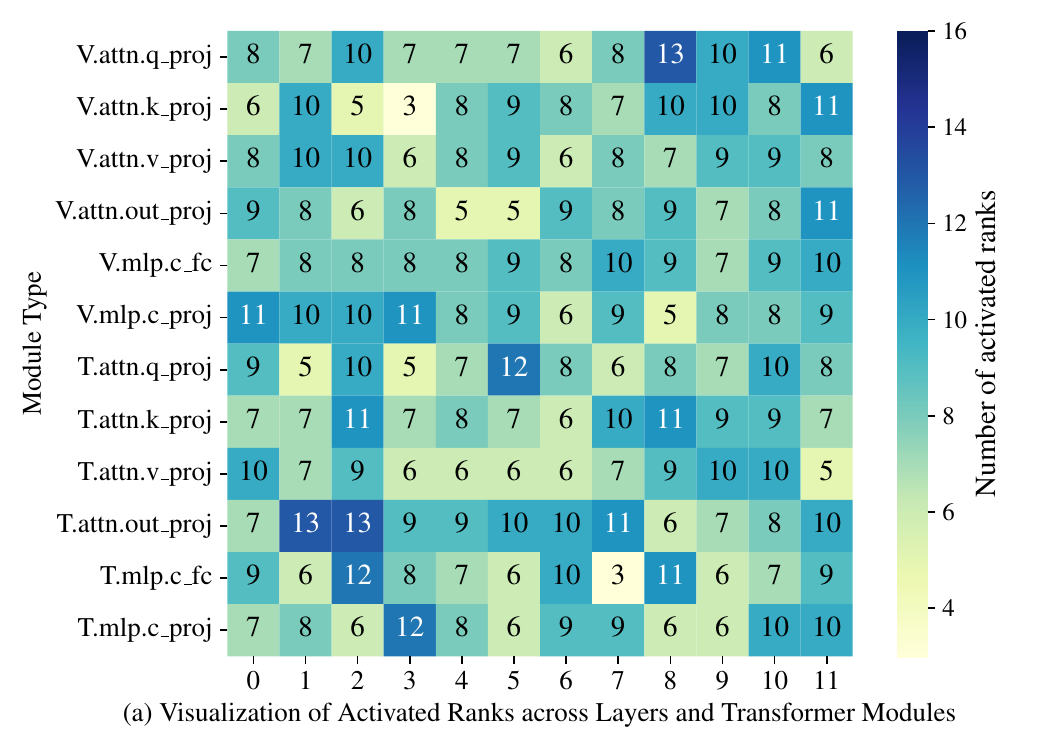}\\
        \includegraphics[width=0.95\textwidth]{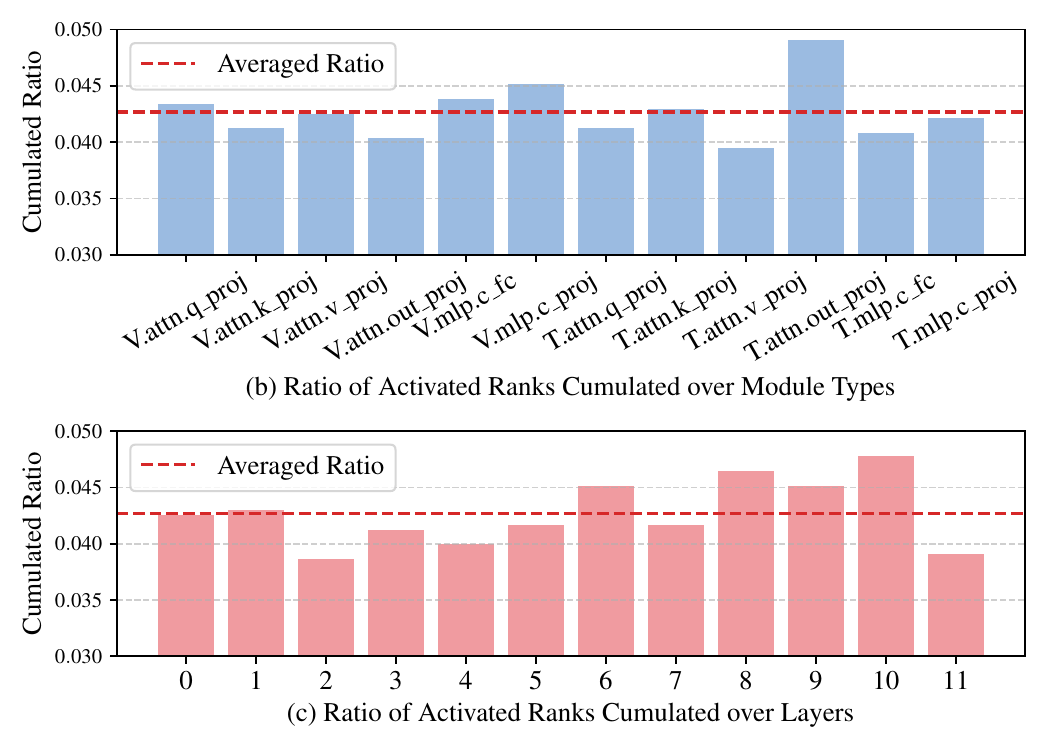}\\
        \vspace{-0.2cm}
        \caption{Visualization and statistical analysis of rank activation on the Aircraft dataset.}
        \label{fig:aircraft_rank_stat}
    \end{minipage}
    \hfill
    \begin{minipage}[t]{0.48\textwidth}
        \centering
        \includegraphics[width=0.95\textwidth]{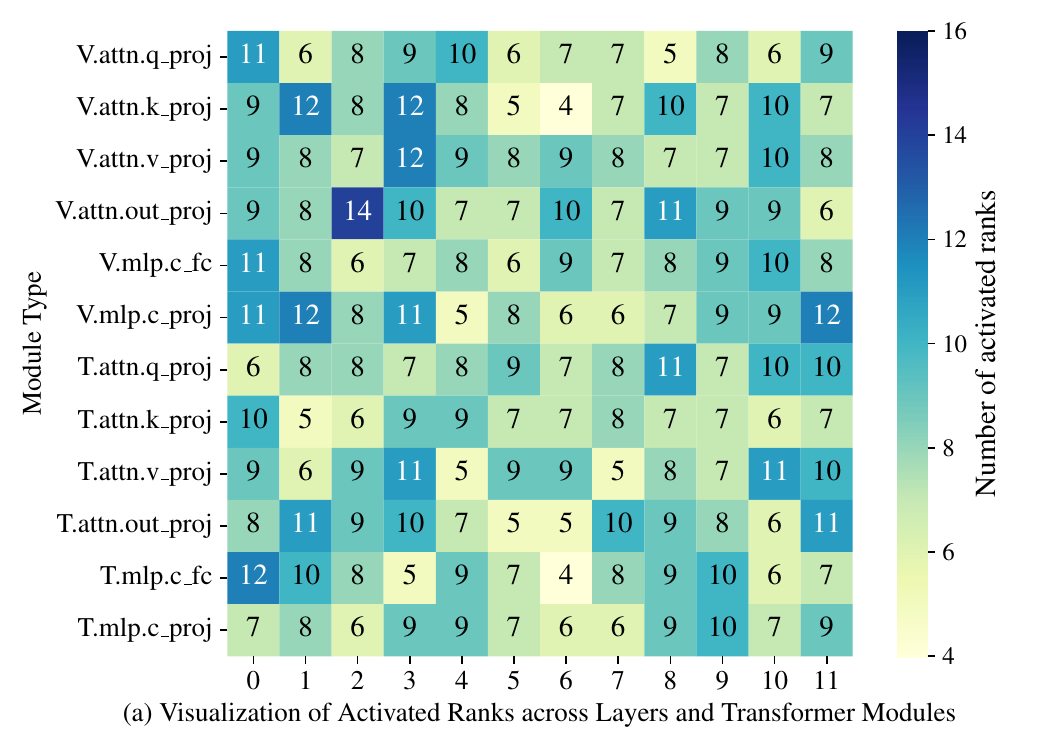}\\
        \includegraphics[width=0.95\textwidth]{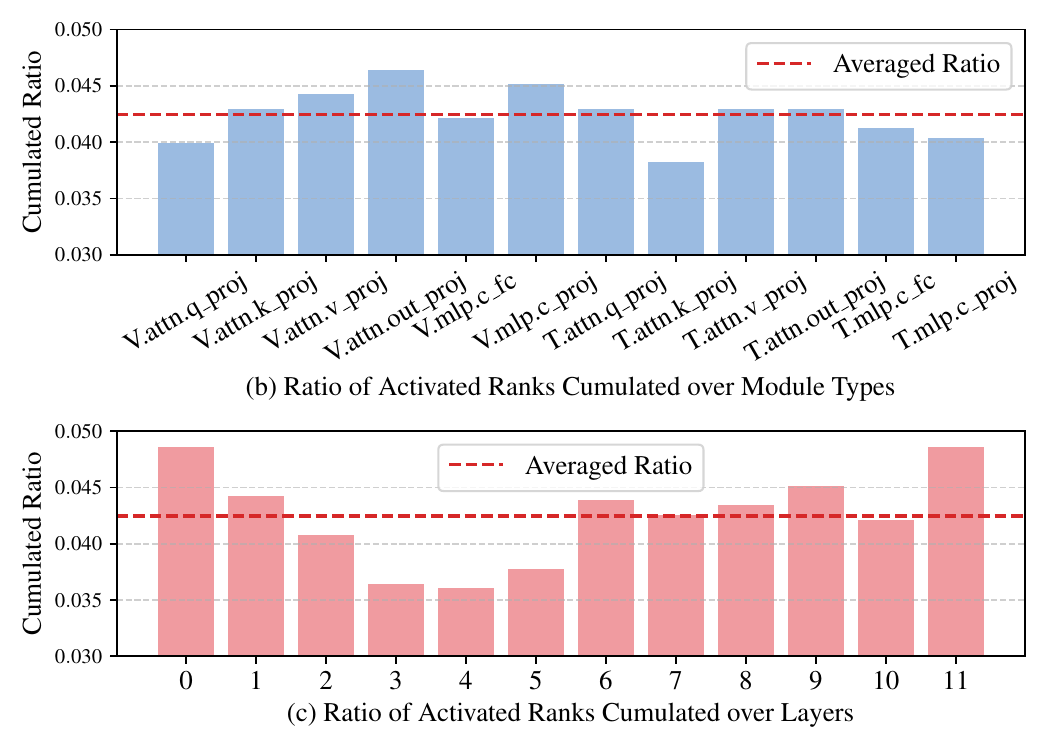}\\
        \vspace{-0.2cm}
        \caption{Visualization and statistical analysis of rank activation on the Oxford Pets dataset.}
        \label{fig:pets_rank_stat}
    \end{minipage}
\end{figure*}

\begin{figure*}[!tb]
\vspace{-0cm}
  \centering
  \begin{minipage}[t]{0.49\linewidth}
    \centering
    \includegraphics[width=0.45\linewidth]{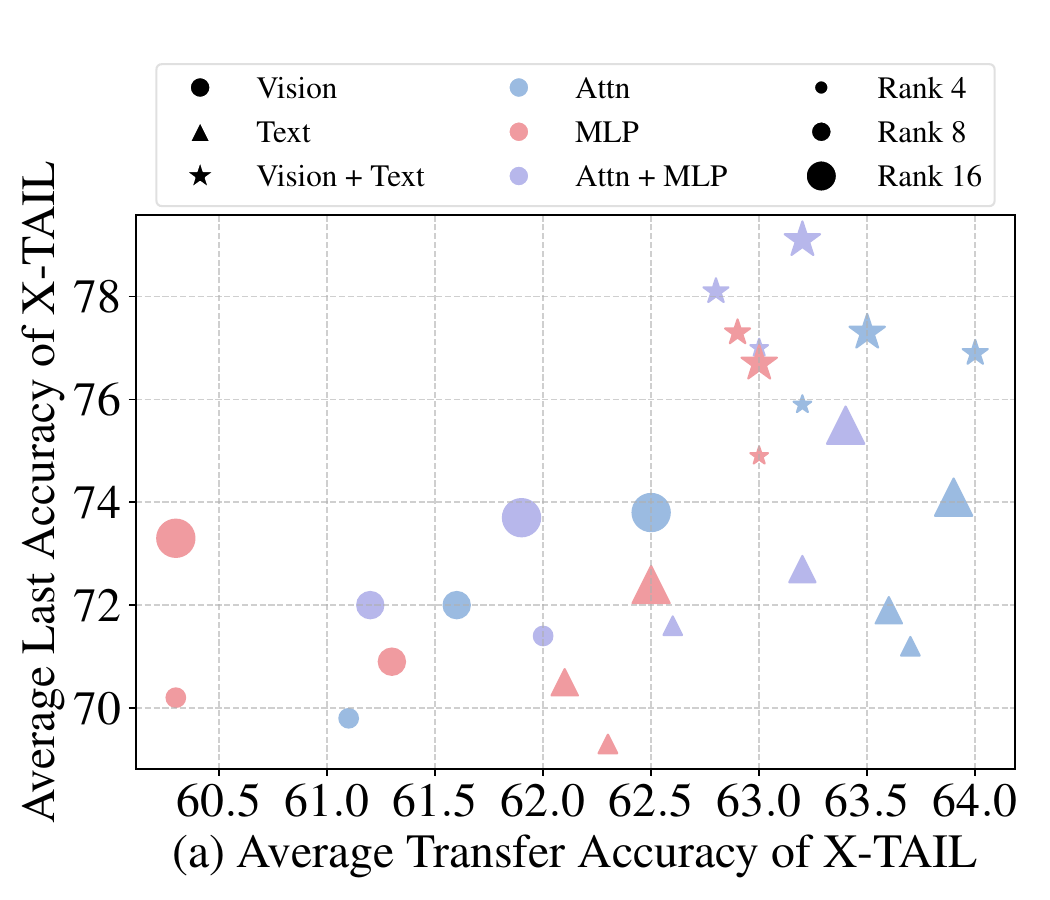}
    \includegraphics[width=0.45\linewidth]{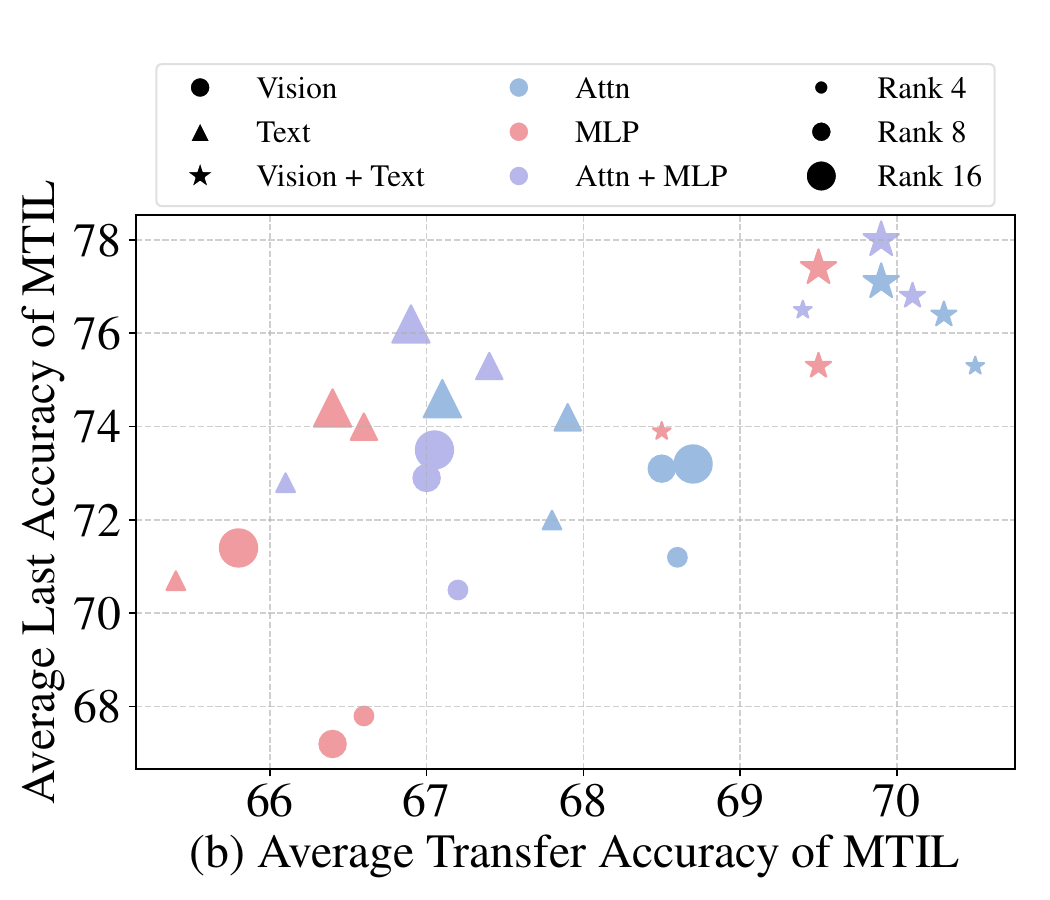}
    \captionof{figure}{Ablations on different insertion locations and the impact of varying the initial rank.}
    \label{fig:lora_loc_ablate}
  \end{minipage}\hfill
  \begin{minipage}[t]{0.49\linewidth}
    \centering
    \includegraphics[width=0.45\linewidth]{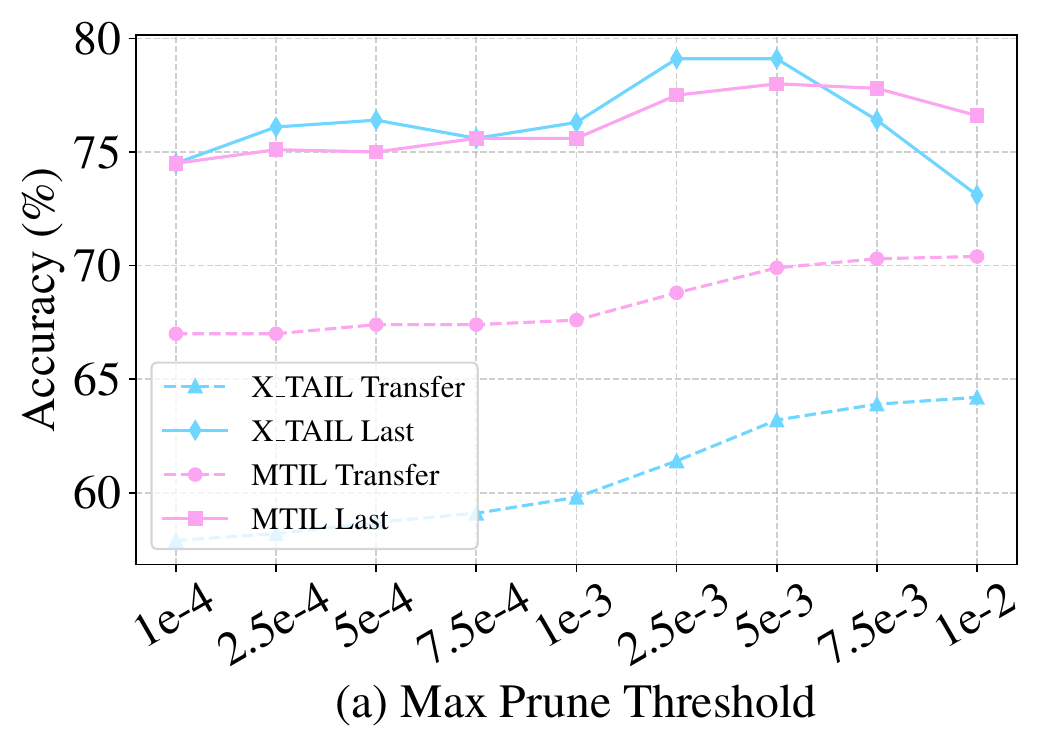}
    \includegraphics[width=0.45\linewidth]{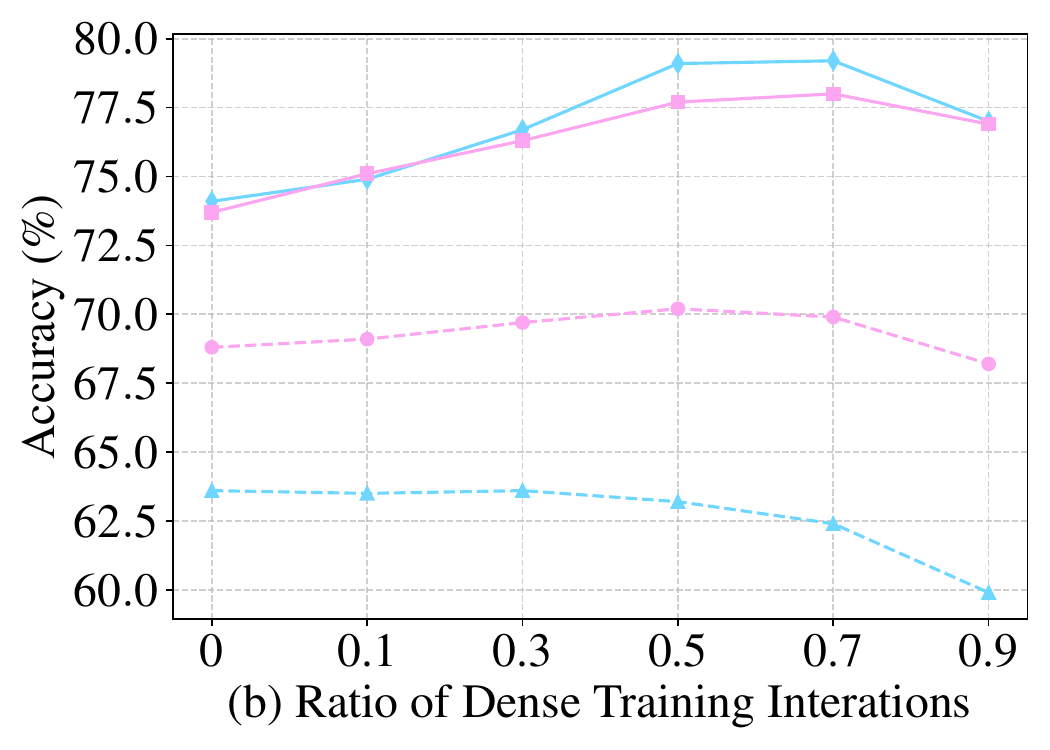}
    \captionof{figure}{Analyses (a) maximum pruning threshold and (b) ratios of dense training iterations.}
    \label{fig:sora_ablate}
  \end{minipage}
\end{figure*}

\begin{table}[!tb]
\centering
\small
\resizebox{0.95\linewidth}{!}{\begin{tabular}{@{}lcccccc@{}}
\toprule
After task $t$ & 1 & 2 & 4 & 6 & 8 & 10 \\ \midrule
LoRA ($r{=}16$)  & 4.25 & 6.59 & 11.34 & 16.94 & 22.94 & 50.11 \\
\rowcolor{Purple!8}
\textbf{CoDyRA} & \textbf{3.13} & \textbf{3.83} & \textbf{4.71} & \textbf{5.75} & \textbf{6.21} & \textbf{7.51} \\ \bottomrule
\end{tabular}}
\vspace{-0.2cm}
\caption{Cumulative parameter shift $\|\theta_t-\theta_0\|_F$.
}
\label{tab:drift}
\end{table}
\begin{table}[!tb]
\centering\small
\resizebox{0.7\linewidth}{!}{
\begin{tabular}{lcc}
\toprule
Method & Effective Rank & $\|\Delta\mathbf{W}\|_F$ \\ \midrule
LoRA ($r{=}1$)  & 1.0  & 4.21 \\
LoRA ($r{=}4$)  & 4.0  & 5.37 \\
LoRA ($r{=}8$)  & 8.0  & 6.25 \\
LoRA ($r{=}16$) & 16.0 & 7.44 \\
\midrule
\rowcolor{FigCap}
\textbf{CoDyRA} & \textbf{8.3} & \textbf{3.22} \\ \bottomrule
\end{tabular}}
\vspace{-0.15cm}
\caption{Per-task update magnitude on CLIP CL. CoDyRA attains the lowest $\|\Delta\mathbf{W}\|_F$ despite moderate effective rank.}
\label{tab:norm_lowest}
\end{table}

\noindent\textbf{Bounded parameter shift via rank minimization.}
A natural concern with merge-after-each-task is that accumulated perturbations drift the backbone from its pretrained state. We measure $\|\theta_t-\theta_0\|_F$ across CLIP CL tasks (Table~\ref{tab:drift}): CoDyRA grows sublinearly ($\sqrt{t}$ fit) to $7.51$, fixed-rank LoRA grows near-linearly to $50.11$. This gap between the two isolates rank minimization from norm regularization. Pruning components to zero (Figs.~\ref{fig:aircraft_rank_stat}--\ref{fig:pets_rank_stat}) freezes their parameters, confining the shift to a small task-relevant subspace; consistent with this, consecutive-task update overlap (linear CKA) is lower for CoDyRA ($0.0038$) than for fixed-rank LoRA ($0.0094$; Appendix~\ref{supp:cka}). \reva{CoDyRA also attains the lowest per-task update magnitude: at effective rank $8.3$, $\|\Delta\mathbf{W}\|_F = 3.22$, below every fixed-rank LoRA including $r{=}1$ (Table~\ref{tab:norm_lowest}), isolating the magnitude mechanism (App.~\ref{supp:two_mechanisms}). Directional interference is characterized separately via gradient-conflict and norm-matched probes (App.~\ref{supp:norm_matched}).}

\noindent\textbf{General capability preserved with modest positive transfer.}
Model complementation freezes the PTM, so its capabilities can at best be \emph{gated}, never updated. Reference-constrained methods update the backbone but anchor each step with stored data, risking visible forgetting. Under our general regime, the update-based criterion targets forgetting on $\theta_{t-1}$, covering both general capability and prior-task knowledge. Table~\ref{tab:unseen_transfer} reports zero-shot accuracy on three held-out datasets. Model complementation baselines \citep{boosting,rail} are pinned to the initial CLIP zero-shot accuracy by construction, while CoDyRA's merged updates yield small but consistent gains, indicating the merged shift carries modest positive transfer rather than catastrophic perturbation. More detailed breakdown analysis in Appendix~\ref{supp:drift}.

\begin{table}[!tb]
\centering
\resizebox{\linewidth}{!}{
\begin{tabular}{@{}lcccc@{}}
\toprule
\multirow{2}{*}{Method} & \multicolumn{4}{c}{Zero-shot Retention on Unseen Data} \\
 & CIFAR100 & Places365 & ImageNet-1k & \textit{Average} \\ \midrule
Zero-Shot \citep{clip} & 68.24 & 33.77 & 66.72 & 56.24 \\
MoE-Adapter$\dagger$ \citep{boosting} & 68.24 & 33.77 & 66.72 & 56.24 \\
RAIL$\dagger$ \citep{rail} & 68.24 & 33.77 & 66.72 & 56.24 \\ \midrule
\rowcolor{Purple!8}
\textbf{\ours} & \textbf{68.95} & \textbf{36.52} & \textbf{68.52} & \textbf{58.00} \\ \bottomrule
\end{tabular}
}
\vspace{-0.2cm}
\caption{General capability after X-TAIL CLIP CL on three held-out benchmarks. Complementation methods cap at zero-shot; CoDyRA modestly improves.}
\label{tab:unseen_transfer}
\vspace{-0.3cm}
\end{table}

\noindent\textbf{Rank allocation analysis.}
Per-module active-rank distributions (Figs.~\ref{fig:aircraft_rank_stat}, \ref{fig:pets_rank_stat}; extended in Appendix~\ref{supp:vis_heatmap}) differ systematically across both modules and tasks. Aircraft concentrates rank in the text-encoder Attention Output projection and deeper layers, while Pets shifts allocation to the vision-encoder MLP and to first/last layers; the same mechanism redistributes capacity to where each task needs it. This instantiates Takeaway~3 and aligns with the dimensional clause of the forgetting bound: confining each task's update to a small subspace leaves the rest of the parameter space available for subsequent tasks.

\subsection{Ablation Studies}

\noindent\textbf{Effects of insertion locations and initial ranks.}
We study the impact of insertion locations and initial ranks (Fig.~\ref{fig:lora_loc_ablate}), consistent with our earlier analyses (Fig.~\ref{fig:lora_forgetting}). Training restricted to the vision encoder significantly reduces transfer accuracy, indicating degraded pre-trained capabilities. At certain locations, increasing rank improves adaptation but does not always hinder transfer accuracy.

Interestingly, restricting updates to only Attention modules can even surpass our reported state-of-the-art Transfer results (Table~\ref{tab:few_mtil}, Table~\ref{tab:few_xtail}), but at the cost of weaker adaptation. To balance adaptation and retention, we update across all modules by default with $r_{\text{init}}{=}16$ (a standard LoRA default); the method prunes from this ceiling, so the effective rank per module is determined by the $\ell_1$ penalty rather than the initialization.

\noindent\textbf{Pruning threshold and dense-iteration ratio.}
Fig.~\ref{fig:sora_ablate} shows a clean trade-off: higher $\kappa_{\max}$ prunes more noisy ranks (improves Last) but eventually limits adaptation, and the dense ratio sets when pruning begins (too early kills useful ranks; too late retains noise). Within each benchmark a single setting is used across all tasks (dense ratio 0.7 for MTIL, 0.5 for X-TAIL; selected on the first task order and fixed thereafter).

\section{Conclusion}
\vspace{-0.1cm}
We studied continual learning through \emph{capacity control}, the effective rank of each parameter update. An empirical probe identifies a moderate-rank sweet spot that varies by module and task; a theoretical result connects rank to forgetting. Building on these findings, CoDyRA applies sparsity-promoting regularization on per-component importance weights to protect general capability and prior-task knowledge under a single update-based criterion. Rank minimization serves as a forgetting regularizer in the CL scenario. Across MTIL, X-TAIL, and TRACE, CoDyRA achieves strong accuracy with minimal forgetting and no inference overhead.

\section*{Limitations}
While our method performs competitively across MTIL, X-TAIL, and TRACE, several limitations remain. Our rank-based forgetting regularization is agnostic to the direction of the update relative to the loss landscape, and does not explicitly model inter-task dependencies (e.g., task similarity or shared rank structure), since prior-task updates are merged into the backbone after each task. Extending capacity control with alignment-aware regularization and explicit inter-task coupling is a promising future research direction.

\section*{Acknowledgements}
This work was partially supported by the ARC DECRA Fellowship (DE230101591), and the ARC Discovery Project Grant (DP260103379). H. Lu was affiliated with CSIRO Data61 through a PhD scholarship and acknowledges the support of the Google PhD Fellowship.

\bibliography{custom}

\clearpage
\newpage
\appendix

\subsubsection*{The Use of Large Language Models (LLMs)} 
Large language models (LLMs) were employed in preparing this manuscript, specifically to assist with grammar checking and polishing. Our experiments involve training and evaluating LLMs as part of the experimental setup. All technical content, experimental design, and analysis were conceived and carried out by the authors.

\section{Method and Theory Details}
\label{supp:method}
This section supplements Sec.~\ref{sec:theory} and Sec.~\ref{sec:our_method} of the main paper. We provide proofs of the propositions stated in the theoretical bounds (Appendix~\ref{supp:theory}), the full derivation of the soft-thresholding update used by CoDyRA (Appendix~\ref{supp:derivation}), and a discussion connecting rank minimization to forgetting (Appendix~\ref{supp:ortho}).

\subsection{Proof of the Rank-Dependent Forgetting Bound}
\label{supp:theory}

\noindent\textbf{Notation.} Let $\theta_{t-1}\in\mathbb{R}^{N}$ be the model parameters after task $t{-}1$, and $\theta_t = \theta_{t-1}+\Delta\theta$ after a LoRA update on task $t$, with $\Delta\theta$ supported on a single weight matrix $\mathbf{W}\in\mathbb{R}^{d\times k}$ updated by $\Delta\mathbf{W} = \mathbf{B}\mathbf{A}$, $\mathbf{B}\in\mathbb{R}^{d\times r}$, $\mathbf{A}\in\mathbb{R}^{r\times k}$, and effective rank $\rho := \mathrm{rank}(\Delta\mathbf{W}) \le r$. Since $\Delta\theta$ is zero outside $\mathbf{W}$, $\|\Delta\theta\|_F = \|\Delta\mathbf{W}\|_F$; the proof below uses $\Delta\theta$ for the Taylor expansion and $\Delta\mathbf{W}$ for LoRA-specific bounds. The reference loss is $\ell(\theta;\mathcal{D}_{\mathrm{ref}})$, and forgetting is $F(\theta_t):=\ell(\theta_t;\mathcal{D}_{\mathrm{ref}})-\ell(\theta_{t-1};\mathcal{D}_{\mathrm{ref}})$.

\noindent\textbf{Proposition 1 (Rank-dependent forgetting bound).}
\emph{If $\ell(\cdot;\mathcal{D}_{\mathrm{ref}})$ is $L$-smooth on a convex neighborhood of $\theta_{t-1}$ containing $\theta_t$, then}
\begin{equation*}
F(\theta_t) \;\le\; \|\nabla_\theta\ell(\theta_{t-1};\mathcal{D}_{\mathrm{ref}})\|_F\,\|\Delta\theta\|_F + \tfrac{L}{2}\|\Delta\theta\|_F^2,
\end{equation*}
\emph{with $\|\Delta\theta\|_F \le \sqrt{\rho}\,\|\mathbf{B}\|_2\|\mathbf{A}\|_2$. Moreover, $\Delta\mathbf{W}$ has row and column space of dimension at most $\rho$.}

\noindent\textit{Proof.} By the descent lemma for $L$-smooth functions,
\(\ell(\theta_t)\le\ell(\theta_{t-1})+\langle\nabla\ell(\theta_{t-1}),\Delta\theta\rangle+\tfrac{L}{2}\|\Delta\theta\|_F^2.\)
Rearranging gives $F(\theta_t)\le\langle\nabla\ell(\theta_{t-1}),\Delta\theta\rangle+\tfrac{L}{2}\|\Delta\theta\|_F^2$; Cauchy--Schwarz yields the displayed inequality. For the norm bound, $\Delta\mathbf{W}=\mathbf{B}\mathbf{A}$ has at most $\rho$ non-zero singular values, so $\|\Delta\mathbf{W}\|_F\le\sqrt{\rho}\,\|\Delta\mathbf{W}\|_2\le\sqrt{\rho}\,\|\mathbf{B}\|_2\|\mathbf{A}\|_2$ (rank--Frobenius inequality + submultiplicativity). The subspace clause follows from $\mathrm{rank}(\mathbf{B}\mathbf{A})\le\rho$. $\square$

\noindent\emph{Remarks.} (i) Both terms in the upper bound shrink with $\rho$ (given fixed $\|\mathbf{B}\|_2, \|\mathbf{A}\|_2$); the bound holds for any $\mathcal{D}_{\mathrm{ref}}$, so the same rank-based criterion covers pretrained-capability retention and prior-task retention without access to either distribution. (ii) The bound controls magnitude and dimensional reach, not alignment between $\Delta\theta$ and $\nabla\ell$; alignment-aware control is deferred to future work (Limitations).


Decomposing $\Delta\mathbf{W}=\sum_{i=1}^{r}w_i\,\mathbf{B}_{:,i}\mathbf{A}_{i,:}$ (Eq.~\ref{eq:dyra}) gives $\rho \le \|\mathbf{w}\|_0$. The $\ell_1$ penalty on $\mathbf{w}$ in Eq.~\ref{eq:loss} is the convex envelope of $\|\cdot\|_0$ on $[-1,1]^r$ \citep{tibshirani1996regression}; the proximal soft-thresholding update (Eq.~\ref{eq:sparse_update}) drives small $w_i$ to exactly zero, directly reducing $\rho$ and tightening the bound.

\reva{\subsection{Empirical Norm Stability Across Ranks}
\label{supp:norm_stability}

Prop.~\ref{prop:forget}'s rank-based statement assumes near-constant operator norms $\|\mathbf{A}\|_2, \|\mathbf{B}\|_2$. We verify this on the CLIP CL stream by measuring these norms (averaged across updated matrices), reported in Table~\ref{tab:supp_norm_stability}.

\begin{table}[!ht]
\centering\small
\begin{tabular}{lcccc}
\toprule
Method & eff-rank & $\|\mathbf{A}\|_2$ & $\|\mathbf{B}\|_2$ & $\|\Delta\mathbf{W}\|_F$ \\ \midrule
LoRA ($r{=}1$)  & 1.0  & 0.759 & 0.389 & 4.21 \\
LoRA ($r{=}8$)  & 8.0  & 0.742 & 0.389 & 6.25 \\
LoRA ($r{=}16$) & 16.0 & 0.763 & 0.389 & 7.44 \\
CoDyRA          & 8.3  & 0.739 & 0.348 & 3.22 \\ \bottomrule
\end{tabular}
\caption{Operator norms remain near-constant across fixed rank ($\|\mathbf{A}\|_2 \approx 0.74$, $\|\mathbf{B}\|_2 \approx 0.39$ for LoRA); $\|\Delta\mathbf{W}\|_F$ grows with rank only through the summation of more rank-one components. CoDyRA leaves $\|\mathbf{A}\|_2$ near LoRA's value and lowers $\|\mathbf{B}\|_2$ modestly ($0.389{\to}0.348$), further reducing $\|\Delta\mathbf{W}\|_F$ by suppressing high-magnitude low-relevance components through the gates.}
\label{tab:supp_norm_stability}
\end{table}

Two mechanisms keep operator norms stable throughout training, not merely at convergence: weight decay applied to $\mathbf{A}, \mathbf{B}$ (none on the gates $\mathbf{w}$), and the proximal soft-thresholding step, which carries each component's magnitude in $w_i$ rather than in $\mathbf{B}, \mathbf{A}$. Rank therefore governs the bound as claimed in Prop.~\ref{prop:forget}.}

\reva{\subsection{Two Mechanisms: Interference and Magnitude}
\label{supp:two_mechanisms}

A second-order expansion of forgetting $F(\Delta\mathbf{W}) := \ell(\mathbf{W}_0 + \Delta\mathbf{W}) - \ell(\mathbf{W}_0)$ gives, under $L$-smoothness (descent lemma):
\begin{equation}
F(\Delta\mathbf{W}) \;\le\; \underbrace{\langle \nabla\ell(\mathbf{W}_0), \Delta\mathbf{W} \rangle}_{\text{interference (directional)}} + \underbrace{\tfrac{L}{2}\|\Delta\mathbf{W}\|_F^2}_{\text{magnitude}}.
\end{equation}
This is a pre-Cauchy--Schwarz version of Prop.~\ref{prop:forget}: applying Cauchy--Schwarz to the first term yields the $\|\nabla\ell\|_F\,\|\Delta\mathbf{W}\|_F$ bound used in Prop.~\ref{prop:forget}, which absorbs the direction-dependence of the first-order interference. Rank enters both terms via $\|\Delta\mathbf{W}\|_F \le \sqrt{\rho}\,\|\mathbf{B}\|_2\|\mathbf{A}\|_2$ (magnitude), and via the row/column subspace bound $\le \rho$ (interference confinement).

CoDyRA reduces forgetting through the magnitude mechanism. By selectively pruning low-relevance components while retaining task-relevant ones, it attains $\|\Delta\mathbf{W}\|_F$ lower than every fixed-rank baseline including $r{=}1$ (Table~\ref{tab:norm_lowest} in the main paper), without collapsing to the extreme-low-rank regime where directional interference dominates. The interference mechanism is characterized separately in App.~\ref{supp:norm_matched}.}

\subsection{Full Derivation of Eq.~\ref{eq:sparse_update}}
\label{supp:derivation}

For clarity, in this subsection we omit the per-task superscript $t$ of the main paper notation and instead use subscript $t$ to denote the training \emph{iteration}. At iteration $t$, the regularized training loss (Eq.~\ref{eq:loss} in the main paper) is
\begin{equation}
\label{eq:supp_loss}
    \mathcal{L}_{\text{train}}(\Delta_t) \;:=\; \mathcal{L}_{\text{sup}}(\Delta_t) \,+\, \lambda_t \sum_{m=1}^{M} \|\mathbf{w}^{m}_t\|_1,
\end{equation}
where $\Delta_t = \{\mathbf{w}^{m}_t,\mathbf{B}^{m}_t,\mathbf{A}^{m}_t\}_{m=1}^M$ collects all trainable parameters at iteration $t$, $\mathcal{L}_{\text{sup}}$ is the supervised loss, $M$ is the number of LoRA-equipped weight matrices, and $\lambda_t\ge 0$ is a scheduled $\ell_1$ regularization strength.

To handle the non-differentiable $\ell_1$ term, we apply proximal gradient descent~\citep{beck2009fast}. With learning rate $\eta_t>0$, define the gradient-only step on the supervised loss as
\begin{equation}
\label{eq:supp_grad_step}
    \hat{\mathbf{w}}^{m}_t \;:=\; \mathbf{w}^{m}_t \,-\, \eta_t \, \nabla_{\mathbf{w}^m} \mathcal{L}_{\text{sup}}(\Delta_t).
\end{equation}
The proximal update of the importance weights is then
\begin{equation}
\label{eq:supp_prox}
    \mathbf{w}_{t+1}^m \,\leftarrow\, \arg\min_{\mathbf{w}^m} \,\Big\{ \eta_t \lambda_t \,\|\mathbf{w}^m\|_1 + \tfrac{1}{2}\,\|\mathbf{w}^m - \hat{\mathbf{w}}^{m}_t\|_2^2 \Big\},
\end{equation}
which trades off sparsity (the $\ell_1$ term) against proximity to the gradient step (the quadratic term).

Eq.~\ref{eq:supp_prox} admits a closed form: the well-known elementwise soft-thresholding operator
\begin{equation}
\label{eq:supp_soft}
    \mathcal{T}_{\kappa}(x) \;=\; \mathbbm{1}(|x|>\kappa) \cdot \big(x - \mathrm{sign}(x)\cdot\kappa\big),
\end{equation}
yielding
\begin{equation}
\label{eq:supp_update}
    \mathbf{w}_{t+1}^m \,\leftarrow\, \mathcal{T}_{\kappa_t}\!\big(\hat{\mathbf{w}}^{m}_t\big), \qquad \kappa_t \,=\, \eta_t\,\lambda_t,
\end{equation}
applied componentwise. Here $\mathbbm{1}(\cdot)$ returns $1$ when its argument is true and $0$ otherwise, and $\mathrm{sign}(\cdot)$ returns the sign of its input. Eq.~\ref{eq:supp_update} is the dynamic update of the main paper's Eq.~\ref{eq:sparse_update}. In practice we schedule $\lambda_t$ (rather than treating it as a fixed constant) so that the effective threshold $\kappa_t=\eta_t\lambda_t$ follows the linear ramp from $0$ to $\kappa_{\max}$ described in Sec.~\ref{sec:our_method} and App.~\ref{supp:impl}.

\reva{\subsection{Comparison with Orthogonal-Subspace Regularization Methods}\label{supp:ortho}}
As shown in Eqs.~(\ref{eq:dyra}) and~(\ref{eq:loss}), we minimize the effective rank to regularize the update magnitude $\|\Delta \mathbf{W}^{t,m}\|_F$ via $R(\Delta \mathbf{W}^{t,m})$. Minimizing $\|\mathbf{w}^{t,m}\|_1$ with a small effective rank encourages minimization of $\|\Delta \mathbf{W}^{t,m}\|_F$ for given $\mathbf{A}^{t,m}$ and $\mathbf{B}^{t,m}$ (according to Eq.~(\ref{eq:dyra})). This regularization limits the update $\Delta \mathbf{W}^{t,m}$ to the minimal dimension necessary, and softly reduces capacity in updates that conflict with critical information encoded in the learned parameters, without introducing additional assumptions. 

Another line of work regularizes updates (e.g., LoRA weights or gradients) to lie in subspaces orthogonal to existing parameters \cite{wang2023orthogonal,inflora}. Orthogonal-regularization methods and CoDyRA both aim to mitigate interference, but adopt different strategies with distinct trade-offs, which can be summarized along three aspects.


(1) Orthogonal regularization aims to constrain updates to lie in subspaces orthogonal to previous parameters, which effectively reduces interference but also limits the ability to acquire new knowledge. This rigid geometric constraint may suppress learning new knowledge, and thus is typically softened or relaxed in practice.
In contrast, CoDyRA regularizes update simplicity (via rank sparsity) rather than geometry, allowing alignment with prior knowledge when beneficial while keeping updates compact to avoid unintended overwriting. 
Both aim to alleviate interference (for stability) but trade off learning new tasks (for plasticity).


(2) Explicit orthogonal-regularization methods typically require storing the information of the already-learned subspace (e.g., projection matrices or singular vectors) to compute orthogonality losses, introducing additional memory overhead and dependence on past task statistics. 
In contrast, CoDyRA mitigates interference implicitly through sparsity over rank importance. This strictly local objective requires no access to previous tasks, aligning with our goal of natural, memory-free PTM updates.


(3) Standard orthogonal methods typically assume a fixed rank and focus on rotating it. When this rank exceeds a task’s intrinsic dimensionality, the extra dimensions behave as noise, causing unnecessary weight perturbation.
CoDyRA actively prunes such dimensions: a rank pruned to zero induces, by definition, ``zero interference'' with previous tasks. This yields ``zero overlap'' for irrelevant dimensions while preserving flexible updates for task-relevant ones.


By mitigating forgetting through different mechanisms and trade-offs, CoDyRA and orthogonal-based methods are complementary. For example, the task-relevant ranks selected by CoDyRA (with nonzero $\mathbf{w}^{t,m}$) can be further regularized using orthogonal constraints.

\section{Detailed Experimental Setup and Extended Results}\label{supp:experiment}

\subsection{Further Details of Experiments}
\label{supp:datasets}

\noindent\textbf{Details of CLIP CL experiment settings.}
The MTIL setting consists of 1,201 classes drawn from 11 diverse datasets:
Aircraft~\citep{maji2013fine},
Caltech101~\citep{fei2004learning},
CIFAR100~\citep{krizhevsky2009learning},
DTD~\citep{cimpoi2014describing},
EuroSAT~\citep{helber2019eurosat},
Flowers~\citep{nilsback2008automated},
Food~\citep{bossard2014food},
MNIST~\citep{deng2012mnist},
OxfordPet~\citep{parkhi2012cats},
Cars~\citep{krause20133d}, and SUN397~\citep{xiao2010sun}. In the X-TAIL setting, a total of 10 datasets are used, with CIFAR100~\citep{krizhevsky2009learning} excluded to prevent domain overlap, following the protocol in~\citep{rail}. In line with \citep{rail}, we evaluate MTIL under both 5-shot and full-shot settings, while X-TAIL is evaluated using a 16-shot setting.

\noindent\textbf{Details of LLM CL experiment settings.}
We adopt the TRACE benchmark \citep{wang2023trace}, designed for evaluating continual learning in LLMs such as LLaMA \citep{touvron2023llama,dubey2024llama} and Gemma \citep{team2024gemma}. TRACE comprises eight diverse datasets standardized into a unified format, spanning domain-specific understanding, multilingual processing, code generation, and mathematical reasoning. Specifically, the tasks are: C-STANCE \citep{zhao2023c}, FOMC \citep{shah2023trillion}, MeetingBank \citep{hu2023meetingbank}, Py150 \citep{lu2021codexglue}, ScienceQA \citep{lu2022learn}, NumGLUE-cm \citep{mishra2022numglue}, NumGLUE-ds \citep{mishra2022numglue}, and 20Minuten \citep{gonzales2021new}.

Given this diversity, we employ task-specific metrics to capture performance nuances: {Accuracy} is used for classification and reasoning tasks (C-STANCE, FOMC, ScienceQA, NumGLUE); {ROUGE-L} for summarization (MeetingBank); {SARI} for multilingual text simplification (20Minuten); and a {similarity score} for code generation (Py150).

\subsection{Evaluation Metrics}
\label{supp:metrics}
To evaluate the plasticity and stability of our method, we employ the following metrics. Let $N$ denote the total number of tasks, and $A_{i,j}$ denote the performance (e.g., accuracy, ROUGE, or similarity score) on task $j$ after training on task $i$.

\paragraph{CLIP Continual Learning Metrics.}
For the CLIP experiments (MTIL and X-TAIL), we follow the metric definitions of \citep{zscl,boosting,rail}. Let $A_{i,j}$ denote accuracy on task $j$ after the model has been trained on tasks $1{:}i$, and let $N$ be the total number of tasks. The per-task metrics for task $j$ are:
\begin{itemize}
\item \textbf{Transfer Accuracy} ($j\geq 2$). Average zero-shot accuracy on task $j$ before it is learned, capturing retention of pre-trained capabilities:
\begin{equation}
\text{Transfer}_j \;=\; \frac{1}{j-1}\sum_{i=1}^{j-1} A_{i,j}.
\end{equation}
\item \textbf{Average Accuracy.} Full column mean of task $j$ across all $N$ training steps (including pre-learning steps), capturing stability throughout the stream:
\begin{equation}
\text{Avg}_j \;=\; \frac{1}{N}\sum_{i=1}^{N} A_{i,j}.
\end{equation}
\item \textbf{Last Accuracy.} Accuracy on task $j$ after all training is complete:
\begin{equation}
\text{Last}_j \;=\; A_{N,j}.
\end{equation}
\end{itemize}
Tables~\ref{tab:few_mtil} and~\ref{tab:few_xtail} report these per-task values and their mean across the $N$ tasks. Read together: Transfer probes retention of pretrained capabilities; Last probes final-task performance after the full stream; Average mediates between them.

\paragraph{LLM Continual Learning Metrics.}
For the TRACE benchmark, we adopt the standard metrics as in \citep{wang2023trace}:
\begin{itemize}
    \item \textbf{Overall Performance (OP):} The average performance across all learned tasks at the end of training.
    \begin{equation}
        \text{OP} = \frac{1}{N} \sum_{j=1}^{N} A_{N,j},
    \end{equation}
    where $A_{i,j}$ represents the performance on task $j$ after learning task $i$.
    \reva{\item \textbf{Forgetting:} Measures the average performance degradation of previous tasks $j < N$ after learning subsequent tasks. Lower is better.
    \begin{equation}
        \text{Forgetting} = \frac{1}{N-1} \sum_{j=1}^{N-1} (A_{j,j} - A_{N,j})
    \end{equation}}
\end{itemize}

\subsection{Implementation Details}
\label{supp:impl}
\textbf{CLIP experiments.} Following established protocols \citep{zscl,boosting,rail}, we employ the CLIP ViT-B/16 backbone \citep{clip}. CoDyRA is applied to all pre-trained weight matrices in both vision and text encoders and is initialized with a rank of 16. CoDyRA is optimized for 500 iterations per task using AdamW \citep{loshchilov2017decoupled}, with dense training ratios of 0.5 for X-TAIL and 0.7 for MTIL, setting $\kappa_{\max} = 0.005$.

\noindent\textbf{LLM experiments.} We adhere to the variable epoch schedule of prior work \citep{qian2025treelora} (\{2, 1, 3, 2, 1, 2, 2, 3\} epochs). We use an initial rank of 16, a dense training ratio of 0.5, and set $\kappa_{\max} = 0.005$. Optimization employs a learning rate of $1\times 10^{-4}$ and a batch size of 4. Training utilizes DeepSpeed ZeRO-2 with BF16 mixed-precision and a maximum context length of 1024 tokens. All experiments were conducted on a cluster of four Nvidia H100 GPUs.%

\noindent\textbf{Additional reproducibility details.}
The same configuration is used across all CLIP datasets in MTIL/X-TAIL and across all three LLM architectures on TRACE; no per-task or per-model tuning is applied beyond the dense-iteration ratio noted above.
(i)~\emph{LoRA scaling.} We use the standard LoRA $\alpha$ scaling with $\alpha{=}r_{\text{init}}{=}16$, so each rank-one component is scaled by $\alpha/r_{\text{init}}{=}1$; CoDyRA's per-component importance weight $w_i$ is multiplied on top of this and softens to zero under the proximal update.
(ii)~\emph{Initial rank per module.} Identical $r_{\text{init}}{=}16$ is used for every updated weight matrix (Q, K, V, O, FC, Proj in both encoders). The per-module \emph{effective} rank emerges from sparsity-promoting training (Figs.~\ref{fig:aircraft_rank_stat}--\ref{fig:pets_rank_stat}); we do not hand-pick per-module ranks.
(iii)~\emph{Threshold schedule.} The soft-thresholding $\kappa$ in Eq.~\ref{eq:sparse_update} ramps linearly from $0$ at the end of the dense-iteration phase to $\kappa_{\max}{=}0.005$ over the remaining iterations.
(iv)~\emph{Weight decay on $\mathbf{A},\mathbf{B}$.} AdamW weight decay is set to $0.01$ on $\mathbf{A}$ and $\mathbf{B}$ (no weight decay on $\mathbf{w}$ to avoid double-shrinking the sparsity signal). This bounds the operator norms of surviving ranks and is the training-time mechanism behind the near-constant $\|\mathbf{A}\|_2, \|\mathbf{B}\|_2$ we report in App.~\ref{supp:norm_stability}.
(v)~\emph{Task order.} CLIP MTIL/X-TAIL follow the original task order of \citep{zscl,rail}; TRACE follows the order of \citep{wang2023trace,qian2025treelora}.
(vi)~\emph{Prompts and augmentations.} Text prompts are the standard ``a photo of a [CLS]''; image augmentations are random resized crops and horizontal flips. The same prompts and augmentations are used for all baselines we re-run.
(vii)~\emph{Training cost.} On the CLIP CL benchmarks, one full MTIL pass (11 tasks $\times$ 500 iterations) takes roughly $\sim$45 minutes on a single H100. The proximal soft-thresholding step adds negligible wall-clock overhead over standard LoRA fine-tuning. Per-method breakdowns underlying Table~\ref{tab:comp_cost} (main paper): MoE-Adapter's train/test gap arises from its mixture-of-experts design (all experts active during training, only top-2 at inference); RAIL-Primal incurs overhead from high-dimensional feature expansion via projection and regression layers; RAIL-Dual additionally stores image features for all samples across 1,101 classes in X-TAIL.

\reva{\subsection{X-TAIL: Per-Domain Results}\label{supp:results}}

\begin{table*}[!t]
\centering
\resizebox{\linewidth}{!}{
\small
\renewcommand{\arraystretch}{1.05}
\begin{tabular}{l>{\centering\arraybackslash}p{1cm} >{\centering\arraybackslash}p{1cm} >{\centering\arraybackslash}p{1cm} >{\centering\arraybackslash}p{1cm} >{\centering\arraybackslash}p{1cm} >{\centering\arraybackslash}p{1cm} >{\centering\arraybackslash}p{1cm} >{\centering\arraybackslash}p{1cm} >{\centering\arraybackslash}p{1cm} >{\centering\arraybackslash}p{1cm} >{\centering\arraybackslash}p{1.6cm}}
\toprule
{\quad}\textbf{Method} & \rotatebox{90}{Aircraft} & \rotatebox{90}{Caltech101} & \rotatebox{90}{DTD} & \rotatebox{90}{EuroSAT} & \rotatebox{90}{Flowers} & \rotatebox{90}{Food} & \rotatebox{90}{MNIST} & \rotatebox{90}{OxfordPet} & \rotatebox{90}{Cars} & \rotatebox{90}{SUN397} & \textbf{\textit{Average}} \\
\midrule
\rowcolor{Green!10}\multicolumn{12}{l}{\emph{CLIP zero-shot reference}}\\
{\quad}Zero-shot \citep{clip} & 23.5 & 76.8 & 37.3 & 36.7 & 63.6 & 84.0 & 46.7 & 86.7 & 66.1 & 63.7 & 58.5 \\
\midrule
\rowcolor{Green!10}\multicolumn{12}{l}{\emph{Transfer}}\\
{\quad}Zero-shot \citep{clip} & -- & 76.8 & 37.3 & 36.7 & 63.6 & 84.0 & 46.7 & 86.7 & 66.1 & 63.7 & 62.4 \\
\multicolumn{12}{l}{\colorbox{FigComp}{\textbf{\textit{\small Model complementation}}}} \\
{\quad}MoE-Adapter$\dagger$ \citep{boosting} & -- & 71.0 & 34.9 & 19.2 & 63.0 & \textbf{86.6} & 20.0 & 87.2 & 63.7 & 58.6 & 56.0 \\
{\quad}LAE-LoRA \citep{gao2023unified} & -- & 76.2 & \textbf{39.9} & 43.8 & 66.5 & 83.3 & 42.7 & 87.5 & 63.0 & 62.9 & 62.9 \\
{\quad}TreeLoRA \citep{qian2025treelora} & -- & 76.0 & 36.3 & 34.0 & 58.5 & 77.2 & 43.3 & 82.2 & 49.8 & 55.8 & 57.0 \\
{\quad}InfLoRA \citep{inflora} & -- & 75.8 & 34.5 & 29.2 & 58.1 & 73.4 & 38.6 & 79.5 & 47.7 & 50.3 & 54.1 \\
{\quad}CL-LoRA \citep{he2025cl} & -- & 73.3 & 33.7 & 29.5 & 58.5 & 80.3 & 43.1 & 85.5 & 61.5 & 59.2 & 58.3 \\
{\quad}RAIL-Primal$\dagger$ \citep{rail} & -- & \textbf{76.8} & 37.3 & 36.7 & 63.6 & 84.0 & \textbf{46.7} & 86.7 & \textbf{66.1} & \textbf{63.7} & 62.4 \\
{\quad}\textbf{\ours$\dagger$} & -- & 74.3 & 36.8 & \textbf{44.2} & \textbf{69.9} & 83.5 & 42.8 & \textbf{88.9} & 64.6 & 63.4 & \textbf{63.2} \\
\midrule
\multicolumn{12}{l}{\colorbox{FigCont}{\textbf{\textit{\small Model continually learned}}}} \\
\multicolumn{12}{l}{\hspace{0.5em}\colorbox{FigRef}{\textit{\small reference-constrained}}} \\
{\quad}WiSE-FT \citep{wortsman2022robust} & -- & 70.1 & 31.9 & 25.3 & 56.3 & 79.8 & 29.9 & 74.9 & 45.6 & 56.8 & 52.3 \\
{\quad}ZSCL \citep{zscl} & -- & 73.3 & 32.6 & 36.8 & 62.1 & 83.8 & 42.1 & 83.6 & 56.5 & 60.2 & 59.0 \\
{\quad}LwF \citep{li2017learning} & -- & 66.6 & 26.9 & 19.5 & 51.0 & 78.4 & 26.6 & 68.9 & 35.5 & 56.1 & 47.7 \\
{\quad}iCaRL \citep{rebuffi2017icarl} & -- & 71.7 & 35.0 & 43.0 & 63.4 & \textbf{86.9} & 43.9 & 87.8 & 63.7 & 60.0 & 61.7 \\
\multicolumn{12}{l}{\hspace{0.5em}\colorbox{FigNaive}{\textit{\small naive LoRA}}} \\
{\quad}LoRA ($r{=}4$) & -- & 74.9 & \textbf{39.4} & 41.8 & 67.7 & 83.3 & 44.1 & 87.1 & 64.2 & 62.7 & 62.8 \\
{\quad}LoRA ($r{=}8$) & -- & \textbf{77.0} & 38.2 & 38.4 & 67.8 & 83.3 & \textbf{44.6} & 87.6 & 64.0 & 62.8 & 62.6 \\
{\quad}LoRA ($r{=}16$) & -- & 76.4 & 39.1 & 37.5 & 65.9 & 82.7 & 41.9 & 86.9 & 63.4 & 62.7 & 61.8 \\
\multicolumn{12}{l}{\hspace{0.5em}\colorbox{FigCap}{\textit{\small capacity-controlled (ours)}}} \\
{\quad}\textbf{\ours} & -- & 74.3 & 36.8 & \textbf{44.2} & \textbf{69.9} & 83.5 & 42.8 & \textbf{88.9} & \textbf{64.6} & \textbf{63.4} & \textbf{63.2} \\
\midrule
\rowcolor{Green!10}\multicolumn{12}{l}{\emph{Average}}\\
\multicolumn{12}{l}{\colorbox{FigComp}{\textbf{\textit{\small Model complementation}}}} \\
{\quad}MoE-Adapter$\dagger$ \citep{boosting} & 43.6 & 77.9 & 52.1 & 34.7 & 75.9 & \textbf{86.3} & 45.2 & 87.4 & 66.6 & 60.2 & 63.0 \\
{\quad}LAE-LoRA \citep{gao2023unified} & 29.2 & 81.3 & 50.7 & 68.8 & 77.4 & 83.8 & 53.2 & 88.5 & 66.3 & 64.2 & 66.4 \\
{\quad}TreeLoRA \citep{qian2025treelora} & 19.3 & 81.4 & 55.3 & 63.9 & 77.6 & 80.4 & 64.3 & 85.0 & 54.8 & 57.6 & 64.0 \\
{\quad}InfLoRA \citep{inflora} & 38.3 & 84.2 & 60.1 & 45.8 & 79.0 & 77.5 & 61.8 & 82.8 & 52.2 & 52.2 & 63.4 \\
{\quad}CL-LoRA \citep{he2025cl} & 39.7 & 79.2 & 56.7 & 58.5 & 79.1 & 81.1 & 64.0 & 78.0 & 63.6 & 60.6 & 66.0 \\
{\quad}RAIL-Primal$\dagger$ \citep{rail} & 42.4 & \textbf{89.8} & 55.7 & 68.5 & \textbf{84.0} & 83.3 & \textbf{65.3} & 85.8 & \textbf{67.9} & \textbf{64.5} & 70.7 \\
{\quad}\textbf{\ours$\dagger$} & \textbf{43.9} & 81.6 & \textbf{60.6} & \textbf{78.4} & \textbf{84.0} & 84.9 & 64.6 & \textbf{90.5} & 67.4 & 64.4 & \textbf{72.0} \\
\midrule
\multicolumn{12}{l}{\colorbox{FigCont}{\textbf{\textit{\small Model continually learned}}}} \\
\multicolumn{12}{l}{\hspace{0.5em}\colorbox{FigRef}{\textit{\small reference-constrained}}} \\
{\quad}WiSE-FT \citep{wortsman2022robust} & 27.1 & 76.5 & 40.9 & 31.3 & 68.7 & 81.6 & 31.4 & 74.7 & 51.7 & 58.4 & 54.2 \\
{\quad}ZSCL \citep{zscl} & 36.0 & 75.0 & 40.7 & 40.5 & 71.0 & 85.3 & 46.3 & 83.3 & 60.7 & 61.5 & 60.0 \\
{\quad}LwF \citep{li2017learning} & 24.7 & 79.7 & 38.3 & 36.9 & 63.9 & 81.0 & 36.5 & 71.9 & 42.7 & 56.7 & 53.2 \\
{\quad}iCaRL \citep{rebuffi2017icarl} & 25.4 & 72.1 & 37.5 & 51.6 & 65.1 & \textbf{87.1} & 59.1 & 88.0 & 63.7 & 60.1 & 61.0 \\
\multicolumn{12}{l}{\hspace{0.5em}\colorbox{FigNaive}{\textit{\small naive LoRA}}} \\
{\quad}LoRA ($r{=}4$) & 23.5 & 78.4 & 40.6 & 47.3 & 71.4 & 83.7 & 50.1 & 87.5 & 65.1 & 64.0 & 61.2 \\
{\quad}LoRA ($r{=}8$) & 23.7 & 78.1 & 39.7 & 46.6 & 71.3 & 83.7 & 50.7 & 87.8 & 65.4 & 64.1 & 61.1 \\
{\quad}LoRA ($r{=}16$) & 23.6 & 78.5 & 40.5 & 43.0 & 70.4 & 82.8 & 49.5 & 87.5 & 64.9 & 64.0 & 60.5 \\
\multicolumn{12}{l}{\hspace{0.5em}\colorbox{FigCap}{\textit{\small capacity-controlled (ours)}}} \\
{\quad}\textbf{\ours} & \textbf{41.4} & \textbf{81.0} & \textbf{58.7} & \textbf{77.8} & \textbf{83.4} & 84.6 & \textbf{64.5} & \textbf{90.4} & \textbf{67.2} & \textbf{64.4} & \textbf{71.3} \\
\midrule
\rowcolor{Green!10}\multicolumn{12}{l}{\emph{Last}}\\
\multicolumn{12}{l}{\colorbox{FigComp}{\textbf{\textit{\small Model complementation}}}} \\
{\quad}MoE-Adapter$\dagger$ \citep{boosting} & 43.2 & 78.7 & 57.6 & 32.8 & 79.4 & 86.0 & 86.7 & 87.8 & 78.2 & 74.2 & 70.5 \\
{\quad}LAE-LoRA \citep{gao2023unified} & 25.6 & 81.5 & 50.2 & 73.2 & 79.5 & 85.0 & 70.4 & 91.1 & 72.9 & \textbf{75.7} & 70.5 \\
{\quad}TreeLoRA \citep{qian2025treelora} & 16.0 & 79.1 & 59.2 & 62.3 & 83.0 & 83.6 & 95.0 & 89.9 & 71.9 & 74.2 & 71.4 \\
{\quad}InfLoRA \citep{inflora} & 38.3 & 85.2 & \textbf{66.6} & 52.9 & 92.9 & 81.6 & 96.6 & 90.4 & 70.1 & 69.3 & 74.4 \\
{\quad}CL-LoRA \citep{he2025cl} & 39.7 & 79.8 & 62.5 & 70.9 & 92.9 & 81.9 & 95.2 & 90.5 & 72.4 & 73.1 & 75.9 \\
{\quad}RAIL-Primal$\dagger$ \citep{rail} & 41.7 & \textbf{94.0} & 66.0 & 86.4 & \textbf{97.2} & 82.4 & 93.1 & 83.6 & 75.0 & 71.3 & 79.1 \\
{\quad}\textbf{\ours$\dagger$} & \textbf{43.9} & 82.4 & \textbf{66.6} & \textbf{93.0} & 93.3 & \textbf{86.3} & \textbf{97.2} & \textbf{94.0} & \textbf{78.5} & 73.5 & \textbf{80.9} \\
\midrule
\multicolumn{12}{l}{\colorbox{FigCont}{\textbf{\textit{\small Model continually learned}}}} \\
\multicolumn{12}{l}{\hspace{0.5em}\colorbox{FigRef}{\textit{\small reference-constrained}}} \\
{\quad}WiSE-FT \citep{wortsman2022robust} & 21.8 & 76.8 & 42.9 & 20.8 & 77.5 & 84.9 & 30.7 & 76.6 & 75.8 & 72.5 & 58.0 \\
{\quad}ZSCL \citep{zscl} & 33.1 & 75.3 & 43.5 & 35.2 & 74.6 & \textbf{87.4} & 50.4 & 84.2 & \textbf{77.3} & 73.4 & 63.4 \\
{\quad}LwF \citep{li2017learning} & 25.5 & 72.1 & 38.9 & 55.4 & 65.5 & 87.3 & 81.9 & 88.6 & 63.6 & 61.5 & 64.0 \\
{\quad}iCaRL \citep{rebuffi2017icarl} & 25.5 & 72.1 & 38.9 & 55.4 & 65.5 & 87.3 & 81.9 & 88.6 & 63.6 & 61.5 & 64.0 \\
\multicolumn{12}{l}{\hspace{0.5em}\colorbox{FigNaive}{\textit{\small naive LoRA}}} \\
{\quad}LoRA ($r{=}4$) & 9.75 & 79.5 & 40.7 & 37.5 & 65.1 & 82.6 & 47.1 & 84.1 & 57.5 & \textbf{76.3} & 58.0 \\
{\quad}LoRA ($r{=}8$) & 12.7 & 78.7 & 39.6 & 42.6 & 62.4 & 82.3 & 46.8 & 83.6 & 60.2 & 76.1 & 58.5 \\
{\quad}LoRA ($r{=}16$) & 13.7 & 78.6 & 40.3 & 39.4 & 66.1 & 82.7 & 50.9 & 85.7 & 60.3 & 76.1 & 59.4 \\
\multicolumn{12}{l}{\hspace{0.5em}\colorbox{FigCap}{\textit{\small capacity-controlled (ours)}}} \\
{\quad}\textbf{\ours} & \textbf{37.7} & \textbf{81.5} & \textbf{65.1} & \textbf{89.9} & \textbf{91.4} & 85.5 & \textbf{96.8} & \textbf{93.3} & \textbf{77.3} & 73.5 & \textbf{79.2} \\
\bottomrule
\end{tabular}}
\caption{X-TAIL per-domain results. CoDyRA (ours) trains a rank-bounded LoRA on the current task and merges it into the backbone, storing no past-task data or parameters. $\dagger$: uses domain prediction at inference similar to \citep{boosting}. The two big groups (\emph{model complementation} vs.\ \emph{model continually learned}) are separated by a horizontal rule; \textbf{bold} marks the best per column within each big group.}
\label{tab:few_xtail}
\vspace{-0.2cm}
\end{table*}

Table~\ref{tab:few_xtail} compares X-TAIL baselines under the two big groups (\emph{model complementation} vs.\ \emph{model continually learned}) introduced in Table~\ref{tab:few_mtil}, separated by a horizontal rule. Within \emph{model complementation}, per-task LoRA variants (LAE-LoRA, TreeLoRA, InfLoRA, CL-LoRA) and inference-time routing methods (MoE-Adapter$\dagger$, RAIL-Primal$\dagger$) accumulate per-task structure without altering $\theta$; CoDyRA$\dagger$ leads each metric (Transfer 63.2, Average 72.0, Last 80.9) within this group. Within \emph{model continually learned}, the three reference-constrained baselines and three fixed-rank LoRA settings both modify $\theta$ without rank control; their Last gap to CoDyRA's adaptive rank selection (79.2) isolates the contribution of \colora{rank minimization}, and CoDyRA tops 9/11 columns of the trained Average block. Increasing the fixed LoRA rank improves learning at the cost of forgetting (Last drifts down from $r{=}4$ to $r{=}16$), consistent with Fig.~\ref{fig:lora_forgetting} and Takeaway~2.

\reva{\noindent\textbf{O-LoRA and TreeLoRA on CLIP CL.} Table~\ref{tab:supp_olora_treelora} reports the head-to-head comparison on 5-shot MTIL. CoDyRA leads on all three metrics.

\begin{table}[!ht]
\centering\small
\resizebox{\linewidth}{!}{
\begin{tabular}{lccc}
\toprule
Method & Transfer & Average & Last \\ \midrule
O-LoRA \citep{wang2023orthogonal} & 65.7 & 69.6 & 75.1 \\
TreeLoRA \citep{qian2025treelora} & 63.4 & 70.6 & 76.3 \\
\rowcolor{Purple!8}
\textbf{CoDyRA} & \textbf{70.1} & \textbf{73.3} & \textbf{78.0} \\ \bottomrule
\end{tabular}}
\caption{O-LoRA and TreeLoRA compared to CoDyRA on 5-shot MTIL (CLIP CL).}
\label{tab:supp_olora_treelora}
\end{table}

\noindent\textbf{Compute cost.} Both O-LoRA and TreeLoRA train 4.4M parameters, matching CoDyRA. However, they carry additional inference-time memory to maintain per-task subspaces (44.32M for O-LoRA, 22.50M for TreeLoRA), whereas CoDyRA merges each update into the backbone and adds no inference overhead.}

\reva{\subsection{Full-shot MTIL: Additional Results}\label{supp:full_mtil_analysis}}
Table~\ref{supp:full_mtil} reports full-shot MTIL under the same family taxonomy as Tables~\ref{tab:few_mtil} and \ref{tab:few_xtail}. Within \emph{model complementation}, CoDyRA$\dagger$ leads on Transfer (69.1), Average (77.0), and Last (85.8); other prompt-based methods \citep{l2p,dualprompt,sprompts,tang2025mind} are competitive on Last (S-Prompts 83.4, DIKI 85.1) but cap Transfer at zero-shot by construction. Within \emph{model continually learned}, plain CoDyRA (capacity-controlled LoRA merged into the backbone, no auxiliary module) leads on Average (76.0) and Last (84.9), improving over the strongest reference-constrained baseline ZSCL by +0.6 and +1.3, respectively. CoDyRA's Transfer matches CoDyRA$\dagger$'s by construction (capacity control alone preserves the unseen-task zero-shot signal on the merged backbone). Together, the two rows isolate the contribution of capacity-controlled backbone updates (plain CoDyRA over ZSCL) from that of the auxiliary domain predictor (CoDyRA$\dagger$ over CoDyRA).

\begin{table*}[thb]
\centering
\resizebox{\linewidth}{!}{
\begin{tabular}{l>{\centering\arraybackslash}p{1.2cm} >{\centering\arraybackslash}p{1.2cm}>{\centering\arraybackslash}p{1.2cm} >{\centering\arraybackslash}p{1.2cm} >{\centering\arraybackslash}p{1.2cm} >{\centering\arraybackslash}p{1.2cm} >{\centering\arraybackslash}p{1.2cm} >{\centering\arraybackslash}p{1.2cm} >{\centering\arraybackslash}p{1.2cm} >{\centering\arraybackslash}p{1.2cm} >{\centering\arraybackslash}p{1.6cm}}
\toprule
{\quad} 
\makecell[l]{ } &
\makecell[c]{\rotatebox{90}{Aircraft}} & 
\makecell[c]{\rotatebox{90}{Caltech101}} &
\makecell[c]{\rotatebox{90}{DTD}} & 
\makecell[c]{\rotatebox{90}{EuroSAT}} & 
\makecell[c]{\rotatebox{90}{Flowers}} & 
\makecell[c]{\rotatebox{90}{Food}} & 
\makecell[c]{\rotatebox{90}{MNIST}} & 
\makecell[c]{\rotatebox{90}{OxfordPet}} & 
\makecell[c]{\rotatebox{90}{Cars}} & 
\makecell[c]{\rotatebox{90}{SUN397}} & 
\makecell[c]{\textit{Average}} \\
\midrule
{\ours} {\emph{Transfer}} & -- & {$74.3^{\pm0.52}$} & $36.8 ^ {\pm
0.23}$	&$44.2 ^ {\pm
0.56}$	&$69.9 ^ {\pm
0.56}$	&$83.5 ^ {\pm
0.23}$  &$42.8 ^ {\pm
0.18}$	&$88.9 ^ {\pm
0.42}$  &$64.6 ^ {\pm
0.47}$	&$63.4 ^ {\pm
0.56}$  &$63.2 ^ {\pm
0.28}$ \\
{\ours} {\emph{Average}}& $41.4 ^ {\pm
0.28}$	&$81.0 ^ {\pm
0.38}$	&$58.7 ^ {\pm
0.26}$	&$77.8 ^ {\pm
0.47}$	&$83.4 ^ {\pm
0.39}$	&$84.6 ^ {\pm
0.28}$	&$64.5 ^ {\pm
0.14}$	&$90.4 ^ {\pm
0.4}$	&$67.2 ^ {\pm
0.23}$	&$64.4 ^ {\pm
0.47}$	&$71.3 ^ {\pm
0.18}$\\
{\ours} {\emph{Last}}& $37.7 ^ {\pm
0.42}$	&$81.5 ^ {\pm
0.24}$	&$65.1 ^ {\pm
0.63}$	&$89.9 ^ {\pm
0.55}$	&$91.4 ^ {\pm
0.38}$	&$85.5 ^ {\pm
0.16}$	&$96.8 ^ {\pm
0.08}$	&$93.3 ^ {\pm
0.3}$	&$77.3 ^ {\pm
0.66}$	&$73.5 ^ {\pm
0.21}$	&$79.2 ^ {\pm
0.18}$ \\
\bottomrule
\end{tabular}}
\caption{Statistical significance of CoDyRA results corresponding to Table~\ref{tab:few_xtail}.}
\label{tab:supp_training_stability}
\vspace{-0.3cm}
\end{table*}

\subsection{Training Stability on X-TAIL}\label{supp:training_stability}
Table~\ref{tab:supp_training_stability} reports the mean and standard deviation of CoDyRA's results in Table~\ref{tab:few_xtail}, averaged over three independent runs. 
The consistently low variance across metrics demonstrates that our method is stable to random initialization, highlighting CoDyRA's robust adaptive rank selection.

\subsection{Isolating the CL Framing from the Rank-Selection Mechanism}
\label{supp:adalora}

\reva{The importance-weight and soft-thresholding machinery overlaps with AdaLoRA; our contribution is repurposing this machinery as a forgetting regularizer, a role adaptive-rank methods neither claim nor test.}
To isolate the contribution of the CL-aware framing from the rank-selection mechanism itself, we apply AdaLoRA~\citep{adalora} and DyLoRA~\citep{dylora} to the same X-TAIL CL setup as CoDyRA: identical all-weight placement on every Attention/MLP matrix in both encoders, identical training budget, identical optimizer and base LR, and identical maximum rank ($r{=}16$). Both baselines share the rank-selection idea (singular-value importance for AdaLoRA; nested sub-rank training for DyLoRA) but were designed for single-task adaptation rather than CL: they impose no mechanism that ties update capacity to forgetting. Table~\ref{tab:supp_adalora} reports the comparison.

\begin{table}[!ht]
\centering\small
\resizebox{\linewidth}{!}{
\begin{tabular}{lccc}
\toprule
Method & Transfer & Average & Last \\ \midrule
AdaLoRA \citep{adalora} & 59.1 & 67.6 & 76.1 \\
DyLoRA \citep{dylora}   & 60.6 & 69.0 & 77.1 \\
\rowcolor{Purple!8}
\textbf{CoDyRA}         & \textbf{63.2} & \textbf{71.3} & \textbf{79.2} \\ \bottomrule
\end{tabular}}
\caption{Rank-selective LoRA baselines without CL framing on X-TAIL, matched to CoDyRA's placement, budget, optimizer, and $r{=}16$.}
\label{tab:supp_adalora}
\end{table}

CoDyRA outperforms both baselines on all three metrics. Since the underlying mechanism (adaptive per-component importance) is comparable, the gap isolates the contribution of the CL framing: the $\ell_1$-on-importance regularizer doubles as a forgetting regularizer (Prop.~\ref{prop:forget}), whereas AdaLoRA's SVD-importance scoring and DyLoRA's nested training optimize only the current-task objective.

\begin{table*}[!t]
\centering
\resizebox{\linewidth}{!}{
\small
\renewcommand{\arraystretch}{1.05}
\begin{tabular}{l>{\centering\arraybackslash}p{1cm} >{\centering\arraybackslash}p{1cm} >{\centering\arraybackslash}p{1cm} >{\centering\arraybackslash}p{1cm} >{\centering\arraybackslash}p{1cm} >{\centering\arraybackslash}p{1cm} >{\centering\arraybackslash}p{1cm} >{\centering\arraybackslash}p{1cm} >{\centering\arraybackslash}p{1cm} >{\centering\arraybackslash}p{1cm} >{\centering\arraybackslash}p{1cm} >{\centering\arraybackslash}p{1.5cm}}
\toprule
{\quad}\textbf{Method} & \rotatebox{90}{Aircraft} & \rotatebox{90}{Caltech101} & \rotatebox{90}{CIFAR100} & \rotatebox{90}{DTD} & \rotatebox{90}{EuroSAT} & \rotatebox{90}{Flowers} & \rotatebox{90}{Food} & \rotatebox{90}{MNIST} & \rotatebox{90}{OxfordPet} & \rotatebox{90}{Cars} & \rotatebox{90}{SUN397} & \textbf{\textit{Average}} \\
\midrule
\rowcolor{Green!10}\multicolumn{13}{l}{\emph{CLIP zero-shot reference}}\\
{\quad}Zero-shot \citep{clip} & 24.3 & 88.4 & 68.2 & 44.6 & 54.9 & 71.0 & 88.5 & 59.4 & 89.0 & 64.7 & 65.2 & 65.2 \\
\midrule
\rowcolor{Green!10}\multicolumn{13}{l}{\emph{Transfer}}\\
{\quad}Zero-shot \citep{clip} & -- & 88.4 & 68.2 & 44.6 & 54.9 & 71.0 & 88.5 & 59.4 & 89.0 & 64.7 & 65.2 & 69.4 \\
\multicolumn{13}{l}{\colorbox{FigComp}{\textbf{\textit{\small Model complementation}}}} \\
{\quad}L2P$\dagger$ \citep{l2p} & -- & 55.7 & 50.9 & 30.4 & 41.4 & 49.3 & 71.8 & 36.3 & 77.5 & 55.3 & 53.4 & 53.2 \\
{\quad}DualPrompt$\dagger$ \citep{dualprompt} & -- & 66.7 & 51.4 & 28.7 & 33.7 & 45.6 & 70.9 & 59.5 & 77.7 & 49.5 & 50.4 & 52.4 \\
{\quad}S-Prompts$\dagger$ \citep{sprompts} & -- & 67.3 & 49.4 & 26.4 & 39.7 & 47.1 & 70.2 & 34.3 & 78.9 & 56.7 & 52.2 & 52.2 \\
{\quad}MoE-Adapter$\dagger$ \citep{boosting} & -- & 87.9 & 68.2 & 44.4 & 49.9 & \textbf{70.7} & \textbf{88.7} & 59.7 & \textbf{89.1} & 64.5 & 65.5 & 68.9 \\
{\quad}DIKI$\dagger$ \citep{tang2025mind} & -- & \textbf{92.9} & \textbf{69.0} & 43.2 & 48.2 & 67.4 & 85.2 & \textbf{63.0} & 87.9 & 63.8 & \textbf{66.2} & 68.7 \\
{\quad}\textbf{\ours$\dagger$} & -- & 92.4 & 68.8 & \textbf{45.2} & \textbf{50.0} & 69.4 & 84.2 & 62.3 & 88.8 & \textbf{64.6} & 65.0 & \textbf{69.1} \\
\midrule
\multicolumn{13}{l}{\colorbox{FigCont}{\textbf{\textit{\small Model continually learned}}}} \\
\multicolumn{13}{l}{\hspace{0.5em}\colorbox{FigRef}{\textit{\small reference-constrained}}} \\
{\quad}LwF \citep{li2017learning} & -- & 56.6 & 44.6 & 32.7 & 39.3 & 46.6 & 68.0 & 46.0 & 77.4 & 31.9 & 60.5 & 50.4 \\
{\quad}LwF-VR \citep{ding2022don} & -- & 73.5 & 55.6 & 35.6 & 41.5 & 47.0 & 68.3 & 53.9 & 69.3 & 26.8 & 51.9 & 52.3 \\
{\quad}WiSE-FT \citep{wortsman2022robust} & -- & 86.0 & 67.4 & \textbf{45.4} & \textbf{50.4} & 69.1 & \textbf{87.6} & 61.8 & 86.8 & 60.1 & \textbf{66.8} & 68.1 \\
{\quad}iCaRL \citep{rebuffi2017icarl} & -- & 77.1 & 61.0 & 40.5 & 45.3 & 54.4 & 74.6 & 47.9 & 76.7 & 36.3 & 58.6 & 57.2 \\
{\quad}ZSCL \citep{zscl} & -- & 55.7 & 50.9 & 30.4 & 41.4 & 49.3 & 71.8 & 36.3 & 77.5 & 55.3 & 53.4 & 53.2 \\
\multicolumn{13}{l}{\hspace{0.5em}\colorbox{FigCap}{\textit{\small capacity-controlled (ours)}}} \\
{\quad}\textbf{\ours} & -- & \textbf{92.4} & \textbf{68.8} & 45.2 & 50.0 & \textbf{69.4} & 84.2 & \textbf{62.3} & \textbf{88.8} & \textbf{64.6} & 65.0 & \textbf{69.1} \\
\midrule
\rowcolor{Green!10}\multicolumn{13}{l}{\emph{Average}}\\
\multicolumn{13}{l}{\colorbox{FigComp}{\textbf{\textit{\small Model complementation}}}} \\
{\quad}L2P$\dagger$ \citep{l2p} & 38.0 & 85.2 & 78.2 & 61.3 & 72.9 & 74.9 & 79.7 & 59.1 & 82.0 & 59.7 & 55.4 & 67.9 \\
{\quad}DualPrompt$\dagger$ \citep{dualprompt} & 37.8 & 84.3 & 78.5 & 60.1 & 71.1 & 73.2 & 79.1 & 73.9 & 82.3 & 55.1 & 52.8 & 68.0 \\
{\quad}S-Prompts$\dagger$ \citep{sprompts} & 37.5 & 92.5 & 77.5 & 58.2 & 76.4 & 74.1 & 78.8 & 57.9 & 83.0 & 60.8 & 54.4 & 68.3 \\
{\quad}MoE-Adapter$\dagger$ \citep{boosting} & \textbf{50.2} & 91.9 & 83.1 & \textbf{69.4} & 78.9 & \textbf{84.0} & \textbf{89.1} & 73.7 & 89.3 & \textbf{67.7} & 66.9 & 76.7 \\
{\quad}DIKI$\dagger$ \citep{tang2025mind} & 45.1 & 95.5 & 83.1 & 64.8 & 79.9 & 83.5 & 87.0 & \textbf{76.2} & 89.6 & 67.0 & \textbf{67.1} & 76.3 \\
{\quad}\textbf{\ours$\dagger$} & 48.3 & \textbf{96.4} & \textbf{83.2} & 69.1 & \textbf{80.0} & \textbf{84.0} & 86.0 & 75.6 & \textbf{90.5} & 67.5 & 66.3 & \textbf{77.0} \\
\midrule
\multicolumn{13}{l}{\colorbox{FigCont}{\textbf{\textit{\small Model continually learned}}}} \\
\multicolumn{13}{l}{\hspace{0.5em}\colorbox{FigRef}{\textit{\small reference-constrained}}} \\
{\quad}LwF \citep{li2017learning} & 36.3 & 86.9 & 72.0 & 59.0 & 73.7 & 60.0 & 73.6 & 74.8 & 80.0 & 37.3 & 58.1 & 64.7 \\
{\quad}LwF-VR \citep{ding2022don} & 29.6 & 87.7 & 74.4 & 59.5 & 72.4 & 63.6 & 77.0 & 66.7 & 81.2 & 43.7 & 60.7 & 65.1 \\
{\quad}WiSE-FT \citep{wortsman2022robust} & 26.7 & 86.5 & 64.4 & 57.1 & 65.7 & 58.7 & 71.1 & 70.5 & 75.8 & 36.9 & 54.6 & 60.7 \\
{\quad}iCaRL \citep{rebuffi2017icarl} & 35.5 & 89.2 & 72.2 & 60.6 & 68.8 & 70.0 & 78.2 & 62.3 & 81.8 & 41.2 & 62.5 & 65.7 \\
{\quad}ZSCL \citep{zscl} & 45.1 & 92.0 & 80.1 & 64.3 & \textbf{79.5} & 81.6 & \textbf{89.6} & \textbf{75.2} & 88.9 & 64.7 & \textbf{68.0} & 75.4 \\
\multicolumn{13}{l}{\hspace{0.5em}\colorbox{FigCap}{\textit{\small capacity-controlled (ours)}}} \\
{\quad}\textbf{\ours} & \textbf{47.5} & \textbf{95.3} & \textbf{82.1} & \textbf{68.2} & 79.0 & \textbf{83.1} & 85.0 & 74.6 & \textbf{89.6} & \textbf{66.5} & 65.2 & \textbf{76.0} \\
\midrule
\rowcolor{Green!10}\multicolumn{13}{l}{\emph{Last}}\\
\multicolumn{13}{l}{\colorbox{FigComp}{\textbf{\textit{\small Model complementation}}}} \\
{\quad}L2P$\dagger$ \citep{l2p} & 38.0 & 87.1 & 84.2 & 72.9 & 86.0 & 96.1 & 89.2 & 99.0 & 94.1 & 79.6 & 76.0 & 82.0 \\
{\quad}DualPrompt$\dagger$ \citep{dualprompt} & 37.8 & 87.1 & 84.6 & 71.8 & 89.2 & 96.3 & 89.1 & 99.1 & 94.5 & 79.9 & 76.5 & 82.3 \\
{\quad}S-Prompts$\dagger$ \citep{sprompts} & 37.5 & 95.1 & 83.7 & 70.2 & 97.5 & 96.5 & 89.0 & 99.1 & 94.0 & 79.5 & 75.8 & 83.4 \\
{\quad}MoE-Adapter$\dagger$ \citep{boosting} & \textbf{49.8} & 92.2 & 86.1 & \textbf{78.1} & 95.7 & 94.3 & \textbf{89.5} & 98.1 & 89.9 & \textbf{81.6} & \textbf{80.0} & 85.0 \\
{\quad}DIKI$\dagger$ \citep{tang2025mind} & 45.2 & 95.7 & \textbf{86.3} & 72.9 & \textbf{98.0} & \textbf{97.0} & 89.2 & \textbf{99.4} & 94.2 & \textbf{81.6} & 76.6 & 85.1 \\
{\quad}\textbf{\ours$\dagger$} & 48.3 & \textbf{96.8} & 85.9 & 77.8 & 97.5 & 96.2 & 88.3 & 99.0 & \textbf{94.9} & 80.6 & 78.6 & \textbf{85.8} \\
\midrule
\multicolumn{13}{l}{\colorbox{FigCont}{\textbf{\textit{\small Model continually learned}}}} \\
\multicolumn{13}{l}{\hspace{0.5em}\colorbox{FigRef}{\textit{\small reference-constrained}}} \\
{\quad}LwF \citep{li2017learning} & 26.3 & 87.5 & 71.9 & 66.6 & 79.9 & 66.9 & 83.8 & \textbf{99.6} & 92.1 & 66.1 & 80.4 & 74.6 \\
{\quad}LwF-VR \citep{ding2022don} & 20.5 & 89.8 & 72.3 & 67.6 & 85.5 & 73.8 & 85.7 & \textbf{99.6} & 93.1 & 73.3 & 80.9 & 76.6 \\
{\quad}WiSE-FT \citep{wortsman2022robust} & 27.2 & 90.8 & 68.0 & 68.9 & 86.9 & 74.0 & 87.6 & \textbf{99.6} & 92.6 & 77.8 & 81.3 & 77.7 \\
{\quad}iCaRL \citep{rebuffi2017icarl} & 35.8 & 93.0 & 77.0 & 70.2 & 83.3 & 88.5 & 90.4 & 86.7 & 93.2 & 81.2 & \textbf{81.9} & 80.1 \\
{\quad}ZSCL \citep{zscl} & 40.6 & 92.2 & 81.3 & 70.5 & 94.8 & 90.5 & \textbf{91.9} & 98.7 & 93.9 & \textbf{85.3} & 80.2 & 83.6 \\
\multicolumn{13}{l}{\hspace{0.5em}\colorbox{FigCap}{\textit{\small capacity-controlled (ours)}}} \\
{\quad}\textbf{\ours} & \textbf{47.6} & \textbf{95.7} & \textbf{84.9} & \textbf{76.8} & \textbf{96.5} & \textbf{95.2} & 87.5 & 98.1 & \textbf{94.0} & 79.6 & 77.6 & \textbf{84.9} \\
\bottomrule
\end{tabular}}
\caption{Full-shot MTIL. Rows within each subgroup are sorted by Last (ascending). \textbf{Bold} marks the best per column within each big group (model complementation vs.\ model continually learned). $\dagger$: uses domain prediction at inference similar to \citep{boosting}; plain CoDyRA merges the LoRA into the backbone with no auxiliary module.}
\label{supp:full_mtil}
\end{table*}
\subsection{Effects of Additional Adaptive Prediction Head on Visual Representations}
The related continual learning methods focus on updating the representation with the backbone model (\textit{e.g.}, \citep{l2p,zscl,boosting}, and \ours) or newly added representation alignment or prediction head \citep{mcdonnell2024ranpac,rail}. 
These two types of methods can be seen as orthogonal. We study how the representation updated by the proposed method can work with the additional adaptive prediction head, relying on the aggregation-based adapters in \citep{rail}. 

We provide a more detailed analysis of the effects of the additional prediction head on the visual representations in X-TAIL setting in Table \ref{tab:supp_ours+rail}. Note that the Dual version of RAIL \citep{rail} stores all visual representations of all samples for all seen classes and domains in memory, it is not included in Tables \ref{tab:few_xtail} and \ref{tab:few_mtil} for a fair comparison. 

RAIL is a regression-based method that applies an additional prediction head on top of the visual representations extracted by the frozen pre-trained vision encoder of the CLIP model.
This approach is fully orthogonal to our method, as our focus is on continually incorporating new knowledge into the pre-trained model through dynamic low-rank parameter updates, and we can seamlessly integrate their techniques into our method. Because RAIL heavily relies on the frozen pre-trained model, the Transfer accuracy is limited by the zero-shot performance of the pre-trained model. In contrast, our approach gradually accumulates knowledge into the pre-trained model, enhancing its capability on unseen data and achieving improved Transfer accuracy. Moreover, by leveraging our continually enhanced pre-trained model, coupling our method with RAIL significantly outperforms the original RAIL, which uses a frozen pre-trained model, on the Average and Last metrics.

\begin{table}[!t]
\centering
\small
\begin{tabular}{@{}lccc@{}}
\toprule
 & Transfer & Average & Last \\ \midrule
RAIL-Primal & 62.4 & 70.7 & 79.1 \\
\textbf{\ours} + RAIL-Primal & \textbf{63.1} & \textbf{74.1} & \textbf{83.4} \\ \midrule
RAIL-Dual & 62.4 & 71.9 & 82.4 \\
\textbf{\ours} + RAIL-Dual & \textbf{63.1} & \textbf{74.2} & \textbf{83.6} \\ \bottomrule
\end{tabular}
\caption{Experimental results for integrating aggregation-based adapters in the X-TAIL setting. Metrics for ``Transfer'', ``Average'', and ``Last'' are reported as averages across all datasets.}
\label{tab:supp_ours+rail}
\end{table}

\subsection{Evaluation with Other VLMs}
To evaluate the broader applicability of our method, we conducted preliminary experiments with BLIP \citep{li2022blip} on X-TAIL (Table \ref{tab:blip}). Our method modestly improves upon RAIL-Primal and exceeds the zero-shot pre-trained model, suggesting applicability across different VLMs.

\subsection{Evaluation on Standard Parameter-Efficient Fine-Tuning (PEFT) Task}
Leveraging our dynamic rank-adaptive parameter updates, our method naturally extends to Parameter-Efficient Fine-Tuning (PEFT) tasks. Table~\ref{tab:peft} presents 16-shot PEFT results, using the same hyperparameters as the main paper, showing that \ours\ outperforms vanilla LoRA while reducing final trainable parameters by pruning ranks with zero importance. Notably, across multiple downstream tasks (Caltech101, Cars, and OxfordPet), \ours\ achieves higher accuracy while using fewer parameters compared to LoRA with a fixed rank of 16.

By automatically selecting the most relevant ranks from training data, our method provides a more adaptive and efficient alternative to conventional low-rank tuning. It improves downstream accuracy while reducing the final parameter count.

\reva{\section{Analyses and Diagnostics}}

\reva{\subsection{Multi-Dataset Validation of the Findings}
\label{supp:multi_dataset}

We repeat the rank/placement probe of Sec.~\ref{sec:motivation} across three datasets (DTD, EuroSAT, Oxford Flowers) with three ranks ($r{\in}\{4, 8, 16\}$), five placements (attn, mlp, vision, text, all), and three seeds. Tables~\ref{tab:supp_dtd}--\ref{tab:supp_flowers} report $\Delta$Plasticity (higher better) and Forgetting (lower better) as mean$\pm$std.

\begin{table}[!ht]
\centering\small
\resizebox{\linewidth}{!}{
\begin{tabular}{lcccccc}
\toprule
Placement & $\Delta$P $r4$ & $\Delta$P $r8$ & $\Delta$P $r16$ & Fgt $r4$ & Fgt $r8$ & Fgt $r16$ \\ \midrule
attn   & 30.3$\pm$0.5 & 30.8$\pm$1.0 & 31.1$\pm$0.5 & 2.25$\pm$0.07 & 3.02$\pm$0.79 & 4.41$\pm$0.49 \\
mlp    & 30.1$\pm$0.5 & 30.3$\pm$0.8 & 31.0$\pm$0.5 & 5.50$\pm$0.26 & 6.28$\pm$0.46 & 6.71$\pm$1.04 \\
vision & 29.6$\pm$0.6 & 29.9$\pm$0.1 & 30.2$\pm$0.5 & 14.78$\pm$0.73 & 15.45$\pm$0.78 & 18.43$\pm$1.09 \\
text   & 27.1$\pm$0.5 & 26.9$\pm$0.5 & 26.9$\pm$0.4 & 3.52$\pm$0.59 & 2.04$\pm$0.43 & 1.98$\pm$0.39 \\
all    & 30.7$\pm$1.1 & 31.3$\pm$0.5 & 31.0$\pm$1.2 & 5.04$\pm$0.65 & 6.36$\pm$0.10 & 8.24$\pm$1.26 \\ \bottomrule
\end{tabular}}
\caption{DTD: $\Delta$Plasticity (higher is better) and Forgetting (lower is better) across placements and ranks (3 seeds).}
\label{tab:supp_dtd}
\end{table}

\begin{table}[!ht]
\centering\small
\resizebox{\linewidth}{!}{
\begin{tabular}{lcccccc}
\toprule
Placement & $\Delta$P $r4$ & $\Delta$P $r8$ & $\Delta$P $r16$ & Fgt $r4$ & Fgt $r8$ & Fgt $r16$ \\ \midrule
attn   & 46.8$\pm$0.3 & 47.1$\pm$0.4 & 47.3$\pm$0.2 & $-$0.13$\pm$0.14 & $-$0.52$\pm$0.34 & $-$0.18$\pm$0.20 \\
mlp    & 47.3$\pm$0.3 & 46.9$\pm$0.4 & 47.1$\pm$0.4 & 2.46$\pm$0.48 & 4.67$\pm$1.38 & 2.99$\pm$1.88 \\
vision & 47.6$\pm$0.1 & 47.6$\pm$0.4 & 47.7$\pm$0.6 & 4.24$\pm$0.76 & 4.32$\pm$1.10 & 6.95$\pm$0.55 \\
text   & 40.8$\pm$0.5 & 40.7$\pm$0.5 & 40.6$\pm$0.2 & 1.63$\pm$0.02 & 1.05$\pm$0.04 & 0.21$\pm$0.48 \\
all    & 47.0$\pm$0.1 & 47.1$\pm$0.3 & 47.4$\pm$0.4 & 0.52$\pm$0.14 & 0.78$\pm$1.28 & 0.81$\pm$0.28 \\ \bottomrule
\end{tabular}}
\caption{EuroSAT: $\Delta$Plasticity (higher is better) and Forgetting (lower is better) across placements and ranks (3 seeds).}
\label{tab:supp_eurosat}
\end{table}

\begin{table}[!ht]
\centering\small
\resizebox{\linewidth}{!}{
\begin{tabular}{lcccccc}
\toprule
Placement & $\Delta$P $r4$ & $\Delta$P $r8$ & $\Delta$P $r16$ & Fgt $r4$ & Fgt $r8$ & Fgt $r16$ \\ \midrule
attn   & 26.6$\pm$0.5 & 26.6$\pm$0.2 & 27.1$\pm$0.4 & $-$0.41$\pm$0.23 & $-$0.70$\pm$0.40 & $-$0.81$\pm$0.66 \\
mlp    & 26.7$\pm$0.2 & 27.0$\pm$0.4 & 27.1$\pm$0.3 & $-$1.21$\pm$0.17 & $-$0.97$\pm$0.25 & $-$1.07$\pm$0.27 \\
vision & 26.1$\pm$0.3 & 26.4$\pm$0.3 & 26.2$\pm$0.2 & 0.52$\pm$0.21 & 1.21$\pm$0.28 & 2.06$\pm$0.44 \\
text   & 25.9$\pm$0.5 & 26.2$\pm$0.2 & 26.4$\pm$0.1 & 0.47$\pm$0.17 & 0.33$\pm$0.27 & $-$0.08$\pm$0.52 \\
all    & 27.1$\pm$0.1 & 27.1$\pm$0.1 & 27.2$\pm$0.3 & $-$0.89$\pm$0.49 & $-$1.02$\pm$0.40 & $-$0.64$\pm$0.54 \\ \bottomrule
\end{tabular}}
\caption{Oxford Flowers: $\Delta$Plasticity (higher is better) and Forgetting (lower is better) across placements and ranks (3 seeds).}
\label{tab:supp_flowers}
\end{table}

Two patterns hold across all three datasets. (i) Vision-encoder updates forget most on every dataset, while the least-forgetting placement varies (text on DTD, attention on EuroSAT, MLP on Flowers). (ii) The rank effect varies by placement: forgetting can rise, stay flat, or fall as $r$ grows. Together, these patterns show that the best balance depends jointly on placement, rank, and dataset, motivating adaptive per-matrix selection.}

\reva{\subsection{Gradient-Conflict and Norm-Matched Probes}
\label{supp:norm_matched}

\noindent\textbf{Gradient-conflict at truncated rank.} At the pretrained CLIP, we SVD-truncate each task's loss-gradient to rank $\rho$ and measure the norm-normalized mean pairwise $|\cos|$ between tasks, a first-order diagnostic for the interference mechanism (Table~\ref{tab:supp_grad_conflict}).

\begin{table}[!ht]\centering\small
\begin{tabular}{lc}\toprule
rank $\rho$ & mean $|\cos|$ (lower is better) \\ \midrule
1    & 0.0323 \\
8    & 0.0296 \\
full & 0.0252 \\ \bottomrule
\end{tabular}
\caption{Norm-normalized inter-task gradient conflict after SVD truncation to rank $\rho$. Conflict is highest at extreme low rank: the top singular directions are the most shared across tasks.}
\label{tab:supp_grad_conflict}
\end{table}

\noindent\textbf{Norm-matched probe.} To isolate rank from magnitude, we retrain each rank with a penalty enforcing a common $\|\Delta\mathbf{W}\|_F \approx 3.8$ (median of the ranks' natural norms) and measure single-task plasticity and forgetting (3 seeds; Table~\ref{tab:supp_norm_matched}).

\begin{table}[!ht]\centering\small
\resizebox{\linewidth}{!}{
\begin{tabular}{lcccc}\toprule
update & eff-rank & $\|\Delta\mathbf{W}\|_F$ & $\Delta$plasticity & Forgetting \\ \midrule
LoRA $r{=}1$  & 1.0  & 3.83$\pm$0.07 & +22.37 & +0.99$\pm$0.35 \\
LoRA $r{=}2$  & 2.0  & 3.73$\pm$0.03 & +22.93 & +0.36$\pm$0.04 \\
LoRA $r{=}4$  & 4.0  & 3.71$\pm$0.00 & +24.93 & $-$0.16$\pm$0.31 \\
LoRA $r{=}8$  & 8.0  & 3.79$\pm$0.02 & +24.87 & +0.20$\pm$0.08 \\
LoRA $r{=}16$ & 16.0 & 3.97$\pm$0.08 & +22.87 & +0.10$\pm$0.06 \\ \bottomrule
\end{tabular}}
\caption{Norm-matched probe: at fixed $\|\Delta\mathbf{W}\|_F \approx 3.8$, rank still matters. Rank-1 both learns least and forgets most; a moderate rank (4--8) reaches the best trade-off.}
\label{tab:supp_norm_matched}
\end{table}

At matched magnitude, rank still matters: rank-1 both learns least and forgets most (consistent with the gradient-conflict result, where extreme low rank concentrates the update on shared high-conflict directions), while a moderate rank (4--8) spreads the update onto less-conflicting directions. The interference mechanism thus persists after controlling for magnitude.}

\subsection{Cumulative Parameter Shift: Scaling and Constructive Analysis}
\label{supp:drift}

This section expands on Table~\ref{tab:drift} (main paper) along two complementary analyses: the asymptotic scaling of the cumulative shift and its constructive nature.

\noindent\textbf{Asymptotic scaling.} A least-squares fit of $d(t)=at^b$ to the 10-point trajectories yields $b{=}0.49$ ($R^2{=}0.99$) for CoDyRA and $b{=}0.95$ ($R^2{=}0.97$) for fixed-rank LoRA ($r{=}16$). Extrapolating to $t{=}20$ tasks gives projected shifts of $10.6$ and $96.8$, respectively. The sublinear growth of CoDyRA is consistent with the dimensional clause of Prop.~\ref{prop:forget}: pruning an importance weight to zero eliminates further perturbation from the corresponding rank-one component, so the per-task incremental shift decreases as more directions are pruned.

\noindent\textbf{Constructive nature of the accumulated shift.} We characterize how the cumulative update affects unseen-domain capability through two probes. Table~\ref{tab:supp_zs_tracking} tracks zero-shot accuracy on held-out benchmarks after each task merge.

\begin{table}[!ht]
\centering\small
\begin{tabular}{lc}
\toprule
Model state          & Unseen zero-shot (avg.) \\ \midrule
Pretrained ($\theta_0$)   & 56.24 \\
After task 1         & 56.08 \\
After task 2         & 56.92 \\
After task 4         & 57.23 \\
After task 6         & 57.32 \\
After task 8         & 57.71 \\
After task 10        & 58.00 \\ \bottomrule
\end{tabular}
\caption{Zero-shot accuracy on unseen datasets after each CoDyRA merge.}
\label{tab:supp_zs_tracking}
\end{table}

Table~\ref{tab:supp_interp} reports linear interpolation between $\theta_0$ and the final $\theta_{10}$ via $\theta_\alpha = (1-\alpha)\theta_0 + \alpha\,\theta_{10}$, evaluated on the same benchmarks.

\begin{table}[!ht]
\centering\small
\begin{tabular}{lc}
\toprule
$\alpha$ (fraction of merged update) & Unseen zero-shot (avg.) \\ \midrule
0.00 (pretrained)    & 56.24 \\
0.25                 & 56.21 \\
0.50                 & 56.58 \\
0.75                 & 57.84 \\
1.00 (CoDyRA $\theta_{10}$) & 58.00 \\ \bottomrule
\end{tabular}
\caption{Linear interpolation $\theta_\alpha = (1-\alpha)\theta_0 + \alpha\,\theta_{10}$ on unseen-domain zero-shot accuracy.}
\label{tab:supp_interp}
\end{table}

Both trajectories are near-monotone (with a small early dip while initial-task knowledge accumulates), indicating that a larger fraction of the merged update yields stronger zero-shot generalization on held-out data. The accumulated shift therefore acts constructively on the pretrained representation rather than degrading it, a behavior atypical of standard fine-tuning under sustained training.

\subsection{Inter-Task CKA: Empirical Probe for the Dimensional Bound}
\label{supp:cka}

Prop.~\ref{prop:forget} bounds the row and column dimensions of a rank-$\rho$ update by $\rho$; when CoDyRA drives $\rho$ to small values, consecutive task updates should therefore occupy small subspaces with limited overlap. To test this prediction, we compute linear CKA between consecutive task updates on the X-TAIL stream.

For each task pair $(t, t{+}1)$, we flatten the merged updates $\Delta\mathbf{W}^{(t)}$ and $\Delta\mathbf{W}^{(t+1)}$ into vectors $v_t, v_{t+1} \in \mathbb{R}^D$ across all updated weight matrices, then compute linear CKA (Table~\ref{tab:supp_cka}). Values near 1 indicate aligned updates with substantial overlap; values near 0 indicate near-orthogonal updates with minimal interference.

\begin{table}[!ht]
\centering\small
\begin{tabular}{lcc}
\toprule
Task pair (consecutive) & CoDyRA & LoRA ($r{=}16$) \\ \midrule
1--2  & 0.0041 & 0.0113 \\
2--3  & 0.0039 & 0.0116 \\
3--4  & 0.0036 & 0.0101 \\
4--5  & 0.0037 & 0.0105 \\
5--6  & 0.0039 & 0.0133 \\
6--7  & 0.0038 & 0.0092 \\
7--8  & 0.0036 & 0.0060 \\
8--9  & 0.0038 & 0.0059 \\
9--10 & 0.0037 & 0.0071 \\ \midrule
\textbf{Mean} & \textbf{0.0038} & 0.0094 \\ \bottomrule
\end{tabular}
\caption{Linear CKA between consecutive task updates on X-TAIL. Lower values indicate less interfering updates.}
\label{tab:supp_cka}
\end{table}

CoDyRA's mean inter-task CKA ($0.0038$) is smaller than that of fixed-rank LoRA ($0.0094$), providing direct empirical support for the dimensional clause of Prop.~\ref{prop:forget}: by confining each task's update to a small subspace, CoDyRA preserves orthogonal capacity for subsequent tasks.

\subsection{Parameter-space Directional Analysis}
\begin{figure*}[!t]
\centering
\includegraphics[width=0.8\linewidth]{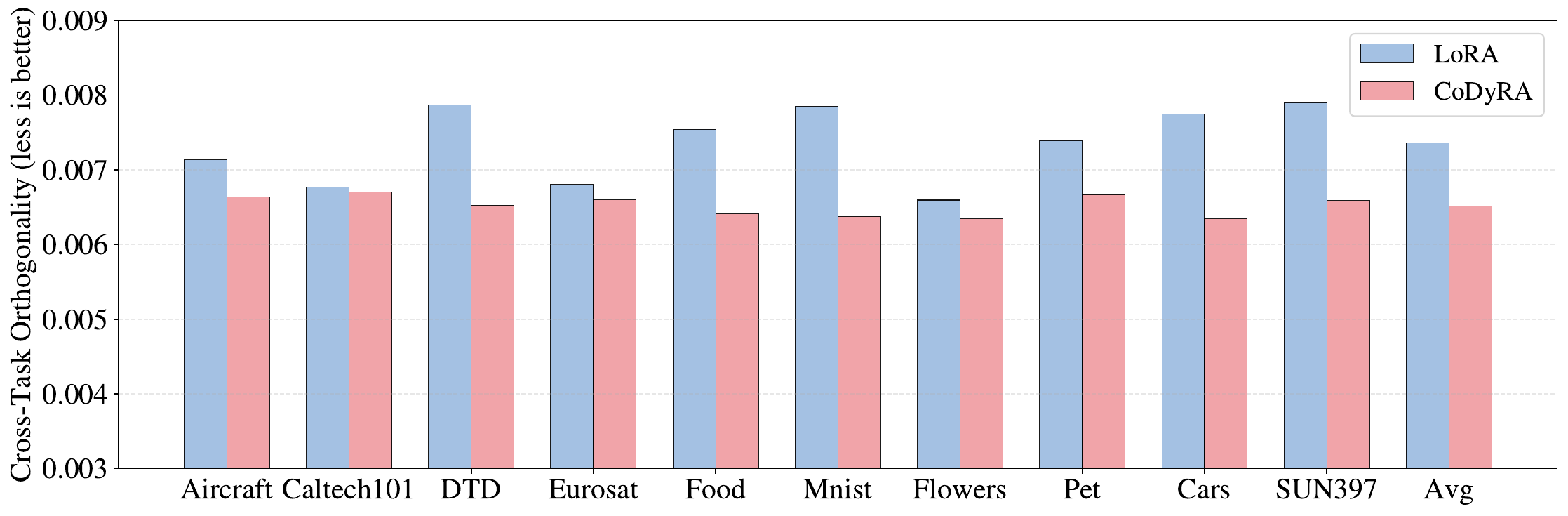}\
\caption{{Parameter-space directional analysis of $\mathbf{A}$-matrix updates. For each task, the column displays the average cosine similarity between the model’s LoRA updates after a specific task and the LoRA updates from models trained on all other tasks. This analysis quantifies directional interference among sequential updates. Lower cosine similarity reflects more orthogonal (less interfering) updates. CoDyRA produces lower similarity than the simple LoRA–merging baseline, demonstrating stronger implicit orthogonality of parameter updates even without explicit orthogonal regularization.}
}
\label{fig:orth}
\vspace{-0.5cm}
\end{figure*}

{To further investigate how CoDyRA structures its parameter updates across sequential tasks, we conduct a parameter-space directional analysis following prior work on directional interference in LoRA-based continual learning. Specifically, we compute the cosine similarity between the LoRA $\mathbf{A}$-matrix updates obtained after each task, comparing every task’s update direction against those from all other tasks. This metric captures the degree to which updates for different tasks align or interfere: higher similarity indicates that tasks induce overlapping update directions, while lower similarity reflects more orthogonal and thus less interfering modifications. As shown in Figure \ref{fig:orth}, CoDyRA consistently produces lower cross-task similarity than a baseline that trains a new LoRA adapter per task and merges them without additional constraints. Despite imposing no explicit orthogonality regularization, CoDyRA naturally encourages more decorrelated update directions, revealing an inherent structural property that helps mitigate destructive interference between tasks.}

\reva{\subsection{Forgetting Analysis on X-TAIL: Extended Discussion}\label{supp:bwt}}

\reva{The headline Forgetting comparison is in the main paper (Table~\ref{tab:bwt_main}). We expand here on the breakdown by baseline family: fixed-rank LoRA ($r{=}16$, Forgetting $14.86\%$) exhibits significant forgetting from unconstrained updates without regularization; ZSCL relies on replayed reference data; MoE-Adapter isolates task knowledge inside adapters with inference-time domain prediction; InfLoRA constrains updates to subspaces orthogonal to prior tasks. CoDyRA mitigates forgetting by minimizing the rank of each update, keeping the model close to its prior state without replay, isolation, or orthogonal constraints; the gap to fixed-rank LoRA isolates the contribution of rank minimization. The same trend appears in Forgetting on TRACE (Table~\ref{tab:llm_cl} of the main paper).}

\subsection{Analysis of Changes/Interference w.r.t. Pre-Trained Weights}
In Sec. \ref{sec:discussion} of the main paper, Fig. \ref{fig:aircraft_rank_stat} and Fig. \ref{fig:pets_rank_stat} visualize and analyze the dynamic ranks assigned to each module. Here, we extend the analysis by comparing parameter changes to pre-trained weights using the amplification factor as in \citep{lora}. The amplification factor is given by:
\begin{equation}
\text{Amp} = \frac{\|\Delta \mathbf{W}\|_F}{\|\mathbf{U}^\text{T} \mathbf{W} \mathbf{V}^\text
{T}\|_F},
\end{equation}
where $\mathbf{W}$ is the pre-trained weight matrix, and $\Delta \mathbf{W}$ is the parameter change introduced by LoRA. For vanilla LoRA and \ours, $\Delta \mathbf{W}$ is computed using Eq.~\ref{eq:lora} and Eq.~\ref{eq:dyra}, respectively. $\mathbf{U}$ and $\mathbf{V}$ represent the top $r$ singular vectors of $\text{SVD}(\Delta \mathbf{W})$, where $r$ is the number of interested ranks in analysis.

\begin{figure*}[!h]
    \centering
    \begin{subfigure}[b]{0.32\linewidth}
        \includegraphics[width=\linewidth]{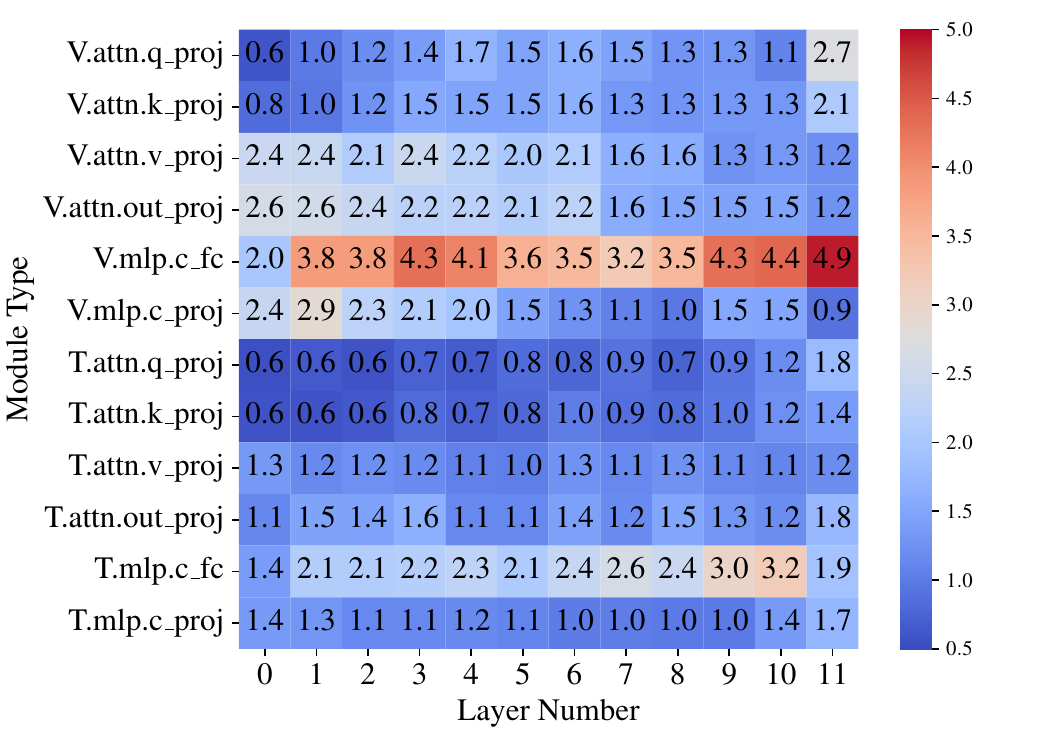}
        \caption{Aircraft - LoRA}
    \end{subfigure}
    \begin{subfigure}[b]{0.32\linewidth}
        \includegraphics[width=\linewidth]{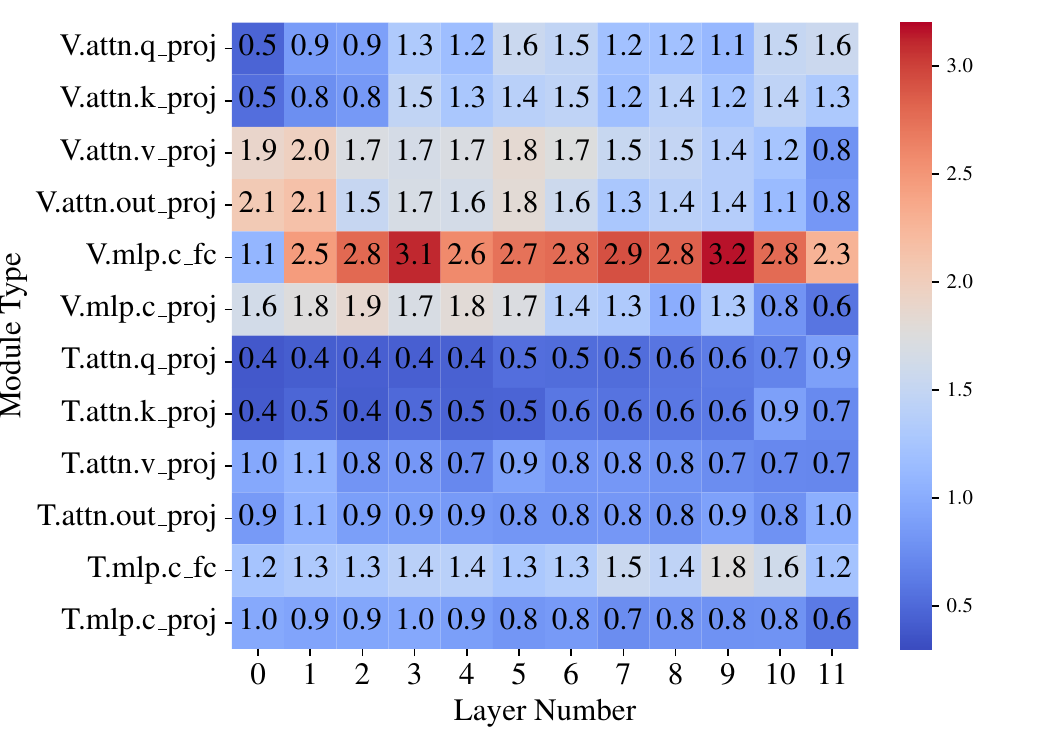}
        \caption{EuroSAT - LoRA}
    \end{subfigure}
    \begin{subfigure}[b]{0.32\linewidth}
        \includegraphics[width=\linewidth]{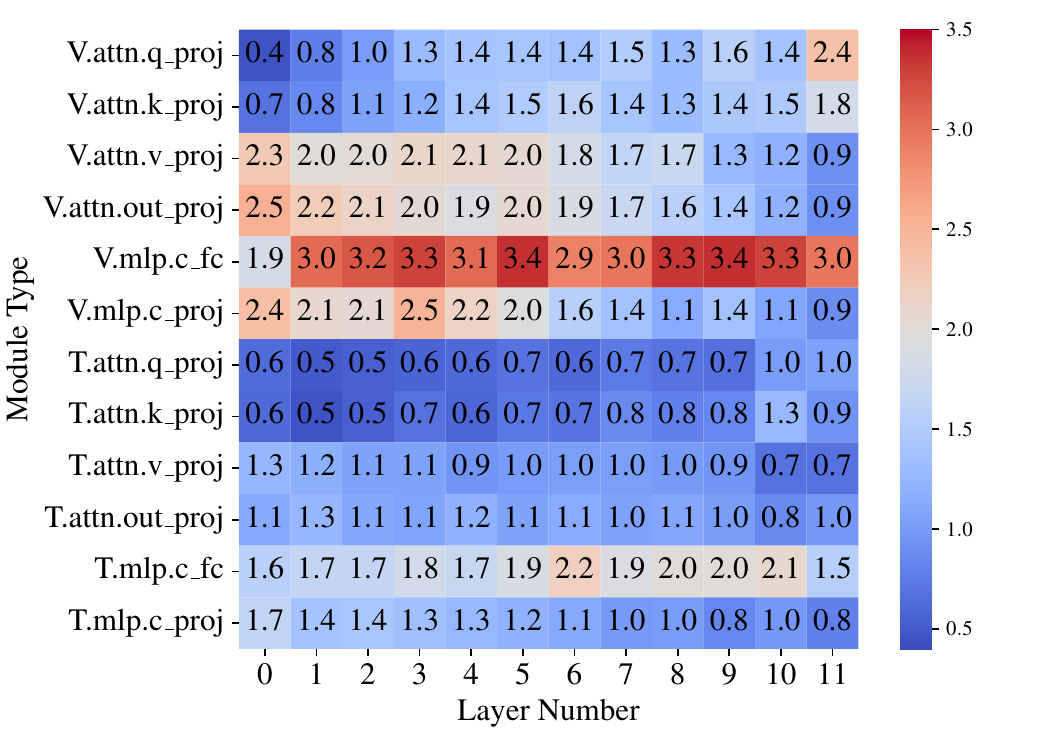}
        \caption{Pets - LoRA}
    \end{subfigure}
    \\
    \begin{subfigure}[b]{0.32\linewidth}
        \includegraphics[width=\linewidth]{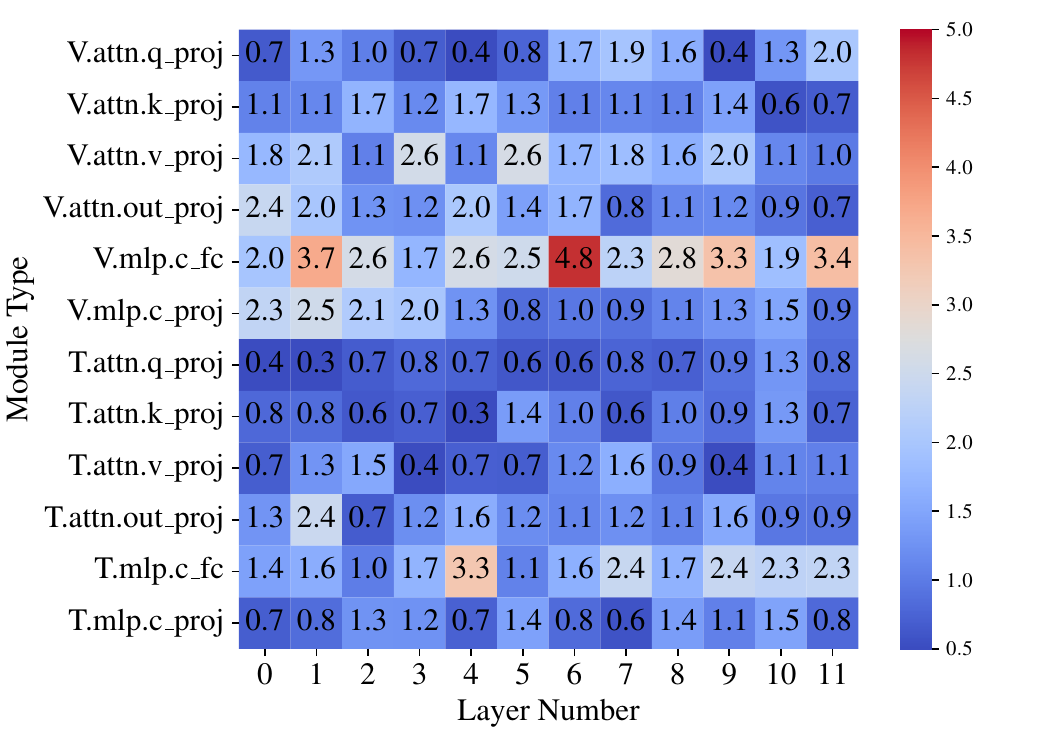}
        \caption{Aircraft - \ours}
    \end{subfigure}
    \begin{subfigure}[b]{0.32\linewidth}
        \includegraphics[width=\linewidth]{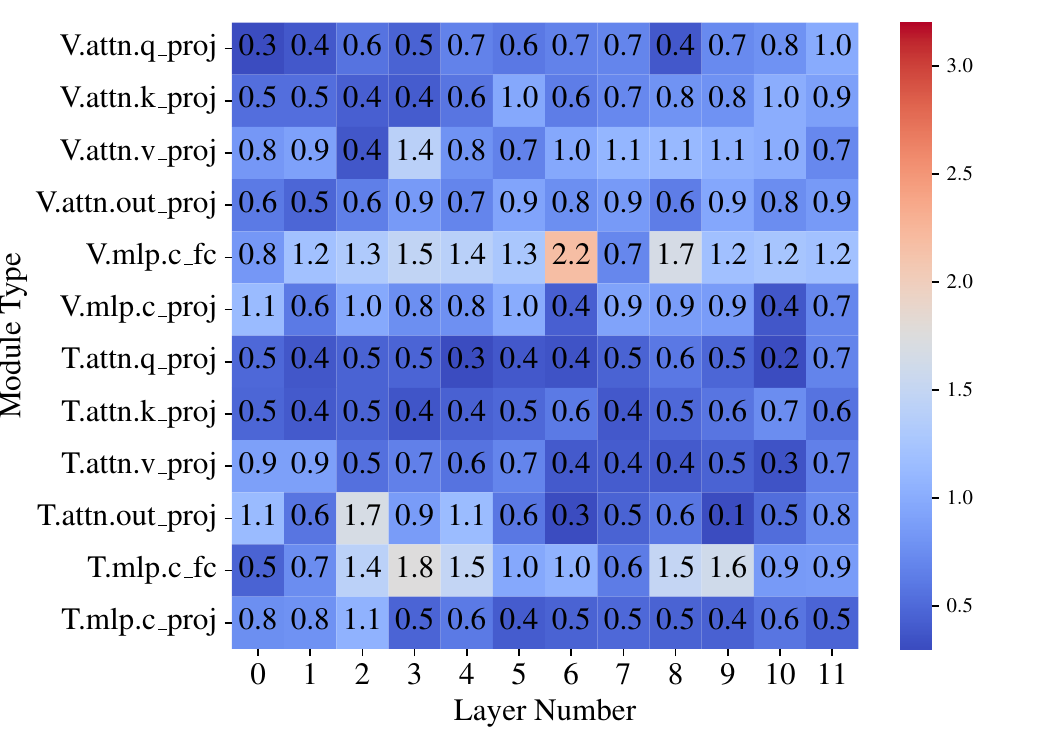}
        \caption{EuroSAT - \ours}
    \end{subfigure}
    \begin{subfigure}[b]{0.32\linewidth}
        \includegraphics[width=\linewidth]{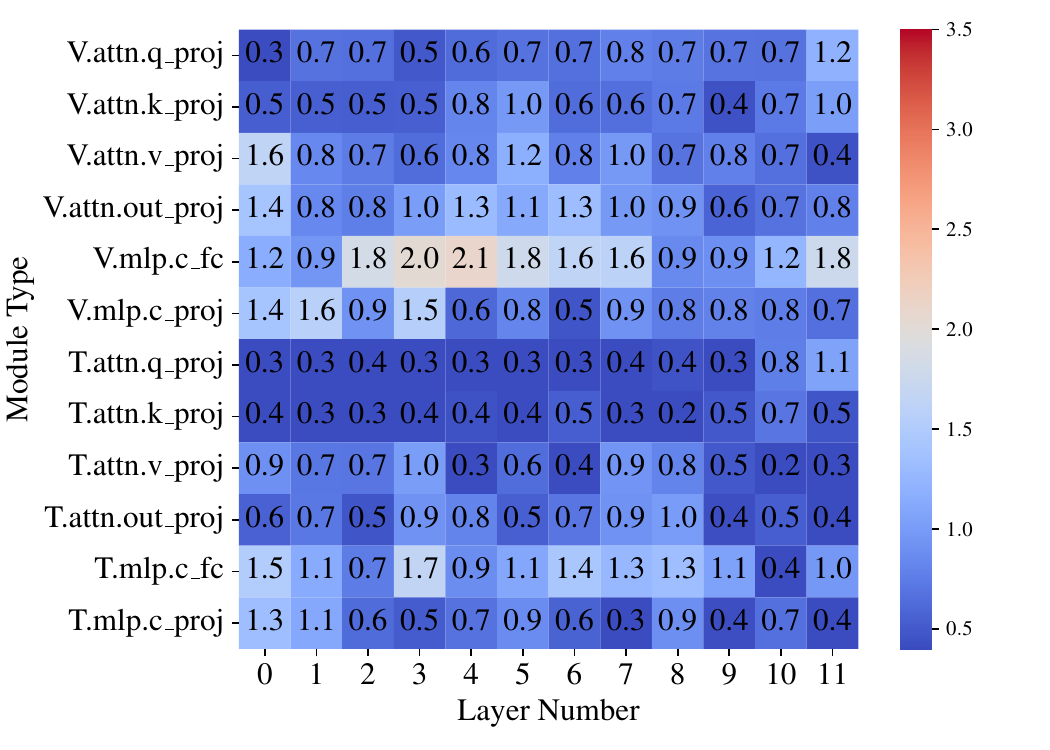}
        \caption{Pets - \ours}
    \end{subfigure}
    \caption{Visualization of amplification factors when trained on (a, d) Aircraft, (b, e) EuroSAT, and (c, f) Pets, comparing (d, e, f) our method of dynamic sparse rank selective LoRA and (a, b, c) vanilla LoRA. It shows that the proposed method with adaptive rank selection can achieve more focused/concentrated updating and less interference.}
    \label{fig:supp_amp}
\end{figure*}

A higher amplification factor indicates that low-rank parameter updates amplify directions less emphasized in the pre-trained weights \citep{lora}. We visualize the amplification factors for both our method and vanilla LoRA trained on the Aircraft, EuroSAT and Oxford Pets datasets in Fig. \ref{fig:supp_amp}. 

The results show that, compared to vanilla LoRA, our method tends to concentrate parameter updates on a more specific subset of transformer modules while reducing updates to pre-trained weights less relevant to the current data. This demonstrates that our method more effectively identifies critical pre-trained weight matrices relevant to the current data and applies targeted dynamic parameter updates. Additionally, by reducing the strength of parameter updates for pre-trained weights less related to the current data, our method minimizes the impact on pre-trained knowledge and knowledge acquired from previous tasks compared to vanilla LoRA.

Note that this analysis differs from the visualizations of number of activated ranks in Fig. \ref{fig:aircraft_rank_stat} and Fig. \ref{fig:pets_rank_stat}. The number of activated ranks reflects the quantity of significant ranks contributing to parameter updates for learning each task. In this analysis, the amplification factor directly measures the parameter change to each pre-trained weight matrix.

\subsection{Parameter-Matched Placement Ablation}
\label{supp:placement_matched}

To check whether the advantage of all-module placement (Sec.~\ref{sec:motivation} Takeaway 1) comes from a larger parameter budget or from cross-module rank allocation, we re-run the placement ablation with rank tuned so each setting matches the trainable-parameter budget of All ($r{=}16$); results in Table~\ref{tab:supp_placement_matched}.

\begin{table}[!ht]
\centering\small
\begin{tabular}{lcccccc}
\toprule
Placement & Rank & Params & Transfer & Avg. & Last \\ \midrule
Attn-only & 16   & 1.97M & 63.5 & 70.7 & 77.4 \\
Attn-only & 36   & 4.43M & 63.2 & 70.8 & 77.9 \\
MLP-only  & 16   & 2.46M & 63.0 & 69.4 & 76.7 \\
MLP-only  & 29   & 4.46M & 62.4 & 68.3 & 75.5 \\
\rowcolor{Purple!8}
\textbf{All (ours)} & \textbf{16} & \textbf{4.43M} & \textbf{63.2} & \textbf{71.3} & \textbf{79.2} \\ \bottomrule
\end{tabular}
\caption{Parameter-matched placement ablation on X-TAIL. Rank is tuned so each placement matches the trainable-parameter budget of \emph{All} ($r{=}16$).}
\label{tab:supp_placement_matched}
\end{table}

Observations: (i) Attn ($r{=}36$) shows negligible improvement over Attn ($r{=}16$), consistent with our Sec.~\ref{sec:motivation} finding that attention modules are stability-oriented with diminishing returns from extra capacity. (ii) MLP ($r{=}29$) actually performs worse than MLP ($r{=}16$), confirming that higher-rank MLP updates are more prone to forgetting. (iii) All ($r{=}16$) outperforms both parameter-matched alternatives on Average and Last, demonstrating that the advantage stems from distributing capacity across module types with adaptive pruning rather than from a larger overall budget.

\subsection{More Visualizations and Analyses on Rank allocation results of different datasets}
\label{supp:vis_heatmap}
\begin{figure*}[h]
    \begin{minipage}[t]{0.45\textwidth}
        \centering
        \includegraphics[width=\textwidth]{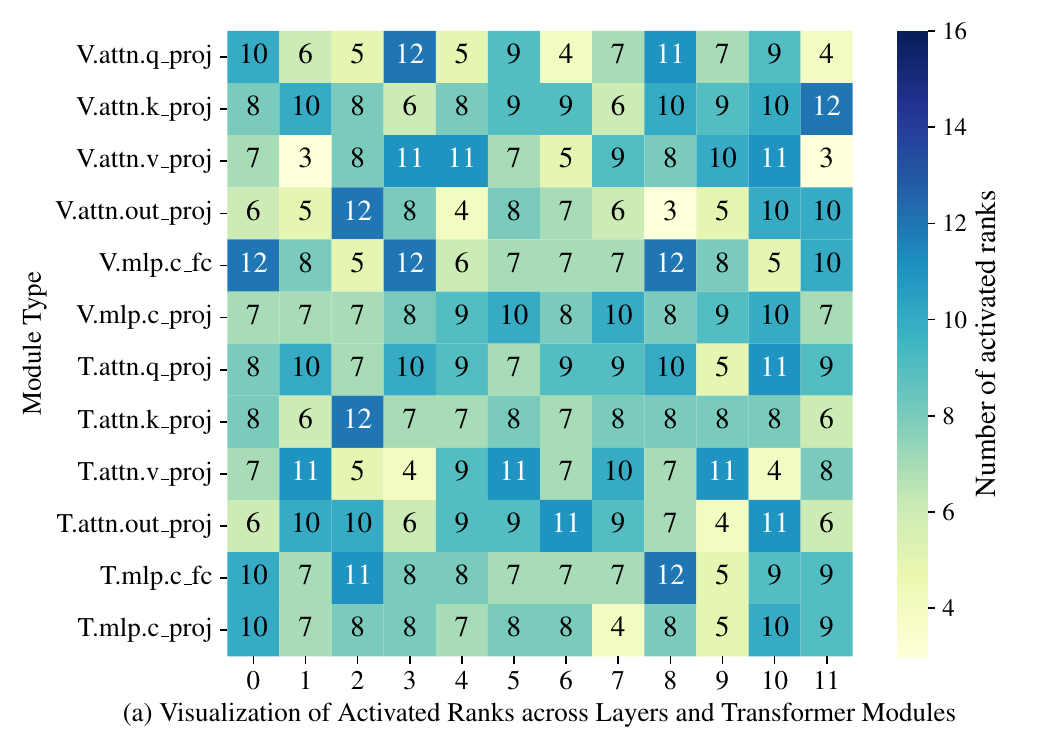}\\
        \includegraphics[width=\textwidth]{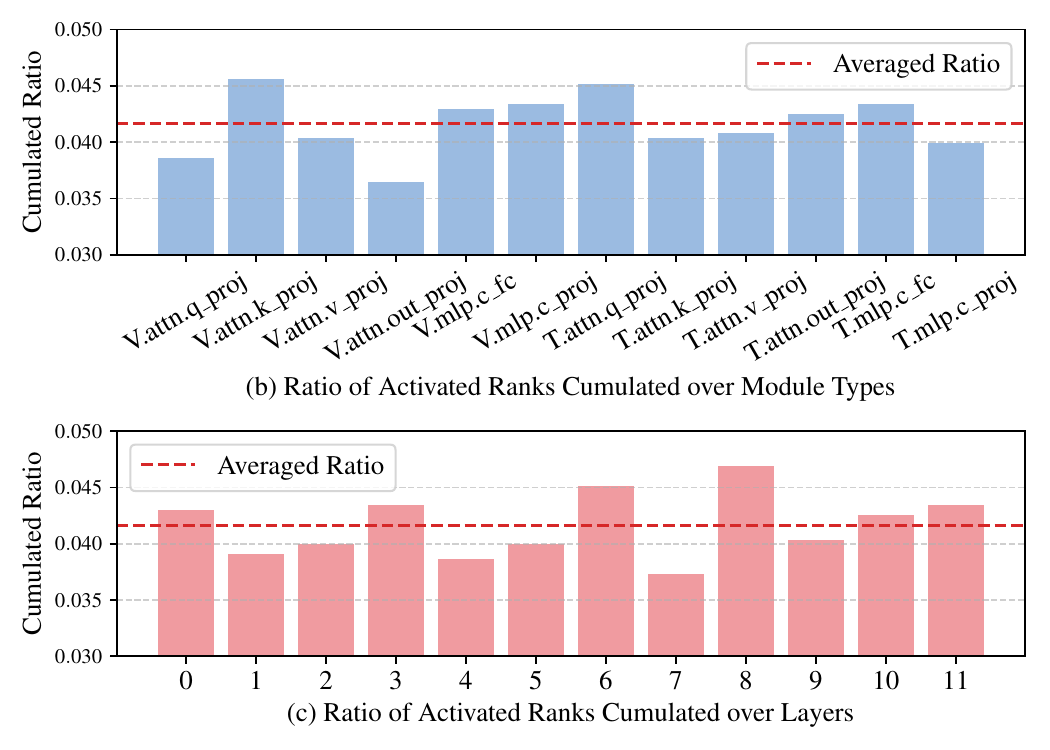}\\
        \caption{Visualization and statistical analysis of rank activation on the Caltech101 dataset using our proposed method.}
        \label{fig:supp_caltech_rank_stat}
    \end{minipage}
    \hfill
    \begin{minipage}[t]{0.45\textwidth}
        \centering
        \includegraphics[width=\textwidth]{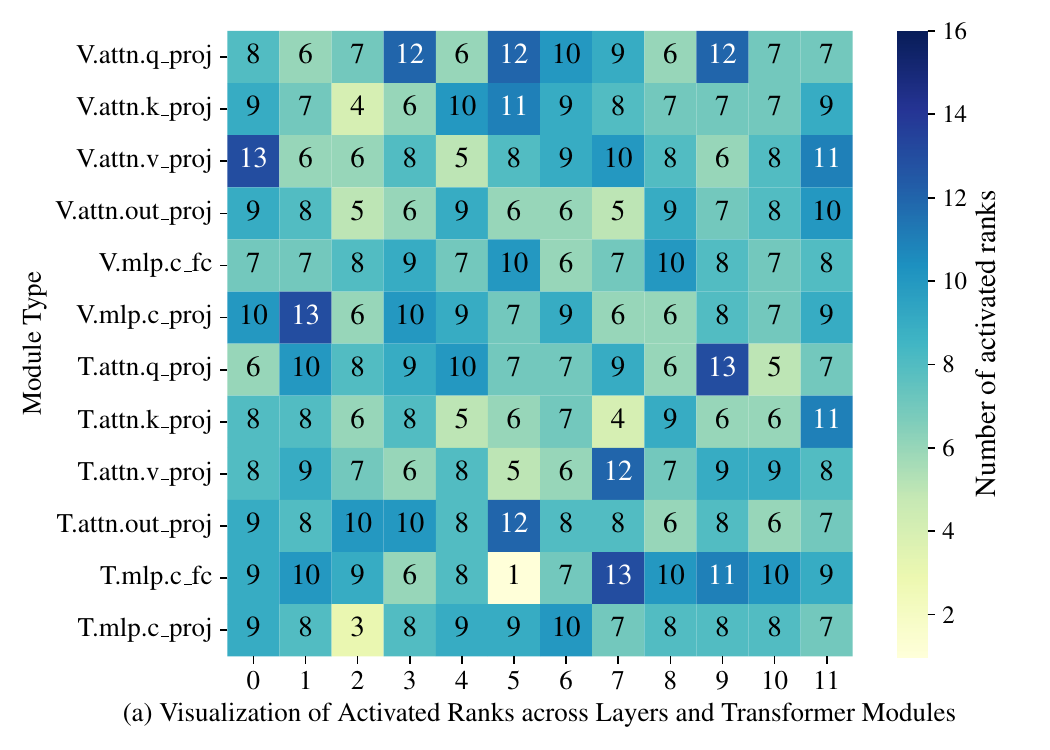}\\
        \includegraphics[width=\textwidth]{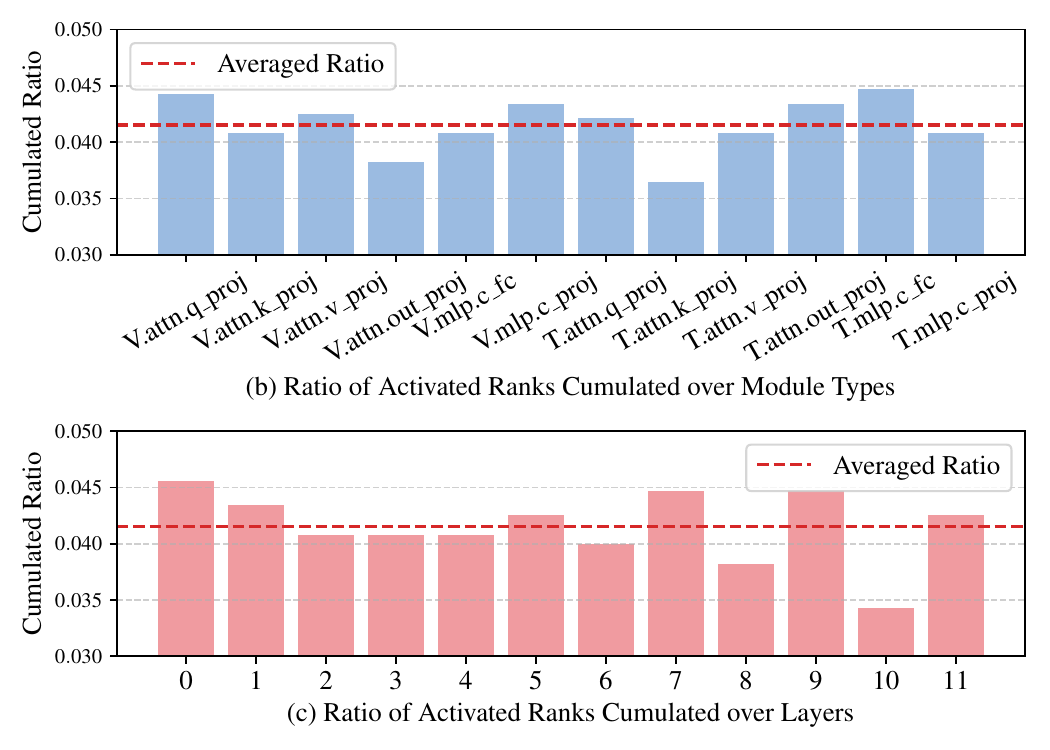}\\
        \caption{Visualization and statistical analysis of rank activation on the DTD dataset using our proposed method.}
    \end{minipage}
\end{figure*}

\begin{figure*}
    \begin{minipage}[t]{0.45\textwidth}
        \centering
        \includegraphics[width=\textwidth]{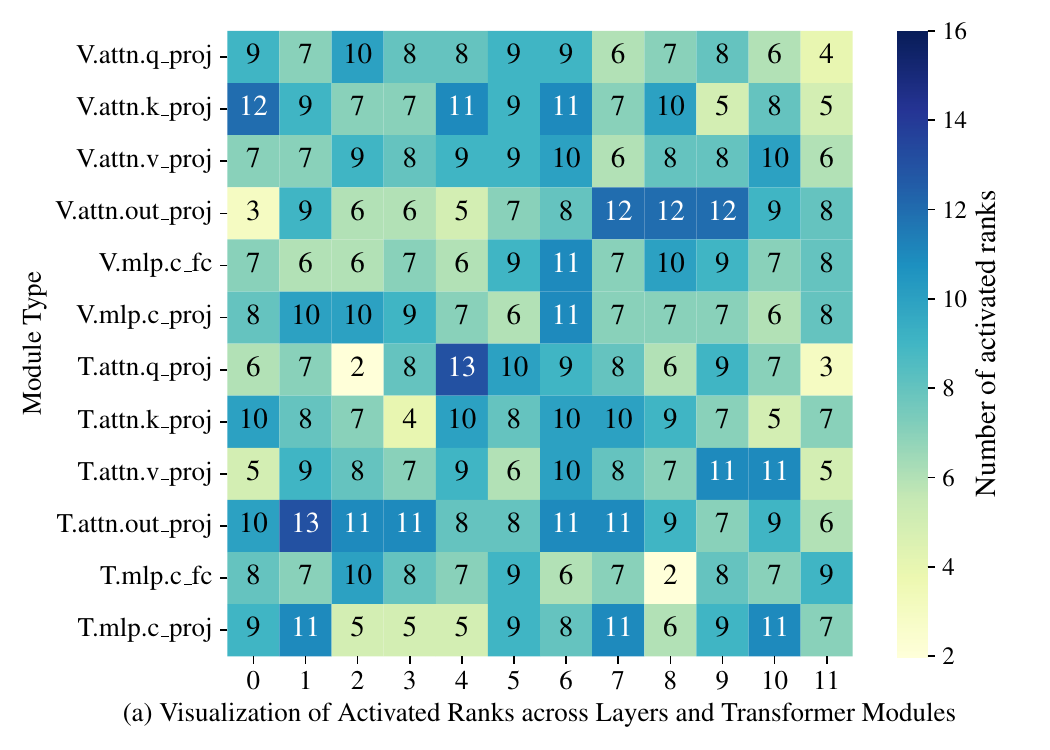}\\
        \includegraphics[width=0.95\textwidth]{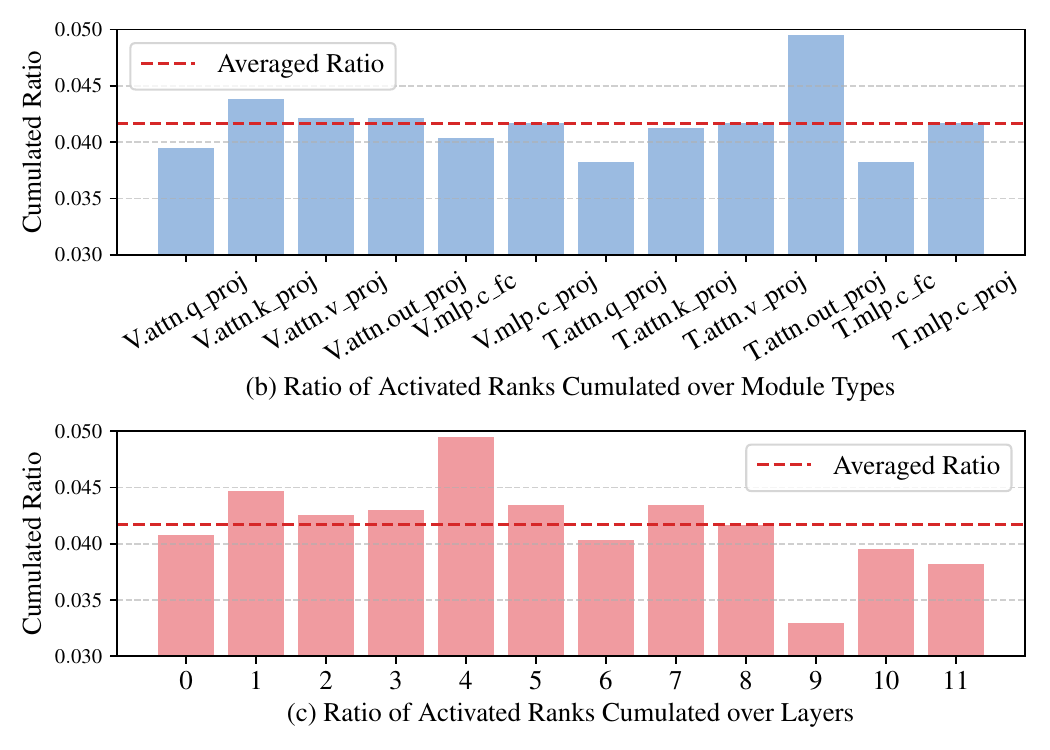}\\
        \caption{Visualization and statistical analysis of rank activation on the Stanford Cars dataset using our proposed method.}
    \end{minipage}
    \hfill
    \begin{minipage}[t]{0.45\textwidth}
        \centering
        \includegraphics[width=\textwidth]{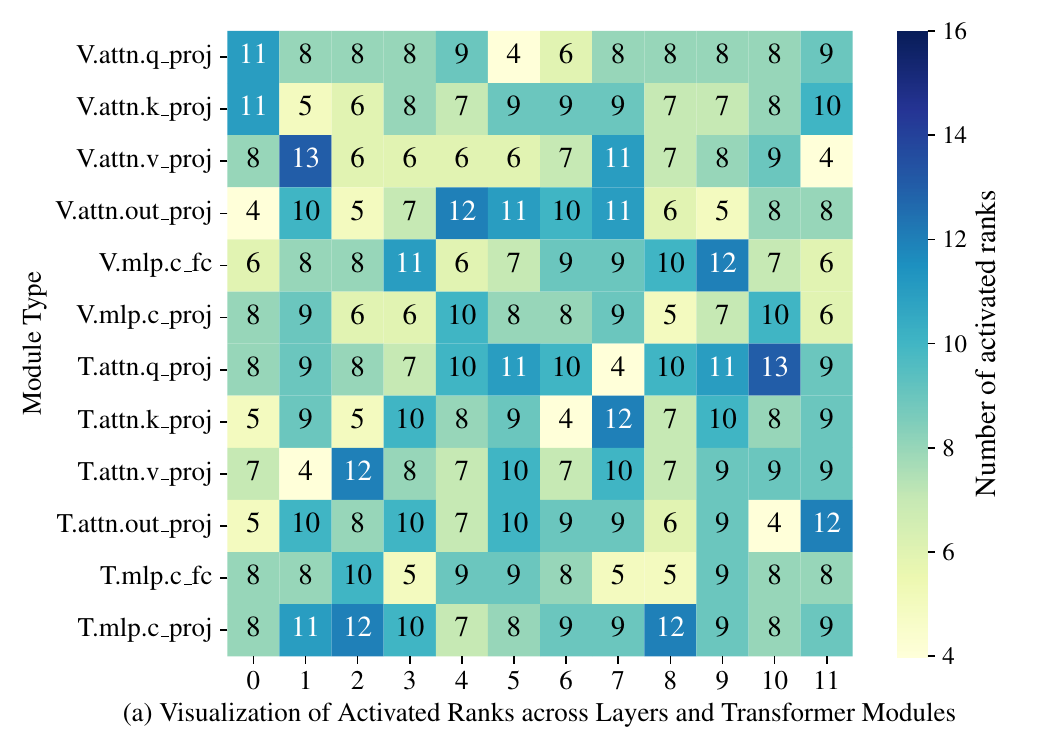}\\
        \includegraphics[width=0.95\textwidth]{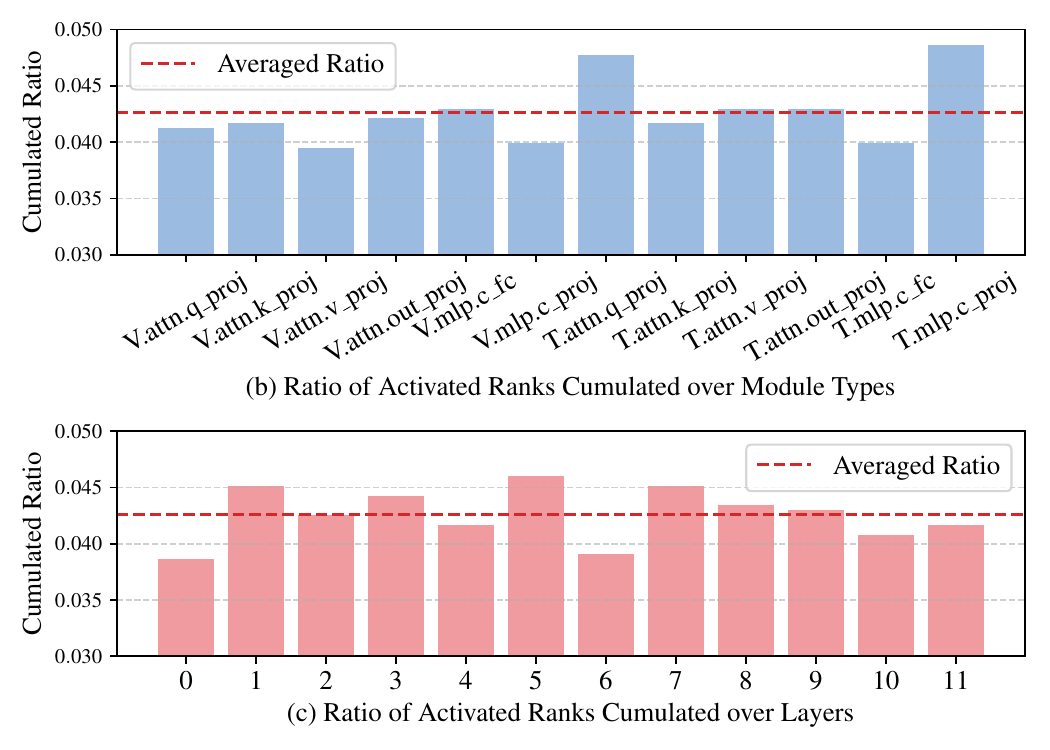}\\
        \caption{Visualization and statistical analysis of rank activation on the SUN397 dataset using our proposed method.}
        \label{fig:supp_sun_rank_stat}
    \end{minipage}
\end{figure*}

In Sec. \ref{sec:discussion}, Fig. \ref{fig:aircraft_rank_stat} and Fig. \ref{fig:pets_rank_stat}, we visualize and analyze the dynamic ranks assigned to each module for the Aircraft and Oxford Pets datasets under the X-TAIL experimental settings. Here, we extend these visualizations and analyses to additional datasets, as shown from Fig. \ref{fig:supp_caltech_rank_stat} to Fig. \ref{fig:supp_sun_rank_stat}.

The visualizations, along with statistical analyses of transformer modules and layers, reveal distinct rank allocation patterns across datasets. These findings suggest that the rank of parameter updates needed for good downstream improvement and knowledge retention varies across data types.

\begin{figure*}[h]
    \begin{minipage}[t]{0.45\textwidth}
        \centering
        \includegraphics[width=\textwidth]{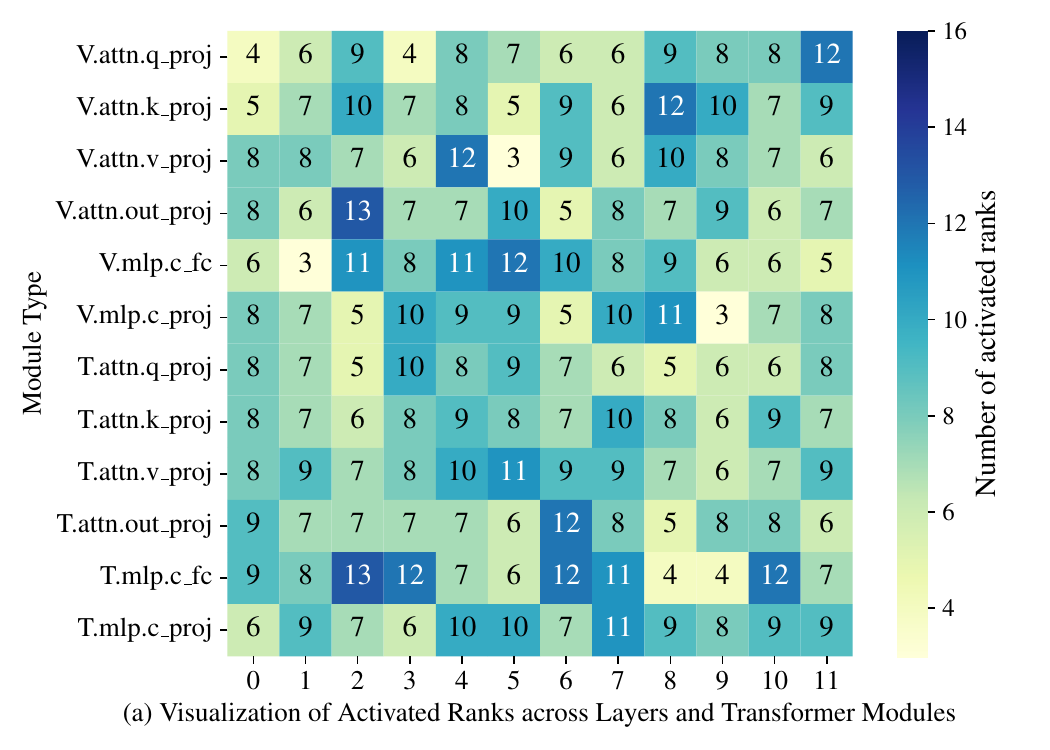}\\
        \includegraphics[width=0.95\textwidth]{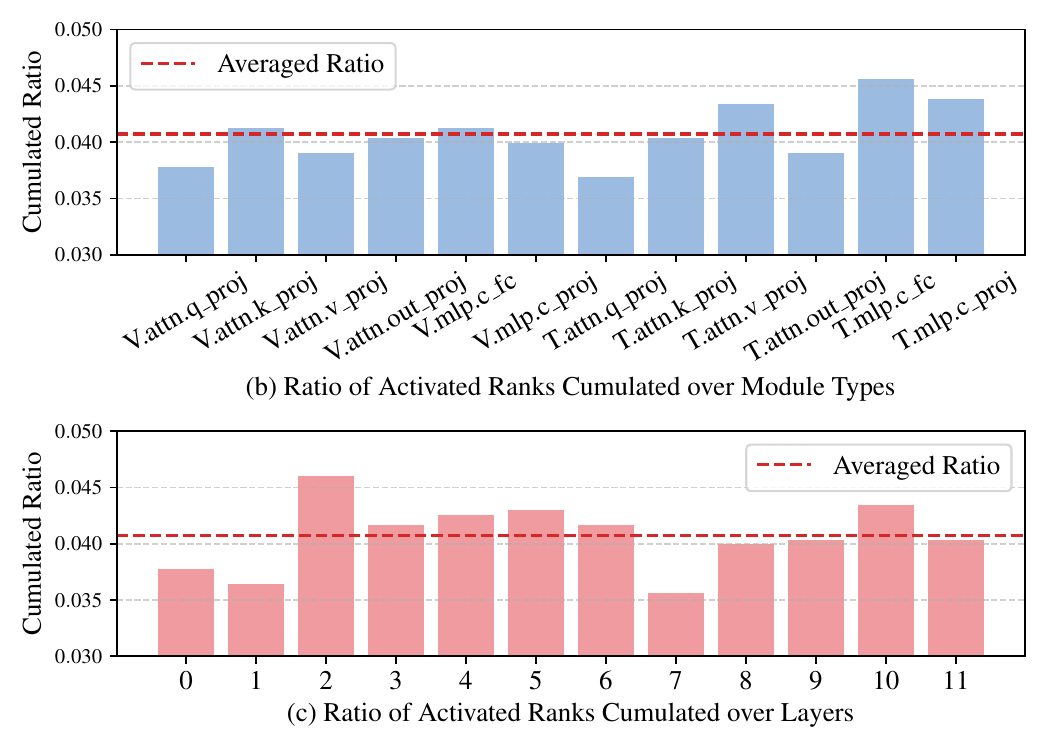}\\
        \caption{Visualization and statistical analysis of rank activation on the EuroSAT dataset using our proposed method.}
    \end{minipage}
    \hfill
    \begin{minipage}[t]{0.45\textwidth}
        \centering
        \includegraphics[width=\textwidth]{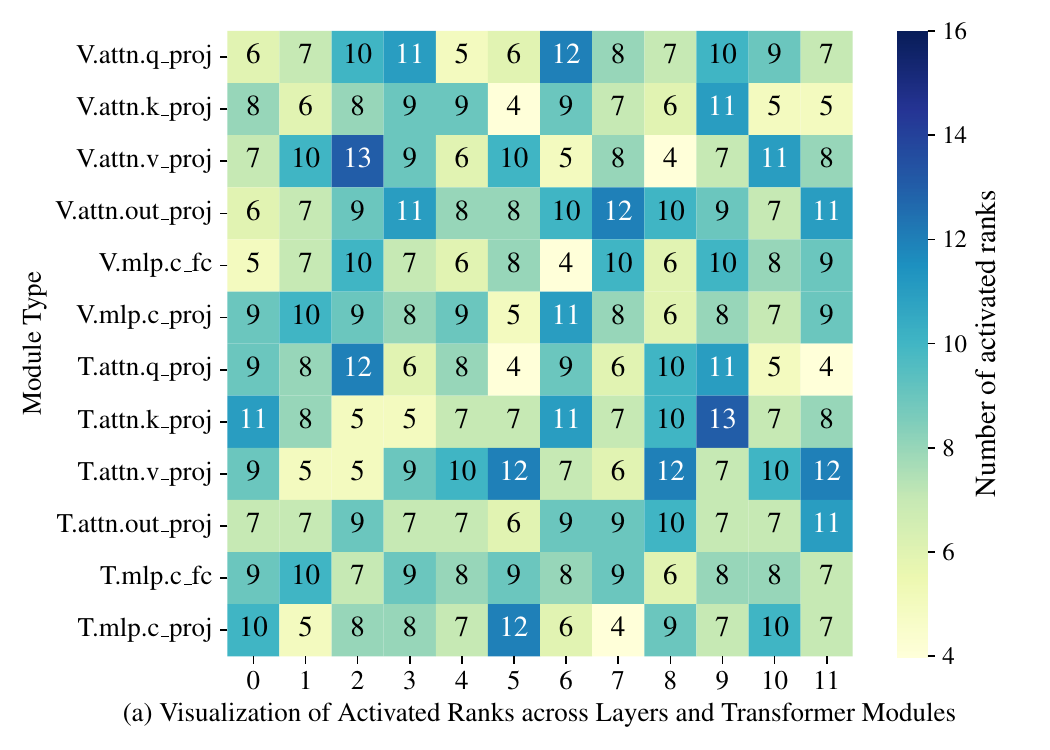}\\
        \includegraphics[width=0.95\textwidth]{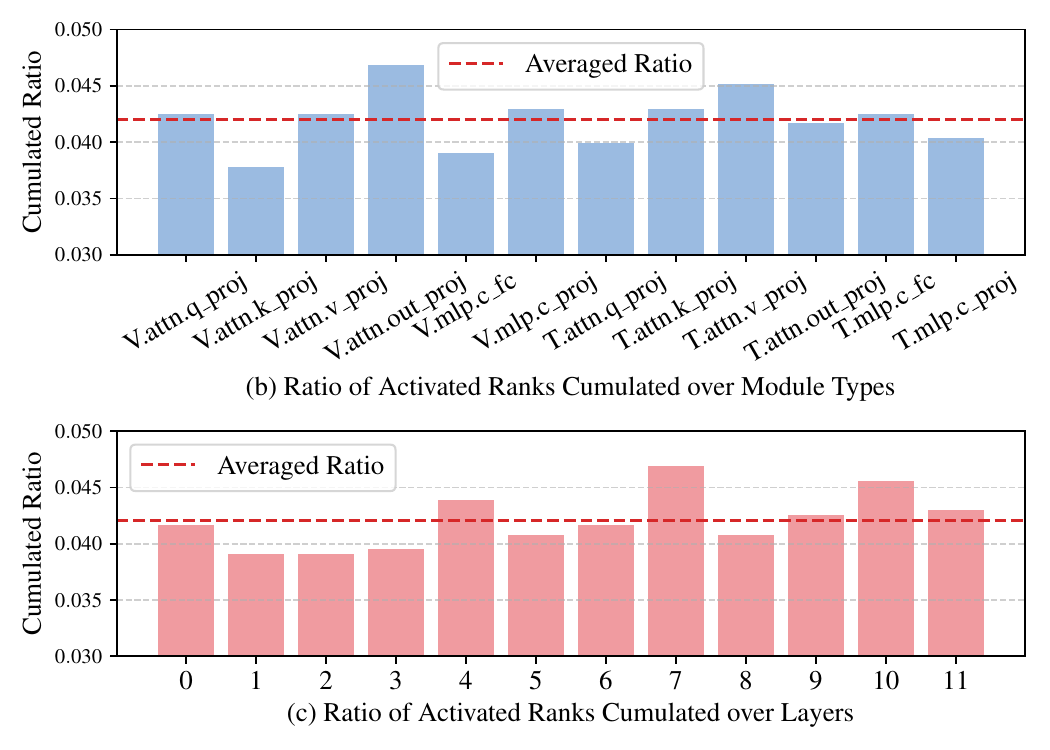}\\
        \caption{Visualization and statistical analysis of rank activation on the Flowers dataset using our proposed method.}
    \end{minipage}
\end{figure*}

\begin{figure*}[h]
    \begin{minipage}[t]{0.45\textwidth}
        \centering
        \includegraphics[width=\textwidth]{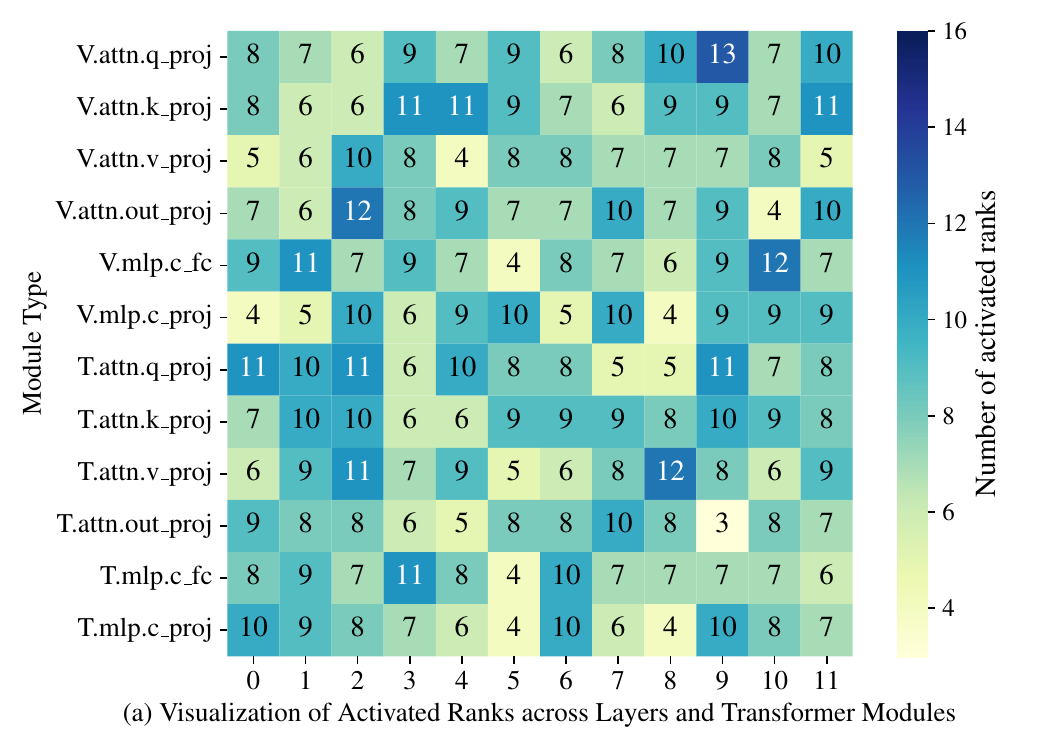}\\
        \includegraphics[width=0.95\textwidth]{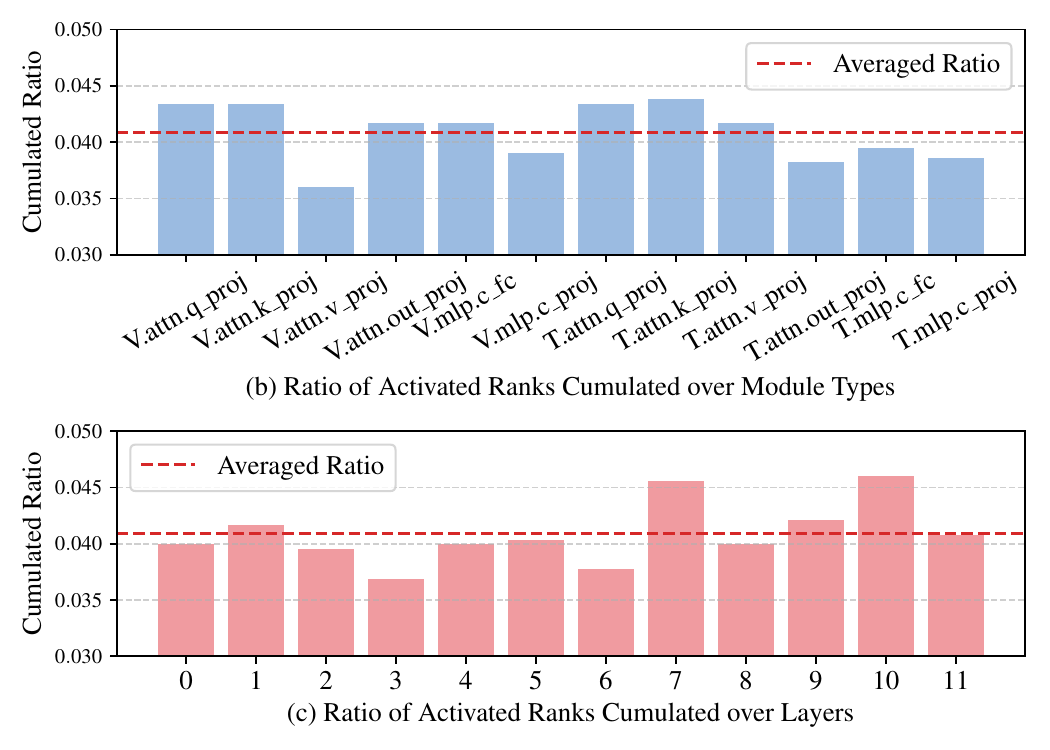}\\
        \caption{Visualization and statistical analysis of rank activation on the Food101 dataset using our proposed method.}
    \end{minipage}
    \hfill
    \begin{minipage}[t]{0.45\textwidth}
        \centering
        \includegraphics[width=\textwidth]{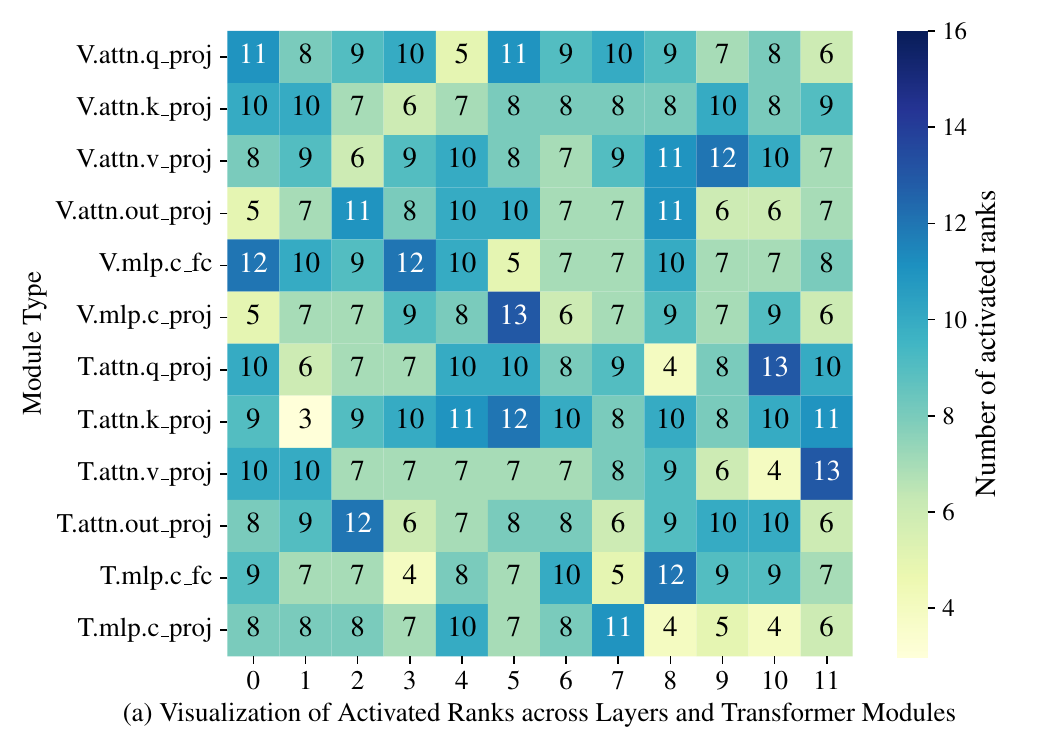}\\
        \includegraphics[width=0.95\textwidth]{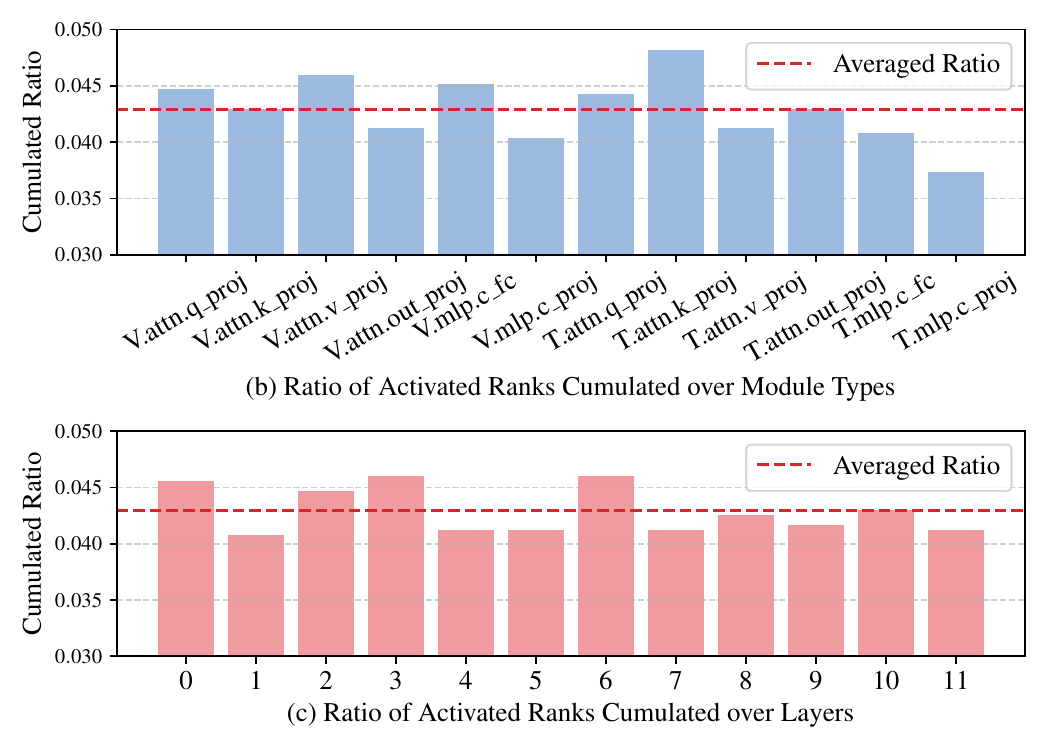}\\
        \caption{Visualization and statistical analysis of rank activation on the MNIST dataset using our proposed method.}
    \end{minipage}
\end{figure*}

\begin{table}[!t]
\centering
\small
\begin{tabular}{@{}lccc@{}}
\toprule
 & Transfer  & Average & Last  \\ \midrule
Zero-shot & 51.03 & -- & 46.45 \\
RAIL-Primal$\dagger$ & 51.03 & 62.21 & 75.72 \\ \midrule
\textbf{\ours$\dagger$} & \textbf{51.59} & \textbf{63.60} & \textbf{76.36} \\ \bottomrule
\end{tabular}
\caption{Averaged performance using BLIP.}
\label{tab:blip}
\end{table}

\begin{table}[!t]
\centering
\resizebox{\linewidth}{!}{
\small
\begin{tabular}{@{}lccc@{}}
\toprule
\multirow{2}{*}{} & \multicolumn{3}{c}{Acc. (higher is better)  /  Final Used Params. (lower is better)} \\ \cmidrule(l){2-4}
 & Caltech101 & Cars & OxfordPet   \\ \midrule
Zero-shot & 88.40  & 64.70  & 89.00  \\
LoRA ($r=16$) & 96.43 / 4.4M & 80.66 / 4.4M & 92.94 / 4.4M  \\ \midrule
\textbf{\ours~ ($r_{init} = 16$)} &  \textbf{97.20 / 3.95M} & \textbf{81.61 / 3.75M} & \textbf{94.44 / 3.85M}  \\ \bottomrule
\end{tabular}
}
\caption{Performance of 16-shot PEFT on downstream tasks.}
\label{tab:peft}
\end{table}

\end{document}